\documentclass[12pt]{article}
\usepackage[utf8]{inputenc}
\usepackage[round]{natbib}
\usepackage{url}
\usepackage{times}
\usepackage{latexsym}
\usepackage{graphicx,psfrag,epsf}
\usepackage{amsfonts}
\usepackage{amssymb}
\usepackage{amsmath}
\usepackage{enumerate}
\usepackage{tikz}
\usetikzlibrary{calc}
\usetikzlibrary{chains}
\usetikzlibrary{scopes}
\usetikzlibrary{shapes.geometric}
\usetikzlibrary{trees}
\usetikzlibrary{topaths}
\usetikzlibrary{positioning}
\usepackage{color}
\newcommand{\ns}{2em}
\newcommand{\veu}{3*\ns} % vertical edge unit
\newcommand{\heu}{4*\ns} % horizontal edge unit
\tikzstyle{hidden}=[draw, circle, fill=gray!50, minimum width=\ns, inner sep=0]
\tikzstyle{observed}=[draw, circle, minimum width=\ns, inner sep=0]
\tikzstyle{observedcond}=[draw=gray!95, circle, minimum width=\ns, inner sep=0, text=gray!80]
\tikzstyle{eliminated}=[draw, circle, minimum width=\ns, color=gray!80, inner sep=0]
\tikzstyle{empty}=[]
\tikzstyle{arrow}=[->, >=latex, line width=1pt]
\tikzstyle{edge}=[-, line width=1pt]
\tikzstyle{lightarrow}=[->, >=latex, line width=1pt, fill=gray!80, color=gray!80]

\newtheorem{definition}{Definition}

\newcommand{\TW}{{\mathcal TW}}

\newcommand{\Aplus}{\mathop{\oplus}}
\newcommand{\Atimes}{\mathop{\odot}}
\newcommand{\BigAplus}{\mathop{\mbox{\Large $\oplus$}}}
\newcommand{\BigAtimes}{\mathop{\mbox{\Large $\odot$}}}
\newcommand{\bfX}{{\mathbf{X}}}
\newcommand{\bfx}{{\mathbf{x}}}
\newcommand{\bfH}{{\mathbf{H}}}
\newcommand{\bfh}{{\mathbf{h}}}
\newcommand{\bfO}{{\mathbf{O}}}
\newcommand{\bfo}{{\mathbf{o}}}

\begin{document}
%-----------------------
% 

\title{Exact or approximate inference in graphical models: why the choice is dictated by the treewidth, and how variable elimination can be exploited}

\author{Nathalie Peyrard$^a$, Marie-Jos\'ee Cros$^a$, Simon de Givry$^a$, Alain Franc$^b$, \\
  St\'ephane Robin$^{c,d}$,
  R\'egis Sabbadin$^a$, Thomas Schiex$^a$, Matthieu Vignes$^{a,e}$\\
\\
\\
  $^a$ INRA UR 875 MIAT,\\
  Chemin de Borde Rouge, 31326 Castanet-Tolosan, France \\
  $^b$ INRA UMR 1202, Biodiversit\'e, G\`enes et
  Communaut\'es, \\
  69, route d'Arcachon,  Pierroton,  33612 Cestas
  Cedex, France \\
  $^c$  AgroParisTech, UMR 518 MIA, 16 rue Claude Bernard, Paris 5e, France\\
  $^d$  INRA, UM R518 MIA, 16 rue Claude Bernard, Paris 5e, France \\
  $^e$  Institute of Fundamental Sciences, Massey University, \\
  Palmerston North, New Zealand}

\date{}

\maketitle

\begin{abstract}
Probabilistic graphical models offer a  powerful framework to account for the dependence structure between variables, which is represented as a graph. However, the dependence between variables may render inference tasks intractable. In this paper we review techniques exploiting the  graph structure for exact inference, borrowed from optimisation and computer science. 
They are  built on the principle of variable elimination whose complexity is dictated in an intricate way by the order in which variables are eliminated. The so-called  treewidth of the graph characterises this  algorithmic complexity:  low-treewidth graphs can be processed efficiently. 
The first message that we illustrate is therefore  the idea that for inference in graphical model, the number of variables is not the  limiting factor, and it is worth checking for the treewidth before turning to
approximate methods. We show how algorithms providing an upper bound of the treewidth can be exploited to derive a 'good' elimination order enabling to perform exact inference. 
The second message  is that when the treewidth is too large, algorithms for approximate inference linked to the principle of variable elimination, such as loopy belief  propagation and variational approaches, can lead to accurate results while being much less time consuming than Monte-Carlo approaches. We illustrate the techniques reviewed in this article on benchmarks of inference problems in genetic linkage analysis and computer vision, as well as  on hidden variables restoration in coupled Hidden Markov Models. 
\end{abstract}


\maketitle

\section{Introduction}

\label{sec: intro}

%%%%pourquoi les GM c'est bien
%Most real  complex systems are made up or modelled by elementary
%objects which locally interact with each other. 
Graphical models ~\citep{Lauritzen96,B06,KF09,barber2012,Murphy2012} are formed by variables linked to
each other by  stochastic relationships. They enable to
model dependencies in possibly high-dimensional heterogeneous data and to
capture uncertainty. 
Graphical models have been applied in a wide range of areas when elementary units locally interact with each other, like image analysis~\citep{Solomon2011}, speech recognition~\citep{baker2009}, bioinformatics~\citep{liu2009,maathuis2010,hohna2014} and ecology~\citep{illian2013,bonneau2014,carriger2016} to name a few.

%%%comment est-ce que l'on peut faire des calculs dans un GM
 In real applications a large number of random variables with a complex dependency structure are involved.
As a consequence,
inference tasks such as the calculation of a normalisation constant, of a
marginal distribution or of the mode of the joint distribution can be challenging.
Three main approaches exist to evaluate such quantities for a given distribution $\Pr$ defining a graphical model:
$(a)$ compute them in an exact manner;
$(b)$ use a stochastic algorithm to sample from the distribution $\Pr$ to get
(unbiased) estimates;
$(c)$ derive an approximation of $\Pr$ for which the exact calculation is
possible.
% The first approach obviously requires to develop clever algorithms since, for
% a model involving (say) $n = 10^6$ variables, a naive implementation may require
% resources in time beyond the age of universe even on largest computers, as the
% complexity (the number of elementary operations) increases exponentially with
% $n$. 
Even if appealing,  exact computation on $\Pr$ can lead to very time and memory consuming procedures for large problems.
%, since the
% number of elements to store or elementary operations to perform increases exponentially with $n$ the number %of  random variables. 
The second approach is probably the most widely used by statisticians and modellers. Stochastic
algorithms such as Monte-Carlo Markov Chains (MCMC) \citep{RC04}, Gibbs sampling~\citep{Geman84,casella1992} and particle
filtering~\citep{GSS93} have become standard tools in many fields of application using
statistical models. The last approach includes  variational
approximation techniques \citep{WaJ08}, which are starting to become common practice in computational statistics.
In essence, approaches of type $(b)$ provide an approximate answer to an exact problem whereas
approaches of type $(c)$ provide an exact answer to an approximate problem.

%%%objectif du papier
In this paper, we focus on approaches of type $(a)$ and $(c)$, and we will review
techniques for exact or approximate inference in graphical models
 borrowed from both optimisation and computer science. They are computationally efficient,
yet not always standard in the statistician toolkit.
% It turns out that the
% graphical model framework is most convenient to bridge the gap between
% statistics and computer sciences as it has been used for a long time in both
% communities. note that this framework is also useful to design stochastic
% algorithms. 
% A graphical model is defined as an $n$-variate distribution $p(x)$ where $x = (x_1, \dots
% x_n)$, that can be factored over a fixed family of subsets $B \in \mathcal{B}$
% of $\{1, \dots, n\}$: $p(x) \propto \prod_B \psi_B(x_B)$. The
% associated graphical representation  is a graph $G$ with vertices $V = \{1, \dots, n\}$ where an
% edge is drawn between vertices $i$ and $j$ if they both belong to some subset $B$
The characterisation of the structure of the graph $G$ associated to a graphical model
 (precise definitions are given in Section \ref{sec: GM}) enables both to determine if the
exact calculation  of the quantities of interest (marginal distribution, normalisation constant, mode)
 can be implemented efficiently and to derive a class of operational algorithms.
When the exact calculation cannot be achieved efficiently, a similar analysis of the problem enables the practitioner to design algorithms to compute an approximation 
of the desired quantities with an associated acceptable complexity. 
Our aim is to provide the reader with the key elements to understand the power of these tools for statistical inference in graphical models. 

% \SR{The trick behind is very simple: make a clever use of distributivity
%between
% $\times$ and $+$. Indeed, evaluating $ab+ac=a(b+c)$ requires two
%multiplications
% and one addition in the former case, and one multiplication and one addition
%in
% the latter. }

%%%% variable elimin et algebre
The central algorithmic tool we focus on in this paper is the variable elimination  concept \citep{Bertele72}.
In Section \ref{sec: exactinference} we adopt a unified algebraic presentation of the different inference tasks
 (marginalisation, normalising constant or mode evaluation)
to emphasise that each of them can be solved using a particular case of a variable elimination scheme. Consequently, the work done to demonstrate that variable elimination is efficient for one task passes on to the other ones.
The key ingredient to design efficient algorithms based on variable elimination  is the clever use of
distributivity  between algebraic operators. For instance  distributivity of the product ($\times$) over the sum ($+$) enables to write
 $(a\times b) + (a\times c) = a\times(b+c)$ and evaluating the left-hand side of this equality requires two multiplications
and one addition while evaluating the right-hand side requires one multiplication and one addition. 
Similarly since $\max(a+b, a+c)= a + \max(b,c)$ it is more efficient to compute the right-hand side from an algorithmic point of view. 
Distributivity enables to minimise the number of operations. 
To perform variable elimination, associativity and commutativity properties are also required, 
% Therefore,
% the ring structure of space $\mathbb{R}^+$ on which each $\psi_B$ takes its
% value is the key ingredient. More precisely, the full ring structure is not
% required, as existence of an inverse for $+$ is not used. Therefore,
and the algebra
behind is that of semi-ring (from which some notations will
be borrowed). 
% Emphasising the algebraic structure on which calculations rely
% enables to present in a common framework calculations for marginalisation or
% partition function on one hand, and for mode on the other. \SR{}{[A-t-on besoin
% d'\^etre aussi pr\' ecis ici ?]} The reason is that
% distributivity is fulfilled for the semi-ring $(\mathbb{R}, \max, +)$ where
% $\max$ plays the role of $+$ and $+$ plays the role of $\times$, as $a +
% \max(b,c) = \max(a+b, a+c)$.
 Inference algorithms using the distributivity  property have been known and
published in the Artificial Intelligence and Machine Learning literature under
 different names, such as sum-prod, or max-sum ~\citep{Pearl88,B06}. They
are typical examples of variable elimination procedures.

% \SR{A key tool to organise calculations for best use of distributivity is to
% translate the dependence structure between random variables as a graph: two
% sites $i,j \in V$ are linked by an edge if there exists a common subset $B$
%they
% belong to. Then, the way to organise use of distributivity is straightforward
% when this graph is a tree: it follows an elimination order of vertices of degree
% one. }

%%% la treewidth comme ingredient cle pour characteriser la complexite de VA
Variable elimination relies on the choice of an order of elimination of the
variables, via successive marginalisation or maximisation operations. The calculations are performed according to this ordering when applying  distributivity.
The topology of the graph $G$ provides key information to optimally organise the
calculations as to minimise the number of elementary operations to perform. 
For example, when the graph is a tree, the most efficient elimination order corresponds to  eliminating recursively the vertices of degree one. One starts from the leaves towards the root, and inner nodes of higher degree successively become leaves.
% Algorithms to compute the
% partition function, 
% marginals on all singletons and the mode are then linear in $n$. It turns out
% that this
%  particular case is an example of much deeper and general results on graphs,
% which enable to characterise a key feature of a graph: its
% treewidth. 
The  notion of an optimal  elimination order for inference in an arbitrary graphical model is
closely linked to the notion of treewidth of the associated graph $G$.  
We  will see in Section \ref{sec: exactinference} the reason why
inference algorithms based on variable elimination with the best elimination order are of linear complexity in
$n$, the number of variables/nodes in the graph, i.e. the size of the graph, but exponential complexity in the  treewidth. Therefore treewidth is one the main characterisation of $G$ to determine if exact inference is possible in practice or not. 
This notion  has lead to the development of several works for solving apparently complex inference problems, which have then been applied  in biology (e.g. \citealt{tamura2014}). More details on these methodological and applied results are provided in the Conclusion Section.

% \SR{For example, exact algorithm exist for computation in quadratic time
% for series-parallel graphs, which have treewidth equal to 2. 
%
% \SR{These results derive from very powerful and deep theorems in discrete
% mathematics which derive in turn from the graph minor theorems, which are
% existence theorems: we know there exists an algorithm in polynomial time, and
% even the degree of the polynomial. But this does not tell how to derive and
% implement this algorithm, apart from some specific cases (as trees, chordal
% graphs, and series-parallel graphs). 
The concept of treewidth has been proposed   in parallel in  computer science 
 \citep{bodlaender94}, in discrete mathematics and graph minor theory  (see \citealt{RS86,L05}).
 Discrete mathematics  existence theorems~\citep{RS86}
establish that there exists an algorithm for computing the treewidth of any graph with 
complexity polynomial in $n$ (but exponential in the treewidth), and the degree of the polynomial is determined. However, this result does not tell
how to derive and implement the algorithm, apart from some very specific cases such as
trees, chordal graphs, and series-parallel graphs~\citep{D65}.  
Section \ref{sec: tw} introduces the reader to several state-of-the-art algorithms that provide an upper bound of the treewidth, together with an associated elimination order. These algorithms are therefore useful tools to test if exact inference is achievable and, if applicable, to derive an exact inference algorithm based on variable elimination. Their behaviour is illustrated on benchmarks borrowed from combinatorial optimisation competitions.

% \SR{If we enlarge the set onto which the potentials $\psi_B$ take their values
% in the Boolean semi-ring $\{\textit{true}, \textit{false}\}$ then a graphical
% model is equivalent to a Constraint Satisfaction Problem~\cite{HB06}, a family
% of problem which has been thoroughly studied in artificial intelligence.
% Multiplication is logical AND ($\land$) and sum is logical OR ($\lor$). 
% Therefore, we show in this manuscript how this notion of treewidth enables to
% link CSP and graphical models, and efficient algorithms derived in one field to
% percolate to the other. We present the tools and machinery needed for that. 
% This is the main objective of this paper. }

%%%message passing, reparametrisation et WCSP
Variable elimination also lead to message passing algorithms \citep{Pearl88} which are now common tools in 
computer science or machine learning  for marginal or mode evaluation. More recently, these algorithms have been reinterpreted 
as a way to re-parameterise the original graphical model into  an updated one with different potential functions by still representing the same join distribution  ~\citep{KF09}. We  explain in Section \ref{sec: VE2MP} how re-parametrisation can be used as a 
pre-processing tool to obtain a new parameterisation with which inference becomes simpler.
Message passing is not the only way to perform re-parametrisation, and we discuss alternative  efficient
algorithms proposed in the context of 
Constraint Satisfaction Problems (CSP, see \citealt{HB06}). These latter ones have, to the best of our knowledge, not yet been exploited in the context of graphical models.

% Also, we will show that some statistical tasks defined above on graphical models
% can be rephrased as constraint satisfaction problems (CSP, \cite{HB06})
% originally defined on the Boolean semi-ring. As CSP have been intensively
% studied in artificial intelligence, the link between CSP and graphical models
% allows to borrow efficient algorithms derived in one field to percolate to the
% other.  

%%%%message passing et methodes variationnelles
As emphasised above, efficient exact inference algorithms can only be designed
for graphical models with limited treewidth, i.e. much less than the number of vertices. Although this is not the case for many graphs, the principles of variable elimination and  message passing for a 
tree can be applied to any graph leading to heuristic inference algorithms.
The most famous heuristics is the Loopy Belief Propagation algorithm (LBP, see \citealt{KFL01}).
% By  heuristics~\cite{Pearl85} we mean a  method not defined as the
% optimisation of particular criterion, as opposed to an approximation.
We recall in Section \ref{sec: variational} the result that establishes LBP as a variational approximation
method. Variational methods rely on the choice of a  distribution which renders
inference  easier. They approximate the original complex graphical model. The
approximate distribution is chosen within a  class of models
for which efficient inference algorithms exist, that is models with small treewidth (0, 1 or 2 in practice). 
We   review some standard choices of approximate distributions, each of them corresponds to a different underlying treewidth.

Finally, Section \ref{sec: illust CHMM} illustrates the techniques reviewed in the article, on the case of Coupled   Hidden Markov Model (CHMM, see \citealt{Brand97}). We first compare them on the problem of mode inference in a CHMM devoted to the study of pest propagation. Then we exemplify the use of different variational methods for  EM-based parameter estimation in CHMM.

%% j'ai reference les section au fil de l'intro
%This paper is organised as follows. In Section~2 graphical models and main
% inference tasks in graphical models are presented. Section~3 is the core of this
% manuscript, where exact inference algorithms running in polynomial time are
% presented. We show they come down to finding an optimal variable elimination
% scheme. We emphasise an algebraic framework enabling a common treatment for sums
% (marginalisation) and maximum (computation of the mode). In
% Section~4, the key notion of treewidth is presented, and some ways to compute
% it, either exactly, or approximately. It is an upper bound of the complexity of
% algorithms in Section~3. In Section~5, some heuristics and approximation methods
% for inference are recalled, combining an approximation of the graphical model by
% truncation and approximation of the inference by maximum entropy principles.
% Finally, the framework of CSP is developed in Section~6 where we emphasise that
% a network of costs functions can be translated into a graphical model via a simple 
% exponential transformation. We end the
% manuscript by our choice of some challenges at the front of current research is
% these fields. 

\section{Graphical Models}
%Graphical models

\label{sec: GM}

\subsection{Models definition}
% \todoTS[inline]{Est-il utile de faire intervenir des points ? Ne peut
%   on pas partir directement d'un ensemble fini de variables aleatoires
%   $\mathcal{X}$ indexe ?}
% \todoNP[inline]{Thomas, OK pour ne pas faire intervenir les points, par contre
% la notation A au lieu de $x_A$ ne nous (= RS, AF, NP) semble pas la plus
% naturelle, donc je suis revenue a $x_A$}.
%Let us consider $V$ a finite set of $n$ points , with elements indexed
%on $\{1, \ldots, n\}$.  

Consider a stochastic system defined by a set of random variables
$\mathbf{X} = (X_1,\ldots,X_n)^\top$. Each variable $X_i$ takes values in
$\Lambda_i$.  A realisation of $\mathbf{X}$ is denoted $\mathbf{x} =
(x_1,\ldots,x_n)^\top$, with $x_i \in \Lambda_i$. The set of all possible
realisations is called the state space, and is denoted $\Lambda =
\prod_{i=1}^n\Lambda_i$.  If $A$ is a subset of $V = \{1, \ldots,n\}$, then
$X_A$, $x_A$ and $\Lambda_A$ are  respectively the subset of random
variables $\{X_i, i \in A\}$, a possible realisation $\{x_i, i \in
A\}$ of $X_A$ and the state space of $X_A$  respectively.  If $p$ is the joint probability
distribution of $\bfX$ on $\Lambda$, we denote for all $\mathbf{x} \in \Lambda$
$$
 p(\mathbf{x}) = \Pr(\mathbf{X} = \mathbf{x}).
$$

Note that we focus here on discrete variables (we will discuss
inference in the case of continuous variables on examples in Section
\ref{sec: conclu}).
A joint distribution $p$ on $\Lambda$ is said to be a
\textit{probabilistic graphical model}~\citep{Lauritzen96,B06,KF09}
indexed on a set $\mathcal{B}$ of parts of $V$ if there exists a set
$\Psi = \{ \psi_B\}_{B \in \mathcal{B}}$ of maps from $\Lambda_B$ to
$\mathbb{R}^+$, called \textit{potential functions}, indexed by
$\mathcal{B}$ such that $p$ can be expressed in the following factorised
form:
\begin{equation}\label{Eq: GM}
  p(\bfx) = \frac{1}{Z}\prod_{B \in \mathcal{B}} \psi_B(x_B),
\end{equation}
where $Z = \sum_{\bfx \in \Lambda} \prod_{B \in \mathcal{B}} \psi_B(x_B)$
is the normalising constant, also called partition function. The
elements  $B \in \mathcal{B}$ are the scopes of the potential functions
and $|B|$ is the arity of the potential function $\psi_B$. The set of
scopes of all the potential functions involving variable $X_i$ is
denoted $\mathcal{B}_i = \{B\in \mathcal{B}: i \in B\}$
%$\mathcal{B}_i = \bigcup_{B\in \mathcal{B} \atop  B \ni i\}$.

\begin{figure}[h!]
\begin{center}
 \begin{tikzpicture}[scale=1.5,rotate=-90]
  \tikzstyle{node}=[draw,circle,fill=gray!50,minimum size = 5pt, inner sep = 2pt,font=\small]
  \tikzstyle{arc}=[->,>=latex]
  \path (0,0) node[node] (a) {1}
        (0,-1) node[node] (b) {2}
        (1,0) node[node] (c) {3}
        (1,-1) node[node] (d) {4}
        ( $(c) + (-30:1)$ ) node[node] (e) {5}
        ( $(e) + (30:1)$ ) node[node] (f) {6}
        ( $(e) + (-30:1)$ ) node[node] (g) {7}
        ( $(e) + (0:1.1)$ ) node {(a)};
  \draw[arc] (a) to (b);
  \draw[arc] (b) to (d);
  \draw[arc] (d) to (e);
  \draw[arc] (e) to (f);
  \draw[arc] (a) to (c);
  \draw[arc] (c) to (e);
  \draw[arc] (e) to (g);
  \end{tikzpicture}\quad\quad
 \begin{tikzpicture}[scale=1.5,rotate=-90]
  \tikzstyle{node}=[draw,circle,fill=gray!50,minimum size = 5pt, inner sep =
2pt]
  \tikzstyle{factor}=[draw,minimum size = 5pt]
  \path (0,0) node[node] (a) {1}
        (0,-0.5) node[factor] (ab) {}
        (0,-1) node[node] (b) {2}
        (0.5,0) node[factor] (ac) {}
        (1,0) node[node] (c) {3}
        (0.5,-1) node[factor] (bd) {}
        (1,-1) node[node] (d) {4}
        ( $(c) + (-30:1)$ ) node[node] (e) {5}
        ( $(c) + (-60:0.577350269)$ ) node[factor] (cde) {}
        ( $(e) + (30:1)$ ) node[node] (f) {6}
        ( $(e) + (30:0.5)$ ) node[factor,rotate=30] (ef) {}
        ( $(e) + (-30:1)$ ) node[node] (g) {7}
        ( $(e) + (-30:0.5)$ ) node[factor,rotate=60] (eg) {}
        ( $(e) + (0:1.1)$ ) node {(b)};
  \draw (c) -- (ac) -- (a) -- (ab) -- (b) -- (bd) -- (d);
  \draw (c) -- (cde) -- (d);
  \draw (cde) -- (e);
  \draw (e) -- (ef) -- (f);
  \draw (e) -- (eg) -- (g);
  \end{tikzpicture}\quad\quad\quad\quad
 \begin{tikzpicture}[scale=1.5,rotate=-90]
  \tikzstyle{node}=[draw,circle,fill=gray!50,minimum size = 5pt, inner sep =
2pt]
  \path (0,0) node[node] (a) {1}
        (0,-1) node[node] (b) {2}
        (1,0) node[node] (c) {3}
        (1,-1) node[node] (d) {4}
        ( $(c) + (-30:1)$ ) node[node] (e) {5}
        ( $(e) + (30:1)$ ) node[node] (f) {6}
        ( $(e) + (-30:1)$ ) node[node] (g) {7}
        ( $(e) + (0:1.1)$ ) node {(c)};
  \draw (c) -- (a) -- (b) -- (d) -- (e) -- (d) --(c) --(e) -- (f) -- (e) -- (g);
  \end{tikzpicture}\quad\quad
 \begin{tikzpicture}[scale=1.5,rotate=-90]
  \tikzstyle{node}=[draw,circle,fill=gray!50,minimum size = 5pt, inner sep =
2pt]
  \tikzstyle{factor}=[draw,minimum size = 5pt]
  \path (0,0) node[node] (a) {1}
        (0,-0.5) node[factor] (ab) {}
        (0,-1) node[node] (b) {2}
        (0.5,0) node[factor] (ac) {}
        (1,0) node[node] (c) {3}
        (0.5,-1) node[factor] (bd) {}
        (1,-1) node[node] (d) {4}
        (1,-0.5) node[factor] (cd) {}
        ( $(c) + (-30:1)$ ) node[node] (e) {5}
        ( $(c) + (-30:0.5)$ ) node[factor,rotate=60] (ce) {}
        ( $(d) + (30:0.5)$ ) node[factor,rotate=30] (de) {}
        ( $(e) + (30:1)$ ) node[node] (f) {6}
        ( $(e) + (30:0.5)$ ) node[factor,rotate=30] (ef) {}
        ( $(e) + (-30:1)$ ) node[node] (g) {7}
        ( $(e) + (-30:0.5)$ ) node[factor,rotate=60] (eg) {}
        ( $(e) + (0:1.1)$ ) node {(d)};
  \draw (c) -- (ac) -- (a) -- (ab) -- (b) -- (bd) -- (d) -- (cd) -- (c) -- (ce)
-- (e) -- (de) -- (d);
  \draw (e) -- (ef) -- (f);
  \draw (e) -- (eg) -- (g);
  \end{tikzpicture}
\end{center}
\caption{From left to right: (a) Graphical representation of a
  directed graphical model where potential functions define the conditional
  probability of each variable given its parents values; (b) The
  corresponding factor graph where every potential function is represented as a
  factor (square vertex) connected to the variables that are involved
  in it; (c) Graphical representation of an undirected graphical
  model. It is impossible from this graph to distinguish between a
  graphical model defined by a unique  potential function on vertices 3, 4 and
  5 from a model defined by 3 pairwise potential functions over each
  pair $(3,4)$, $(3,5)$ and $(4,5)$; (d) The corresponding factor
  graph, which unambiguously defines the potential functions, here
  three pairwise potential functions.}
\label{fig: graphical representation of GM}
\end{figure}
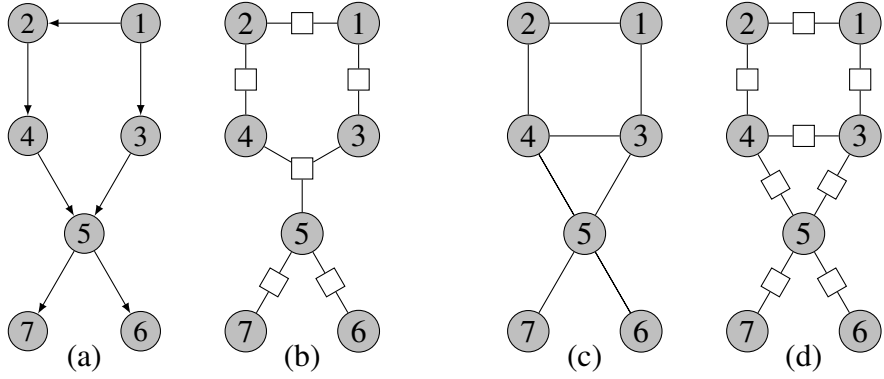

%%
One desirable property of graphical models is that of Markov
local independence: if $p(\bfx)$ can be expressed as in  (\ref{Eq: GM}), then a
variable $X_i$ is (stochastically) independent of all others in $\bfX$
conditionally to the set of variables
$X_{(\cup_{B\in\mathcal{B}_i}B) \setminus i}$.  The set  $X_{(\cup_{B\in\mathcal{B}_i}B) \setminus i}$is called the
Markov blanket of $X_i$, or its neighbourhood \citep[chapter 4]{KF09}. It is denoted $N_i$.
These conditional independences can be represented, by a
graph with one vertex per variable in $\bfX$. The question of encoding
the independence properties associated with a given distribution into
a graph structure has been widely described (e.g. \citealt[chapters 3 and 4]{KF09}), and we will not
discuss it here.  We consider the classical graph $G = (V,E)$
associated to the decomposition dictated in (\ref{Eq: GM}), where an edge
is drawn between two vertices $i$ and $j$ if there exists $B \in
\mathcal{B}$ such that $i$ and $j$ are in $B$.  Such a representation
of a graphical model is actually not as rich as the representation of (\ref{Eq: GM}). For instance, if $n=3$, the two cases $\mathcal{B} =
\{ \{1,2,3\} \}$ and $\mathcal{B} = \{ \{1,2\}, \{2,3\}, \{3,1\} \}$
are represented by the same graph $G$, namely a clique  (i.e. a fully connected set of vertices) of size 3. 
Without loss of generality, we could impose in the definition of a graphical model that 
scopes $B$ correspond to cliques of $G$. In the above example where $\mathcal{B} = \{ \{1,2\}, \{2,3\}, \{3,1\} \}$, this can be done
by defining $\psi'_{1,2,3} = \psi_{12} \psi_{23}\psi_{13}$. The original structure is then lost, and $\psi'$  is more costly to store than the original potential functions.
The factor graph representation goes beyond the limit of the representation $G$: this graphical
representation is a bipartite graph with one vertex per potential
function and one vertex per variable. Edges are only between functions and variables. An edge is present  between a
function vertex (also called factor vertex) and a variable vertex, if and only if
the variable is in the scope of the potential
function.  Figure~\ref{fig: graphical representation of GM} displays
examples of the two graphical representations.
% The notion of \textit{hypergraph} \cite{Berge} goes beyond this limit.
% An hypergraph is essentially a  graph where edges may connect more than 2
% vertices.
% The hypergraph $(V,\mathcal{B})$  associated with a probabilist graphical
% model has the
% set $V$ as the set of vertices and its hyperedges are the scopes of
% the functions $\psi_B$. If $|B|\leq 2, \forall B \in \mathcal{B}$, this
% hypergraph is a graph and is exactly the graph defined above for
% probabilistic graphical models. Otherwise, the graph $(V,E)$ of a
% probabilistic graphical model is exactly the so-called 2-section or
% primal graph of the hypergraph $(V,\mathcal{B})$.

%%
Several families of probabilistic graphical
models exist~\citep{KF09,Murphy2012}.  They can be grouped into directed and
undirected ones. The most classical directed framework is that of
\textit{Bayesian network}~\citep{Pearl88,JN07}. In a Bayesian network,  potential
functions are conditional probabilities of a variable given its
parents. In such models, trivially $Z=1$. 
There is a representation by a directed graph where an edge
is directed from a parent vertex to a child vertex (see Figure~\ref{fig: graphical representation of
  GM} (a)). The undirected graphical representation $G$ is obtained by moralisation, i.e. by adding an edge between two parents of a same variables.
%\SR{}{: $p(x) = \prod_i p(X_i=x_i | X_{par(i)} = x_{par(i)})$}.
  Undirected
probabilistic graphical models (see Figure~\ref{fig: graphical
  representation of GM} (c)) are equivalent to \textit{Markov Random
  Fields} (MRF, \citealt{L01}) as soon as the potential functions take values in
$\mathbb{R}^{+} \setminus \{0\}$.  In a Markov random field (MRF), a potential function is
not necessarily a probability distribution: $\psi_B$ is not required
to be normalised (as opposed to a Bayesian network model).

% \centerline{FIGURE \ref{fig: graphical representation of GM} ABOUT HERE}

\subparagraph{Deterministic Graphical models.} Although the
terminology of 'Graphical Models' is often used to refer to
probabilistic graphical models, the idea of describing a joint
interaction on a set of variables through local functions has also
been used in Artificial Intelligence to concisely describe Boolean
functions or cost functions, with no normalisation constraint. 
Throughout this article we regularly refer to 
these deterministic graphical models, and we explain how the algorithms devoted 
to  their optimisation can be directly applied to compute the mode in a probabilistic graphical model.

In a deterministic
graphical model with only Boolean (0/1) potential functions, each
potential function describes a constraint between variables. If the
potential function takes value 1, the corresponding realisation is
said to satisfy the constraint. If it takes value 0, the realisation does not satisfy it.
The graphical model is known as a
'Constraint Network'. It describes a joint Boolean function on all
variables that takes value 1 if and only if all constraints are
satisfied. The problem of finding a realisation that satisfies all the
constraints, called a solution of the constraint network, is the 'Constraint
Satisfaction Problem' (CSP,~\citealt{HB06}). This framework is used to
model and solve combinatorial optimisation problems. There is a wide
variety of software tools to solve it. 
% When variables are Boolean too,
% and when the Boolean functions are described as disjunctions of
% variables or of their negation, the CSP reduces to the 'Boolean
% Satisfiability' problem (or SAT), the seminal NP-complete
% problem~\citep{Cook71}.

CSP have been extended to describe joint cost functions, decomposed as
a sum of local cost functions, $f_B$ in the `Weighted Constraint
Network' \citep{HB06} or `Cost Function Network'. 
$$
f(\bfx)  = \sum_{B \in \mathcal{B}} f_B(x_B).
$$
In this case,
cost functions take finite or infinite integer or rational
values: infinity enables to express hard constraints while finite 
values encode costs for unsatisfied soft constraints.
The problem of
finding a realisation of minimum cost is the 'Weighted Constraint
Satisfaction Problem' (WCSP), which is  NP-hard. It is easy to
observe that any probabilistic graphical model can be translated in a
weighted constraint network, and vice versa using a simple $-\ln(\cdot)$
transformation.  
$$ 
f_B(x_B) = - \ln(\psi_B), \, \text{with } f_B(x_B) = + \infty \Leftrightarrow  \psi_B(x_B) = 0.
$$
Therefore the WCSP is equivalent to finding a realisation with maximal probability in a probabilistic graphical model.
With this equivalence, it becomes possible to use
exact WCSP resolution algorithms that have been developed in
this field for mode evaluation or for the computation of $Z$, the normalising constant, 
in probabilistic graphical model. See for instance \citet{viricel2016},
for an application on a problem of protein design.

\subsection{Inference tasks in probabilistic graphical models}
\label{subsec: tasks}
Computations on probabilities and potentials rely on two fundamental
types of operations. Firstly, multiplication (or addition in the $\log$ domain)
is used to \emph{combine} potentials to define a joint potential
distribution. Secondly, $ \mathrm{sum} $ or $\max$/$\min$ can be used to
\emph{eliminate} variables and compute marginals or modes of the joint
distribution on subsets of variables. The precise identity of these
two basic operations is not important for the inference algorithms based on variable elimination. 
We therefore adopt a presentation using generic operators to emphasise  this property of the algorithms. 
We denote as $\Atimes$ and as $\Aplus$ the combination
operator and the elimination operator, respectively. To be able to apply the variable elimination algorithm, the only requirement is that $(\mathbb{R}^+,\Aplus,\Atimes)$ defines a commutative
semi-ring. Specifically, the semi-ring algebra offers distributivity:
 $(a \Atimes b) \Aplus (a \Atimes c) = a \Atimes (b\Aplus c)$. For instance, this
corresponds to the distributivity of the product operation over the sum operation, i.e. $(a \times b)
+ (a \times c) = a \times (b + c)$, or to the distributivity of the $\max$ operation over
the sum operation, i.e. $\max(a + b, a + c) = a + \max(b,c) $, or to the
distributivity of the $\max$ operation over the product operation, i.e. $\max(a \times b, a \times
c) = a \times ( \max(b,c)) $. We extend the definition of the two abstract
 operators $\Atimes$ and $\Aplus$ to operators on potential functions, as follows:
\begin{description}
\item[Combine operator:] the combination of two potential
functions $\psi_A$ and $\psi_B$ is a new function $\psi_A\Atimes
\psi_B : \Lambda_{A \cup B} \to \mathbb{R}^+$ defined as
$\psi_A\Atimes \psi_B(x_{A\cup B}) =\psi_A(x_A) \Atimes \psi_B(x_B)$.
\item[Elimination operator:] the elimination of variable
$X_i, i\in B$ from a potential function $\psi_B$ is a new function
$(\Aplus_{x_i} \psi_B): \Lambda_{B\setminus\{i\}} \to \mathbb{R}^+$ defined as $(\Aplus_{x_i}\psi_B)(x_{B \setminus \{i\}}) =
\Aplus_{x_i}(\psi_B(x_{B \setminus \{i\}},x_i))$. For $\Aplus = +$,
$(\Aplus_{x_i} \psi_B)(x_{B \setminus \{i\}})$ represents the marginal sum $\sum_{x_i}
\psi_B(x_{B \setminus \{i\}},x_i)$.
\end{description}
% \end{description}

% \textbf{$\bullet$ conditioning operator}: let us consider a particular
%realisation $x_A$
% of the collection of variables $X_A$, for $A$ a subset of $V$, and $B$ another
% subset of $V$. The result of the conditioning operator applied to the
%potential
% function   $\psi_B$ and $x_A$ is a new function denoted $\psi_B (.
% |x_A)$, from $\Lambda_{B\setminus A}$ to $\mathbb{R}^+$. this new function is
% defined by $\psi_B (x_{B\setminus A} |x_A) = \psi_B(x_{B\setminus A} \ cup
% x_A)$.
%   
% \todo[inline]{ou ca sert?}
% Depending on the task considered, the binary operation $\Atimes$ is
% $\times$ or $+$ (in log-domain), and the $\Aplus$ operation is either
% $+$, $\max$ or $\min$. This means that $(\Aplus_{i} \psi_B)$ is
% respectively obtained by summing, maximising or minimising $\psi_B$
% over $x_i$ values.

Classical counting and optimisation tasks in
graphical models can now be entirely written with these two operators. For simplicity, we
denote by $\Aplus_{x_B}$, where $B\subset V$ a sequence of
eliminations $\Aplus_{x_i}$ for all $i \in B$, the result being
insensitive to the order in a commutative semi-ring. Similarly,
$\Atimes_{B \in \mathcal{B}}$ represents the successive combination of
all potential functions $\psi_B$, with $B \in \mathcal{B}$.  

\paragraph{Counting task.} Under this name we group
all tasks that involve summing over the state space of a subset of
variables in $\bfX$. This includes the computation of the partition
function $Z$ or of any marginal distribution, as well as entropy
evaluation. For $A \subset V$ and $\bar{A} = V \setminus A$, the
marginal distribution $p_A$ of $X_A$ associated to the joint
distribution $p$ is defined as:
 % \begin{equation}
%    \label{eq: marg def}
$$
    p_A(x_A)= \sum_{x_{\bar{A}} \in \Lambda_{\bar{A}}}p\,(x_A,x_{\bar{A}}) =
    \frac{1}{Z} \sum_{ x_{\bar{A}} \in \Lambda_{\bar{A}}} \prod_{B \in \mathcal{B}} \psi_B(x_B)
$$
%  \end{equation}
The function $p_A$ then satisfies ($Z$ is a constant function):
$$p_A\Atimes Z = p_A \Atimes {\Big(\BigAplus_{\bfx} \big(\BigAtimes_{B \in \mathcal{B}} \psi_B\big)\Big)} =  \Big(\BigAplus_{x_{\bar{A}}} \big(\BigAtimes_{B \in \mathcal{B}} \psi_B\big)\Big)$$
where $\Atimes$ combines functions using $\times$ and $\Aplus$ eliminates variables using $+$.

Marginal evaluation is also interesting in the case where some variables are observed.
 If $x_O$ ($O \subset V$) are the values of the observed values, the marginal conditional
distribution can be computed by restricting the domains of variables $X_O$ to the
observed value. This is typically the kind of computational task required in the E-step of an EM algorithm, for parameter estimation of models with hidden data.

% The entropy $H$ of a  probabilistic graphical model $p$ is defined as
%   \begin{equation}
%     \label{eq: entropy def}
%      H(p) = -E[\ln(p(x))],
%  \end{equation}
% where $E[\cdot]$ denotes the mathematical expectation. In the case  of a graphical model, by linearity of the expectation,  the entropy is equal to
%  $$
%      H(p) = \ln(Z) - \sum_{B \in \mathcal{B}} \sum_{x_B \in \Lambda_B}p(x_B) \ln(\psi_B(x_B)).
% $$
%  This expression  is an 
% alternation of use of $\Atimes$ and $\Aplus$  operators (for  $p(x_B)$
% evaluation, for each $B$ and $x_B$).

\paragraph{Optimisation task} The most common optimisation task in a
graphical model corresponds to the evaluation of the most probable
state $\bfx^{*}$ of the random vector $\bfX$, defined as
%\begin{equation}
%  \label{eq: mode def}
$$
\bfx^{*} = \arg\max_{\bfx\in \Lambda} p(\bfx) =
  \arg\max_{\bfx\in\Lambda}\prod_{B\in\mathcal{B}}\psi_B(x_B)
  = \arg\max_{\bfx\in \Lambda}  \sum_{B\in\mathcal{B}} \ln \psi_B(x_B)
$$
%\end{equation}
The maximum itself is $\BigAplus_{\bfx}( \BigAtimes_{B \in \mathcal{B}}
\ln{\psi_B(x_B)})$ with $\Aplus$ and $\Atimes$ set to $\max$ and to $+$, respectively.
The computation of the mode $\bfx^*$ does not require the computation of the
normalising constant $Z$, however evaluating the mode probability value $p(\bfx^*)$ does.
Another optimisation task of interest is the computation of the max-marginals of each variable $X_i$  defined as $p^*(x_i) = \max_{x_{V\setminus i}} p(\bfx)$.

Therefore counting and optimisation tasks can be interpreted as two
instantiations of the same computational task expressed in terms of
combination and elimination operators, namely $\BigAplus_{x_A}
\BigAtimes_{B \in \mathcal{B}} \psi_B$, where $A \subseteq V$.  When the
combination operator $\Atimes$ and the elimination operator $\Aplus$
are set to $\times$ and $+$, respectively, this computational problem
is known as a sum-product problem in the Artificial Intelligence
literature~\citep{Pearl88},\cite[chapter 8]{B06}. When $\Aplus$ and $\Atimes$ are
set to $\max$ and to the sum operator, respectively it is a max-sum
problem~\cite[chapter 8]{B06}. 
In practice, it means that tasks such as solving the E-step of the EM algorithm or computing the mode in a graphical model, belong to the same family of  computational problems.

We will see in Section~\ref{sec: exactinference} that there exists an
exact algorithm solving this general task which exploits the
distributivity of the combination and elimination operators to perform
operations in a smart order. From this generic algorithm, known as
variable elimination~\citep{Bertele72} or bucket
elimination~\citep{dechter99}, one can deduce exact algorithms to
solve counting and optimisation tasks in a graphical model,
by instantiating the operators $\Aplus$ and $\Atimes$.

\subparagraph{Deterministic Graphical models.} The Constraint
Satisfaction Problem is a $\lor$-$\land$ problem as it can can be
defined using $\lor$ (logical 'or') as the elimination operator and
$\land$ (logical 'and') as the combination operator over Booleans. The weighted CSP is
a $\min$-$+$ as it uses $\min$ as the elimination operator and $+$ (or
bounded variants of $+$) as the combination operator. Several other
variants exist~\citep{HB06}, including generic algebraic
variants~\citep{Schiex95a,Bistarelli97,Cooper04,pralet2007,kohlas2003}.

% \centerline{TABLE \ref{tab: GM tasks} ABOUT HERE}
\begin{table}[h!]
\caption{Definitions of the Combine ($\Atimes$) and the Elimination ($\Aplus$) operators for classical tasks on probabilistic and deterministic graphical models.\label{tab: GM tasks}}
\begin{center}
\begin{tabular}{lcc}
\hline
Task & $\Aplus $& $\Atimes$  \\
\hline
Marginal evaluation &   $+$ &  $\times$ \\
Mode evaluation & $\max$ & $+$ \\
Existence of a solution in  a CSP & $\lor$ & $\land$  \\
Evaluation of the  minimum cost in WCSP &  $\min$ & $+$  \\
\hline
\end{tabular}
\end{center}
\end{table}

 \subsection{Example: Coupled HMM}
 \label{sec : def CHMM}
% \textcolor{red}{section a construire. decrire l'exemple d'epidemio avec %observations bruitees qui servira de fil rouge. expliquer qu'on a besoin de %sum-prod dans l'etape E du EM ...}
 
 We introduce now the example of Coupled Hidden Markov Models (CHMM), which can be seen as extensions Hidden Markov Chain (HMC) models to several chains in interactions. In section \ref{sec: illust CHMM} we will use this framework to illustrate the behaviour of exact and approximate algorithms based on variable elimination.
 
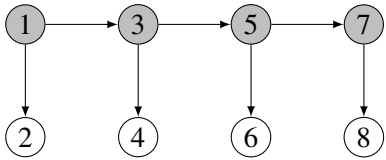
\begin{figure}
\begin{center}
\begin{tikzpicture}[scale=1.5]
\tikzstyle{hiddennode}=[draw,circle,fill=gray!50,minimum size = 5pt, inner sep = 2pt,font=\small]
\tikzstyle{obsnode}=[draw,circle,minimum size = 5pt, inner sep = 2pt,font=\small]
  \tikzstyle{arc}=[->,>=latex]
  \path (0,0) node[hiddennode] (a) {1}
        (0,-1) node[obsnode] (b) {2}
        (1,0) node[hiddennode] (c) {3}
        (1,-1) node[obsnode] (d) {4}
        (2,0) node[hiddennode] (e) {5}
        (2,-1) node[obsnode] (f) {6}
        (3,0) node[hiddennode] (g) {7}
        (3,-1) node[obsnode] (h) {8};
  \draw[arc] (a) to (b);
  \draw[arc] (a) to (c);
  \draw[arc] (c) to (d);
  \draw[arc] (c) to (e);
  \draw[arc] (e) to (f);
  \draw[arc] (e) to (g);
  \draw[arc] (g) to (h);
\end{tikzpicture}
\end{center}
\caption{Graphical representation of a HMM. Hidden variables correspond to vertices 1, 3, 5, 7, and  observed variables to vertices 2, 4, 6, 8.}
\label{fig:HMM}
\end{figure}

 A HMC  (Figure \ref{fig:HMM}) is defined by  two sequences of random variables $\bfO$
and $\bfH$ of same length, $T$.  A realisation $\bfo = (o_1, \ldots
o_T)^\top$ of the variables $\bfO =(O_1, \ldots O_T)^\top$ is observed, while the
states of variables $\bfH = (H_1, \ldots H_T)^\top$ are unknown (hidden).  In the HMC
model the assumption is made that $O_i$ is independent of $H_{V \setminus \{i\}}$  and $O_{V \setminus \{i\}}$ 
given the hidden
variable $H_i$. These independences
are modelled by pairwise potential functions $\psi_{H_i,O_i}, \forall \
1 \leq i\leq T$. Furthermore, hidden variable $H_i$ is independent of $H_1, \ldots, H_{i-2}$ 
 and $O_1, \ldots, O_{i-1}$ given the hidden variable $H_{i-1}$. 
These independences are modelled
by pairwise potential functions $\psi_{H_{i-1},H_i}, \forall \ 1<
i\leq T$. Then the model is fully defined by specifying an additional
 potential function $\psi_{H_1}(h_1)$ to model the initial distribution.  In the classical HMC
formulation \citep{rabiner1989}, these potential functions are
normalised conditional probability distributions i.e.,
$\psi_{H_{i-1},H_i}(h_{i-1}, h_i) = \Pr(H_i = h_i|H_{i-1} = h_{i-1})$,
$\psi_{O_{i},H_i}(o_i,h_i) = \Pr(O_i = o_i|H_i=h_i)$ and $\psi_{H_1}(h_1) =
\Pr(H_1=h_1)$.  As a consequence, the normalising constant $Z$ is equal to
$1$, as it is in Bayesian networks.

% \centerline{FIGURE \ref{fig:HMM} ABOUT HERE}\begin{figure}[h!]

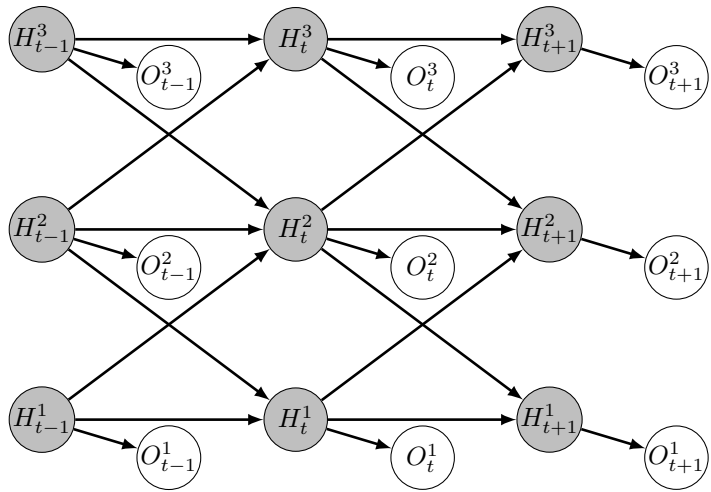
\begin{figure}[h!]
  \begin{center}
  \begin{tikzpicture}
 % \node[empty] (E1tm2) at (-1*\heu, 0) {\footnotesize $\quad$};
  \node[hidden] (H1tm1) at (0, 0) {\footnotesize $H^{1}_{t-1}$};
  \node[hidden] (H1t) at (\heu, 0) {\footnotesize $H^{1}_{t}$};
  \node[hidden] (H1tp1) at (2*\heu, 0) {\footnotesize $H^{1}_{t+1}$};
 % \node[empty] (E1tp2) at (3*\heu, 0) {\footnotesize $\quad$};

 % \node[empty] (Eitm2) at (-1*\heu, \veu) {\footnotesize $\quad$};
  \node[hidden] (Hitm1) at (0, \veu) {\footnotesize $H^{2}_{t-1}$};
  \node[hidden] (Hit) at (\heu, \veu) {\footnotesize $H^{2}_{t}$};
  \node[hidden] (Hitp1) at (2*\heu, \veu) {\footnotesize $H^{2}_{t+1}$};
 % \node[empty] (Eitp2) at (3*\heu, \veu) {\footnotesize $\quad$};

 % \node[empty] (EItm2) at (-1*\heu, 2*\veu) {\footnotesize $\quad$};
  \node[hidden] (HItm1) at (0, 2*\veu) {\footnotesize $H^{3}_{t-1}$};
  \node[hidden] (HIt) at (\heu, 2*\veu) {\footnotesize $H^{3}_{t}$};
  \node[hidden] (HItp1) at (2*\heu, 2*\veu) {\footnotesize $H^{3}_{t+1}$};
 % \node[empty] (EItp2) at (3*\heu, 2*\veu) {\footnotesize $\quad$};
  
  %\draw[arrow] (E1tm2) to (H1tm1);  \draw[arrow] (H1tp1) to (E1tp2);   
  \draw[arrow] (H1tm1) to (H1t);   \draw[arrow] (H1t) to (H1tp1);  
  %\draw[arrow] (Eitm2) to (Hitm1);  \draw[arrow] (Hitp1) to (Eitp2);   
  \draw[arrow] (Hitm1) to (Hit);   \draw[arrow] (Hit) to (Hitp1);  
  %\draw[arrow] (EItm2) to (HItm1);  \draw[arrow] (HItp1) to (EItp2);   
  \draw[arrow] (HItm1) to (HIt);   \draw[arrow] (HIt) to (HItp1);  
  
  \draw[arrow] (H1tm1) to (Hit);   
  %\draw[edge] (H1tm1) to [bend left](HItm1);  
  \draw[arrow] (Hitm1) to (HIt);
  \draw[arrow] (Hitm1) to (H1t);
  \draw[arrow] (HItm1) to (Hit); 
  
   \draw[arrow] (H1t) to (Hitp1);   
  %\draw[edge] (H1tm1) to [bend left](HItm1);  
  \draw[arrow] (Hit) to (HItp1);
  \draw[arrow] (Hit) to (H1tp1);
  \draw[arrow] (HIt) to (Hitp1); 
  
%   \draw[edge] (H1t) to [bend left](HIt);  
%   \draw[edge] (Hit) to (HIt);  
%   \draw[edge] (H1tp1) to (Hitp1);   
%   \draw[edge] (H1tp1) to [bend left](HItp1);  
%   \draw[edge] (Hitp1) to (HItp1);  
  
  \node[observed] (O1tm1) at (.5*\heu, -.2*\veu) {\footnotesize $O^{1}_{t-1}$};
  \node[observed] (O1t) at (1.5*\heu, -.2*\veu) {\footnotesize $O^{1}_{t}$};
  \node[observed] (O1tp1) at (2.5*\heu, -.2*\veu) {\footnotesize $O^{1}_{t+1}$};
  \node[observed] (Oitm1) at (.5*\heu, .8*\veu) {\footnotesize $O^{2}_{t-1}$};
  \node[observed] (Oit) at (1.5*\heu, .8*\veu) {\footnotesize $O^{2}_{t}$};
  \node[observed] (Oitp1) at (2.5*\heu, .8*\veu) {\footnotesize $O^{2}_{t+1}$};
  \node[observed] (OItm1) at (.5*\heu, 1.8*\veu) {\footnotesize $O^{3}_{t-1}$};
  \node[observed] (OIt) at (1.5*\heu, 1.8*\veu) {\footnotesize $O^{3}_{t}$};
  \node[observed] (OItp1) at (2.5*\heu, 1.8*\veu) {\footnotesize $O^{3}_{t+1}$};
  
  \draw[arrow] (H1tm1) to (O1tm1);   
  \draw[arrow] (H1t) to (O1t);
  \draw[arrow] (H1tp1) to (O1tp1);   
  \draw[arrow] (Hitm1) to (Oitm1);   
  \draw[arrow] (Hit) to (Oit); 
  \draw[arrow] (Hitp1) to (Oitp1);  
  \draw[arrow] (HItm1) to (OItm1);  
  \draw[arrow] (HIt) to (OIt);
  \draw[arrow] (HItp1) to (OItp1);  

  \end{tikzpicture}
  \caption{Graphical representation of a coupled HMM with 3 hidden chains.\label{Fig:CoupledHMM}}
 \end{center}
\end{figure}

 Consider now that there is more than one hidden chain:  $I$ signals are observed at times $t \in \{1, \dots T\}$ and we denote $O^i_t$ the variable corresponding to the observed signal
 $i$ at time $t$. Variable $O^i_t$ depends on some hidden state $H^i_t$.
 %, where the series $(H^i_t)_{t = 1, \dots T}$ is a Markov chain. 
% If the series are supposed to be independent, the corresponding graphical model corresponds to Figure~\ref{Fig:CoupledHMM-Markov}. 
The Coupled HMM (CHMM) framework  assumes dependency between  two hidden chains at two consecutive time steps  (see \citealt{Brand97}): $H_t^i$ depends not only of $H_{t-1}^i$, it may depend on some $H_{t-1}^j$ for $j \neq i$. The set of the indices of chains  upon which $H_t^i$ depends (expect $i$) is noted $L_i$.
This results in the graphical structure displayed on \figurename~\ref{Fig:CoupledHMM}, where  $L_2 = \{1,3\}$ and $L_1=L_3 = \{2\}$. Such models have been considered in a series of domains such
 as bioinformatics \citep{CFN13}, electroencephalogram  analysis \citep{ZhG02} or speech recognition \citep{NoO03}. 
% (see \figurename~\ref{Fig:CoupledHMM-DBN}). 
In a CHMM setting,  the joint distribution of the hidden variables $\bfH = (H^i_t)_{i, t}$ and observed variables $\bfO = (O^i_t)_{i, t}$ factorises as
\begin{equation} \label{Eq:cHMM-joint}
\Pr(\bfh, \bfo) 
\propto \prod_{i=1}^I \psi^{init}(h_1^i)\left(\prod_{i=1}^I \prod_{t=2}^T \psi^M(h^i_{t-1},  h_{t-1}^{L_i}, h^i_t)\right) 
\times \left(\prod_{i=1}^I \prod_{t=1}^T \psi^E(h^i_t, o^i_t)\right),
\end{equation}
% where $h_t = (h^i_t)_i$ stands the vector of all hidden variables at time $t$ and where
% $\psi^M$ encodes the Markovian dependency of the hidden variables within a series, 
% $\psi^C$ encodes the coupling between the hidden variables of all series at a given time and  
where $\psi^{init}$ is the initial distribution,  $\psi^M$ encodes the local transition function of $H_t^i$ 
and $\psi^E$ encodes the emission of the observed signal given the corresponding hidden state.
%When dealing with such a model, one is typically interested in the conditional distribution of the hidden states given the observed signal, namely $p(h|o)$ or in the mode of this distribution, namely $\widehat{h} = \arg\max_h p(h|o)$.
A fairly comprehensive  exploration of these models can be found in \citep{Mur02}.

% \centerline{FIGURE \ref{Fig:CoupledHMM} ABOUT HERE}

Potential function $\psi^{init}$, $\psi^M$ and $\psi^E$ can be parameterised by a set of parameters denoted $\theta$. A classical problem for CHMM is have more than one iron in the fire: (a) estimate $\theta$ and (b) compute the mode of the conditional distribution of the hidden variables given the observations. Estimation can be performed using an EM algorithm, and as mentioned previously, the E-step of the algorithm and the mode computation task belong to the same family of computational task in graphical models. Both can be solved using variable elimination, as we show in the next section.

Beforehand, we present a reasonably simple example of CHMM that will be used to illustrate the different inference algorithms introduced in this work.
It models the dynamics of a pest that can spread on a landscape  composed of $I$ crop fields organised on a regular grid.
The spatial neighbourhood of field $i$, denoted $L_i$, is the set of the four closest fields (three on the borders, and two in corners of the grid).
$H_t^i \in \{0,1\}$ ($1 \leq i \leq I, \ 1 \leq t \leq T$) is the state of crop field $i$ at time $t$. 
State 0  (resp. 1) represents the absence (resp. presence) of the pest in the field.
Variable  $H_t^i$ depends on  $H_{t-1}^i$ and of the  $H_{t-1}^j$, for $j \in L_i$.
The conditional probabilities of survival and apparition of the pest in field $i$ are parameterised by 3 parameters: 
$\epsilon$, the probability  of contamination from outside the landscape (long-distance dispersal);
$\rho$, the probability that the pest spreads from an infected field $j \in L_i$ to  field $i$ between two consecutive times; and
  $\nu$, the probability of field persistent infection between two consecutive times.
  %Note here that these parameters do not depend on time $ t $, so they are steady-state %limiting parameters of the dynamics.
 We assume that contamination events from all neighbouring fields are independent.
Then, if ${C_t^i}$ is the number of contaminated neighbours of field $i$ at time $t$ (i.e. $C_t^i = \sum_{j \in L_i} H_t^j$), the contamination potential of field $ i $ at time $ t $ writes:
$$
\psi^M(0, h_{t-1}^{L_i}, 1) = \Pr(H_{t}^i = 1 \mid H_{t-1}^i = 0, h_{t-1}^{j}, j \in L_i) = \epsilon + (1 - \epsilon)(1 - (1- \rho)^{C_t^i} ),
$$
and its persistence in a contaminated state writes:
\begin{eqnarray}
\psi^M(1, h_{t-1}^{L_i}, 1)& = &\Pr(H_{t}^i = 1 \mid h_{t-1}^i = 1, h_t^{j}, j \in L_i) \nonumber \\
 &= & \nu + (1 - \nu) \left(  \epsilon + (1 - \epsilon)(1 - (1- \rho)^{C_t^i} ) \right). \nonumber
\end{eqnarray}

The $(H_t^i)$'s are hidden variables but
monitoring observations are available. A binary variable $ O_t^i $ is observed: it takes value 1, if the pest was declared as present in the field, and 0 otherwise. 
Errors of detection are possible. False negative observations occur since even if the pest is there, it can be difficult to notice, and  missed. On the opposite, false positive observations occur when the pest is mixed up  with another one.
We define the corresponding emission potential as $\psi^E(0,1)  = \Pr(O_t^i = 0 \mid H_t^i = 1) = f_n$ and $\psi^E(1,0)  = \Pr(O_t^i = 1 \mid H_t^i = 0) = f_p$, respectively.

\section{Variable elimination for  exact inference}
%Methods for exact inference 
\label{sec: exactinference}
% Une serie d'algo partant de cas connus (Viterbi, programmation
% dynamique serielle) progressivement complexifies (elimination de
% variable PD non serielle, elimination de variable, ...). La
% complexite des algos depend de l'ordre dans lequel les operations
% d'elimination se font , la largeur d'arbre apparait comme quantite
% naturelle d'interet.

We describe now the principle of variable elimination to solve the general inference tasks presented in Section \ref{subsec: tasks}. We first recall
the Viterbi algorithm for Hidden Markov Chains \citep{rabiner1989},
a classical example of
variable elimination for optimisation (mode evaluation).  Then, we  formally  describe the
variable elimination procedure in the general graphical model framework. The key
element is the choice of an ordering for the sequential elimination of the variables. 
It is closely linked to the notion of treewidth of the
graphical representation of the model. We explain how the complexity of
 a variable elimination algorithm  is  fully characterised by this notion. We also describe the extension to the elimination of blocks of variables.

\subsection{Case of hidden Markov chain models}
As a didactic introduction to exact inference on graphical models by variable
elimination, we consider a well studied stochastic process: the
discrete Hidden Markov Chain model (HMC).

% This is not required if the
% HMM is considered as a specific type of Markov Random Field since $Z$
% will take care of normalisation.
% Assume that the parameters of the HMM, that is the values of the
% potential functions over their domain, are known. A usual
% inference task over HMM is to identify the most probable value of the
% variables in $H$ given a realisation $o$ of the variables in $O$. That
% is, we want to compute $\arg\max_H p(H| O=o)$ or equivalently
% $\arg\max_H \prod_{\psi_B\in \psi} \psi_B(B|o|)$. 
% 
% This is the result of $\max$-eliminating all variables in $H$ from the
% $\times$-join of all the potential functions conditionned by $o$. 
%Let us consider that the parameters of the HMC model are known (that is the
%value of the potential functions over their domain). 

A classical inference task for HMC is to identify the most likely
values of variables $\bfH$ given a realisation $\bfo$ of the variables
$\bfO$. The problem is to compute $\arg\max_\bfh \Pr(\bfH = \bfh| \bfO=\bfo)$, or
equivalently the argument of:
\begin{equation}\label{viterbi}
\max_{h_1, \ldots, h_T} \Big[(\psi_{H_1}(h_1)\psi_{O_{1},H_1}(o_1, h_1))\prod_{i = 2}^T (\psi_{H_{i-1},H_i}(h_{i-1}, h_i)\psi_{O_{i},H_i}(o_i,h_i))\Big]
\end{equation}
%
%This is a conditional optimisation task as defined in the previous
%section.  The most probable state is obtained by applying the
%elimination operator $\Aplus$ (here $\max$) to the combined product
%($\Atimes$) of all potential functions.
The number of possible realisations of $\bfH$ is exponential in $T$.
Nevertheless, this optimisation problem can be solved in a number of
operations linear in $T$ using the well-known Viterbi
algorithm~\citep{rabiner1989}. This algorithm, based on dynamic
programming, performs successive eliminations (by maximisation) of all
hidden variables, starting with $H_T$, and iteratively considering the $ H_i $'s for $ i=T-1, T-2,\ldots $, and finishing by
$H_1$. It successively computes the most likely sequence of hidden variables. By
using distributivity between the $\max$ and the product operators, the
elimination of variable $H_T$ can be done by rewriting (\ref{viterbi})
as:

% $$\max_{h_1,\ldots,h_{T-1}} \Big[\psi_{H_1}(h_1)\psi_{O_{1},H_1}(o_1,h_1) \prod_{i = 2}^{T-1} (\psi_{H_{i-1},H_i}(h_{i-1}, h_i)\psi_{O_{i},H_i}(o_i,h_i).
% \underbrace{\max_{h_T}\psi_{H_{T-1},H_T}(h_{T-1}, h_T) \psi_{O_{T},H_T}(o_T,h_T)}_{\textrm{\scriptsize New potential function}}\Big]$$

\begin{eqnarray}
 && \max_{h_1,\ldots,h_{T-1}} \Big[\psi_{H_1}(h_1)\psi_{O_{1},H_1}(o_1,h_1)     \nonumber \\
&& 
  \prod_{i = 2}^{T-1}   \big (\psi_{H_{i-1},H_i}(h_{i-1}, h_i)\psi_{O_{i},H_i}(o_i,h_i)
\underbrace{\max_{h_T}\psi_{H_{T-1},H_T}(h_{T-1}, h_T) \psi_{O_{T},H_T}(o_T,h_T)}_{\textrm{\scriptsize New potential function}} \big)\Big] \nonumber
\end{eqnarray}

The new potential function created by maximising on $H_T$ depends only
on variable $H_{T-1}$. 
% Indeed, variables $H_T,O_T$ and the potential
% functions involving them are removed from the optimisation
% problem. 
The same principle can then be applied to $ H_{T-1} $ and so forth. This is a simple application of the general variable
elimination algorithm that we describe in the next section.

\subsection{General principle of variable elimination}
% We show now that Viterbe application of a much more
% general algorithmic tool called variable elimination~\cite{Bertele72}
% (also known as bucket elimination~\cite{dechter99}) and which can be
% described as a graphical model transformation technique. 

In Section \ref{sec: GM}, we have seen that counting and optimisation
tasks can be formalised by the same generic algebraic formulation
\begin{equation}\label{elimeq}
\BigAplus_{x_A} (\BigAtimes_{B \in \mathcal{B}} \psi_B)
\end{equation}
where $A\subseteq V$. 

The trick behind variable elimination~\citep{Bertele72} relies on a
clever use of the distributivity property.  Indeed, evaluating $(a
\Atimes b) \Aplus (a \Atimes c)$ as $a \Atimes (b\Aplus c)$ requires
fewer operations. Hence eliminating $ a $ in the second writing leads 
to dealing with fewer algebraic operations. Since distributivity applies 
both for counting and optimising tasks, variable elimination 
can be applied to both tasks.
It also means that if variable elimination is efficient for one
task it will also be efficient for the other one.  As in the HMC
example, the principle of the variable elimination algorithm for
counting or optimising consists in eliminating variables one by one in
an expression of the problem like in (\ref{elimeq}).
 
The elimination of the first variable, say $X_i, i \in A$, is
performed by merging all potential functions involving $X_i$ and
applying operator $\Aplus_{x_i}$ to these potential functions.  Using
commutativity and associativity of both operators, (\ref{elimeq}) can be rewritten as:
$$
 \BigAplus_{x_A} (\BigAtimes_{B \in \mathcal{B}} \psi_B ) = 
\BigAplus_{x_{A \setminus \{i\}}}  \BigAplus_{x_i} \left( (\BigAtimes_{B \in \mathcal{B} \setminus \mathcal{B}_i} \psi_B ) \BigAtimes(\BigAtimes_{B \in \mathcal{B}_i} \psi_B) \right),
%\BigAplus_{x_{A \setminus \{i\}}}  \BigAplus_{x_i} (\BigAtimes_{B \in \mathcal{B} \setminus \mathcal{B}_i} \BigAtimes_{B \in \mathcal{B}_i} \psi_B )
 $$
\noindent where $ \mathcal{B}_i $ is the subset of $V$ defined such as all its elements contain $ i $. Then using distributivity of $\Atimes$ on $\Aplus$, we obtain:
$$ \BigAplus_{x_A} (\BigAtimes_{B \in \mathcal{B}} \psi_B ) =  \BigAplus_{x_{A \setminus \{i\}}} \Big[(\BigAtimes_{B \in \mathcal{B} \setminus \mathcal{B}_i} \psi_B) \Atimes \underbrace{(\BigAplus_{x_i} \BigAtimes_{B \in \mathcal{B}_i} \psi_B)}_{\textrm{\scriptsize New potential function $\psi_{N_i}$}}\Big]$$

%This shows that to continue the elimination of all variables $X_A$, the next step can be reduced
%to the elimination of $X_{A \setminus \{i\}}$
This shows that the elimination of $X_i$ results in a new graphical
model, where variable $X_i$ and the potential functions $\psi_B,
B\in\mathcal{B}_i = \{ B', \: x_i \in B' \} $ do not appear anymore. They are replaced by a new potential
$\psi_{N_i}$ which does not involve $X_i$, but depends on its neighbours in $G$.  The graph associated to
the new graphical model is in a sense similar to the one of the original model. It is updated as follows: vertex $X_i$ is removed, and  neighbours $X_{N_i}$
of $X_i$ are now connected together in a clique because they are all in the scope of $\psi_{N_i}$. The new edges between
the neighbours of $X_i$ are called \emph{fill-in} edges.  For instance,
when eliminating variable $X_1$ in the graph of Figure~\ref{fig: one
  step elim} (left), potential functions $\psi_{1,2}, \ \psi_{1,3}, \
\psi_{1,4}$ and $\psi_{1,5}$ are replaced by $\psi_{2,3,4,5} =
\Aplus_{x_1} (\psi_{1,2} \Atimes \psi_{1,3} \Atimes \psi_{1,4} \Atimes
\psi_{1,4}) $.  The new graph is shown in Figure~\ref{fig: one step
  elim} (right).\\
  
% \centerline{FIGURE \ref{fig: one step elim} ABOUT HERE}
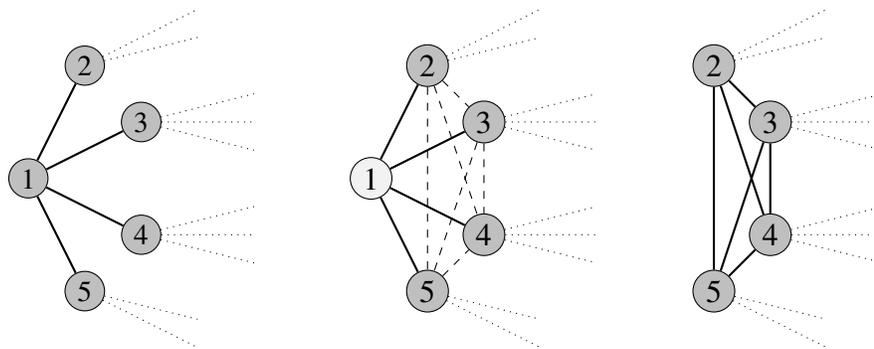
\begin{figure}[h!]
\centering
 \begin{tikzpicture}[scale=0.1875]
    \tikzstyle{vertexlight}=[draw,circle,fill=gray!50,minimum size = 5pt,inner sep=2pt, font=\small]
  \path (0,0) node[vertexlight] (a) {1}
        (4,8) node[vertexlight] (b) {2}
    (8,4) node[vertexlight] (c) {3}
    (4,-8)node[vertexlight] (d) {5}
    (8,-4) node[vertexlight] (e) {4};
  \draw[thick] (a) -- (b) ;
  \draw[thick] (a)--(c);
  \draw[thick] (a)--(d);
  \draw[thick] (a)--(e);
  \draw[dotted] (b)--+(8,4);
  \draw[dotted] (b)--+(8,2);
  \draw[dotted] (c)--+(8,2);
  \draw[dotted] (c)--+(8,0);
  \draw[dotted] (c)--+(8,-2);
  \draw[dotted] (e)--+(8,2);
  \draw[dotted] (e)--+(8,0);
  \draw[dotted] (e)--+(8,-2);
  \draw[dotted] (d)--+(8,-4);
  \draw[dotted] (d)--+(8,-2);
 \end{tikzpicture}\quad\quad\quad
\begin{tikzpicture}[scale=0.1875]
    \tikzstyle{vertexlight}=[draw,circle,fill=gray!50,minimum size = 5pt,inner sep=2pt]
    \tikzstyle{vertexdark}=[draw,circle,fill=gray!10,minimum size = 5pt,inner sep=2pt]
  \path (0,0) node[vertexdark] (a) {1}
        (4,8) node[vertexlight] (b) {2}
    (8,4) node[vertexlight] (c) {3}
    (4,-8)node[vertexlight] (d) {5}
    (8,-4) node[vertexlight] (e) {4};
  \draw[thick] (a) -- (b) ;
  \draw[thick] (a)--(c);
  \draw[thick] (a)--(d);
  \draw[thick] (a)--(e);
  \draw[dashed] (b)--(c);
  \draw[dashed] (b)--(d);
  \draw[dashed] (b)--(e);
  \draw[dashed] (c)--(d);
  \draw[dashed] (c)--(e);
  \draw[dashed] (d)--(e);
  \draw[dotted] (b)--+(8,4);
  \draw[dotted] (b)--+(8,2);
  \draw[dotted] (c)--+(8,2);
  \draw[dotted] (c)--+(8,0);
  \draw[dotted] (c)--+(8,-2);
  \draw[dotted] (e)--+(8,2);
  \draw[dotted] (e)--+(8,0);
  \draw[dotted] (e)--+(8,-2);
  \draw[dotted] (d)--+(8,-4);
  \draw[dotted] (d)--+(8,-2);

 \end{tikzpicture}\quad\quad\quad
\begin{tikzpicture}[scale=0.1875]
    \tikzstyle{vertexlight}=[draw,circle,fill=gray!50,minimum size = 5pt,inner sep=2pt]
  \path (4,8) node[vertexlight] (b) {2}
    (8,4) node[vertexlight] (c) {3}
    (4,-8)node[vertexlight] (d) {5}
    (8,-4) node[vertexlight] (e) {4};
    \draw[thick] (b)--(c);
  \draw[thick] (b)--(d);
  \draw[thick] (b)--(e);
  \draw[thick] (c)--(d);
  \draw[thick] (c)--(e);
  \draw[thick] (d)--(e);
  \draw[dotted] (b)--+(8,4);
  \draw[dotted] (b)--+(8,2);
  \draw[dotted] (c)--+(8,2);
  \draw[dotted] (c)--+(8,0);
  \draw[dotted] (c)--+(8,-2);
  \draw[dotted] (e)--+(8,2);
  \draw[dotted] (e)--+(8,0);
  \draw[dotted] (e)--+(8,-2);
  \draw[dotted] (d)--+(8,-4);
  \draw[dotted] (d)--+(8,-2);

 \end{tikzpicture}\quad\quad\quad
 \caption{Elimination of variable $X_1$ replaces the four pairwise
   potential functions involving variable $X_1$ with a new potential
   $\psi_{N_1}$, involving the four neighbours of vertex 1 in the
   original graph. The new edges created between these four vertices
   are called fill-in edges (dashed edges in the middle figure).}
\label{fig: one step elim}
\end{figure}

% This
%new potential function $\psi_{N_i}$ can be interpreted as a message
%that abstracts the influence of all the potentials involving $X_i$ on
%the rest of the system. 
\paragraph{Interpretation for marginalisation, maximisation and finding the mode of a distribution} When the first elimination step is applied with $\Aplus = +$ and $\Atimes = \times$, 
the probability distribution defined by this
new graphical model is the marginal distribution $p_{V \setminus \{i\}}(x_{V \setminus \{i\}})$ of the original distribution $p$ (up to a constant).
The complete elimination can be obtained by successively eliminating
all variables in $X_A$. 
The result is a graphical model over $X_{V \setminus A}$,
which specifies the marginal distribution $p_{V \setminus A}(x_{V \setminus A})$. When $A=V$, 
the result is a model with a single constant potential
function with value $Z$.

If instead $\Aplus$ is $\max$, and $\Atimes =  \times$ (or $+$ with a log
transformation of the potential functions) and $A=V$, the last potential
function obtained after elimination of the last variable is  equal to the maximum of
the non-normalised distribution. So evaluating  $ Z $ or the maximal probability of a
graphical model can be both obtained with the same variable elimination algorithm, just changing the definition of the $\Aplus$ (and $\Atimes$ if needed) operator(s). \\
Lastly,  if one is interested in the mode itself, an additional 
computation is required. The mode is actually obtained  by induction: if $x^*_{V \setminus
\{i\}}$ is the mode of the graphical model obtained after the elimination of the
first variable,  $X_i$, then the mode of $p$ can be defined as $(x^*_{V
\setminus \{i\}}, x^*_i)$, where $x^*_i$ is a value  in $\Lambda_i$ that
maximises $\Atimes_{B\in\mathcal{B}}\psi_B(x^*_{V \setminus \{i\}},x_i)$. 
This maximisation  is straightforward to derive because $x_i$ can take
only $|\Lambda_i|$ values. $x^*_{V \setminus
\{i\}}$ itself is obtained by completing the mode of the graphical model obtained after elimination of the second variable, and so on.
 We stress here that the procedure requires to keep the intermediary potential functions $\psi_{N_i}$ created during the successive eliminations.

\paragraph{Complexity of the intermediary potential functions and variable elimination ordering: a prelude to the treewidth} When eliminating a variable $X_i$, the task which can be computationally 
expensive is the computation of the intermediate $\psi_{N_i}$.  It
requires to compute the product $\Atimes_{B \in \mathcal{B}_i}
\psi_B(x_B)$ of several potential functions for all elements of
$\Lambda_{N_i\cup\{i\}}$, the state space of $X_{N_i\cup\{i\}}$.
  The time and space complexity of the
operation are entirely determined by the cardinality $|N_i|$ of the set of indices in $N_i$. 
If $K = \max_{j
  \in V} |\Lambda_j|$, the time complexity (i.e. number of elementary operations performed)
is in $O(K^{|N_i|+1})$ and space complexity (i.e. memory space needed) is in $O(K^{|N_i|})$.
Complexity is therefore exponential in $|N_i|$, the number of
neighbours of the eliminated variable in the current graphical
model. The total complexity of the variable elimination is then exponential in the maximum cardinality
$|N_i|$ over all successive eliminations. However note that it is linear in $n$, which means that a large $n$ is not necessarily a problem for having access to exact inference.  Because the graphical model
changes at each elimination  step, this number usually depends on the order
in which variables are eliminated.

As a consequence, the prerequisite to apply variable elimination is to
decide for an ordering of the elimination of the variables. As illustrated in Figure~\ref{fig:
  good and bad order} two different orders can lead to two different $N_i$ subsets. The key message is that the choice of the order
is crucial. It dictates the efficiency of the variable elimination procedure. 
  We now illustrate and formalise this intuition.

\subsection{When is variable elimination efficient ?}
\label{var-elim-eff}

We can understand why the Viterbi algorithm is an efficient algorithm
for mode evaluation in a HMC.  The graph associated to a HMC is
comb-shaped: the hidden variables form a line and each observed
variable is a leaf in the comb (see Figure \ref{fig:HMM}).  So it is possible to design an
elimination order where the current variable to eliminate has a unique neighbour
in the graphical representation of the current model: 
%for instance $O_T > H_T > O_{T-1}> H_{T-1}, \ldots > O_1 > H_1$ 
for instance $H_T > H_{T-1}, \ldots  > H_1$. By convention, the
first eliminated variable is the largest according to this ordering
(note that variables $O_t$ do not have to be eliminated since their value is known).
Following this elimination order, when eliminating a variable using
$\Aplus$, the resulting graphical model has one fewer vertex than the
previous one and \emph{no fill-in edge}. Indeed, the new potential
function $\psi_{N_i}$ is a  function of a single variable since $|N_i| = 1$. 
The Viterbi algorithm as a space complexity of $O(TK)$ and a time complexity of $O(TK^2)$.
%deja dit
% elimination algorithm, if we have kept all the intermediate
% $\psi_{N_i}$, it is easy to identify the mode by working in reverse
% order and always choosing the value of $x_i$ that maximises the
% intermediary graphical model. 
% This task is easy because $x_i$ can take
% only $|\Lambda_i|$ values.

More generally, variable elimination is very efficient, i.e. leads 
to transitional $N_i$ sets of small cardinality, on graphical
models whose graph representation is a  tree. More specifically, for such graph structure, it is
always possible to design an elimination order where the current
variable to eliminate has only one neighbour in the graphical representation
of the current model.

Another situation where variable elimination can be efficient is when the graph
associated to
the graphical model is \emph{chordal} (any cycle of length four or more has a chord i.e.,
an edge connecting
two non adjacent vertices in the cycle), and when the size  of the largest clique
is low. The rationale for this interesting property is explained intuitively here. In Figure \ref{fig: one step elim}, new edges are created
between neighbours of the eliminated vertex. If this neighbourhood is a clique, no
new edge is added. A vertex whose neighbourhood  is a clique is called a
\emph{simplicial vertex}. Chordal graphs have the property that there exists an
elimination order of the vertices, such that every vertex during the elimination
process is simplicial \citep{HABIB2009}. Consequently, there exists an elimination order such that no
fill-in edges are created. Thus, the size of a transitional $N_i$'s is dictated by 
the size of  the clique formed by the neighbours of $i$
%, hence bounded by the size of the largest clique in the graph. \textcolor{red}{NAIVE QUESTION: HOW CAN IT BE LESS? I WOULD REMOVE THIS SENTENCE HERE "The largest possible $ N_i $ is of size equal to 
%or less than the size of the largest clique in the graph."}
Let us note that a tree is a chordal graph, in which all edges and only edges are cliques.
Hence, for a tree, simplicial vertices are vertices of
degree one. The elimination of degree one vertices on a tree is an example of
simplicial elimination on a chordal graph.

For arbitrary graphs, if the maximal scope size of the
intermediate $\psi_{N_i}$ functions created during variable elimination is too large, then memory and
time required for the storage and computation quickly exceed computer
capacities. Depending on the  chosen elimination order, this maximal
scope can be reasonable from a computational point of view, or too large. 
So again, the choice of the elimination order  is crucial. In the case  of CHMM,  we can imagine two different elimination orders: either time slice per time slice, or chain by chain (we omit the observed variables that are known and do not have to be eliminated). For the first order, starting from the  oriented graph of Figure~\ref{Fig:CoupledHMM}, we first moralise it. Then, elimination of the variables $H_T^i$ of the last time step does not add any fill-in edges. However, when eliminating variables $H_{T-1}^i$ for $1 \leq i \leq I-1$, due to the temporal dependences between chain, we create an intermediate potential function depending of $I+1$ variables ($H_{T-1}^I$ and the $H_{T-2}^i$ for all chains). And when successively eliminating temporal slices, the maximal size of the intermediate potential functions created is $I+1$.
For the second elimination order, still starting from the moralised version of the oriented graph, after eliminating all variables $H_t^1$ for $1 \leq t \leq T-1$, we create an intermediate potential function depending of $T+1$ variables ($H_T^1$ and    $H_{t}^2$  for all $t$). And when successively eliminating chains, the maximal size of the intermediate potential functions created is $T+1$. So depending on the values of $I$ and $T$, we will not select the same elimination order.

\subsection{The treewidth to characterise variable elimination complexity}

 The lowest complexity achievable when performing variable elimination is characterised by
a parameter called the \emph{treewidth} of the graph associated to the
original graphical model. This concept
has been repeatedly discovered and redefined. The
treewidth of a graph is sometimes called its induced
width~\citep{dechter88d}, its minimum front size~\citep{Liu92}, its
$k$-tree number~\citep{arnborg85}, its dimension~\citep{Bertele72}, and
is also equal to the min-max clique number of $G$ minus
one~\citep{arnborg85} to name a few. The treewidth is also a key notion in the theory of graph minors~\citep{RS86,L05}.

 We insist here on two definitions. 
The first one \citep{bodlaender94} relies on the notion of \emph{induced graph} (see Definition below). It highlights
the close relationship between fill-in edges and the intermediate $N_i$ sets created during 
variable elimination. The second one~\citep{RS86,bodlaender94} is 
the most commonly used characterisation of the treewidth using so-called
tree decompositions, also known as junction trees, which are key tools to derive variable elimination algorithms.
It underlies the block-by-block elimination procedure described in Section~\ref{sec: tdbbe}.
%This parameter is exactly equal to the
%maximal size of the $N_i$ created during the elimination procedure.

\begin{definition}[induced graph]
Let $G=(V,E)$ be a graph defined by a set of vertices indexed on $V$
  and a set $E$ of edges.  Given an ordering $\pi$ of the vertices of
  $G$, the induced graph $G^{ind}_\pi$ is defined in a constructive way as follows. 
 First, $G$ and $G^{ind}_\pi$ have same vertices. 
Then for each edge in $E$  an oriented edge is added in $G^{ind}_\pi$ going from the first of the two nodes 
according to $\pi$ toward the second. 
Then each vertex $i$ of $V$  is considered one after the other following the order defined by $\pi$. 
When  vertex $i$ is treated, an oriented  edge is created between all pairs of neighbours of $i$
  in $G$ that follow $i$ in the ordering defined by $\pi$. Again the edge is going from the first of the two nodes 
according to $\pi$ toward the second.

\end{definition}

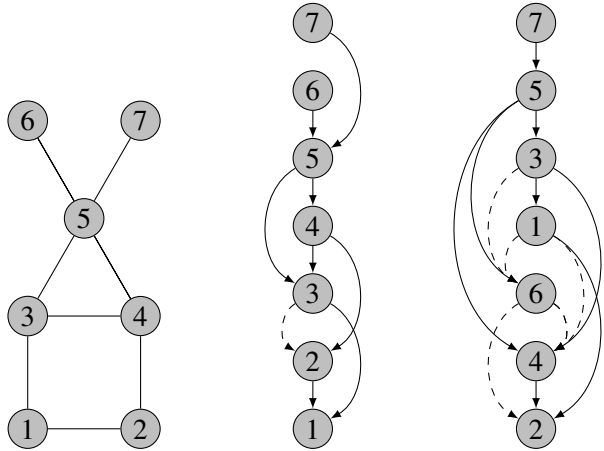
\begin{figure}[h!]
\centering
  \begin{tikzpicture}[scale=1.5,rotate=90]
  \tikzstyle{vertex}=[draw,circle,fill=gray!50,minimum size = 5pt, inner sep =2pt, font=\small]
  \path (0,0) node[vertex] (a) {1}
        (0,-1) node[vertex] (b) {2}
        (1,0) node[vertex] (c) {3}
        (1,-1) node[vertex] (d) {4}
        ( $(c) + (-30:1)$ ) node[vertex] (e) {5}
        ( $(e) + (30:1)$ ) node[vertex] (f) {6}
        ( $(e) + (-30:1)$ ) node[vertex] (g) {7};
  \draw (c) -- (a) -- (b) -- (d) -- (e) -- (d) --(c) --(e) -- (f) -- (e) -- (g);
  \end{tikzpicture}\quad\quad\quad
  \begin{tikzpicture}[scale=0.9]
  \tikzstyle{vertex}=[draw,circle,fill=gray!50,minimum size = 5pt, inner sep = 2pt,font=\small]
  \tikzstyle{arc}=[->,>=latex]
  \path (0,1)  node[vertex] (g) {7}
        (0,0)  node[vertex] (f) {6}
        (0,-1) node[vertex] (e) {5}
        (0,-2) node[vertex] (d) {4}
        (0,-3) node[vertex] (c) {3}
        (0, -4) node[vertex] (b) {2}
        (0, -5) node[vertex] (a) {1};

        \draw[arc] (g) to[bend left=60] (e);
        \draw[arc] (f) to (e);
        \draw[arc] (e) to (d);
        \draw[arc] (e) to[bend right=60] (c);
        \draw[arc] (d) to (c);
        \draw[arc] (d) to[bend left=60] (b);
        \draw[arc,dashed] (c) to[bend right=60] (b);
        \draw[arc] (c) to[bend left=60] (a);
        \draw[arc] (b) to (a);
  \end{tikzpicture}\quad\quad
  \begin{tikzpicture}[scale=0.9]
  \tikzstyle{vertex}=[draw,circle,fill=gray!50,minimum size = 5pt, inner sep =2pt, font=\small]
  \tikzstyle{arc}=[->,>=latex]
  \path (0,1)  node[vertex] (g) {7}
        (0,0)  node[vertex] (e) {5}
        (0,-1) node[vertex] (c) {3}
        (0,-2) node[vertex] (a) {1}
        (0,-3) node[vertex] (f) {6}
        (0, -4) node[vertex] (d) {4}
        (0, -5) node[vertex] (b) {2};

        \draw[arc] (g) to (e);
        \draw[arc] (e) to (c);
        \draw[arc] (e) to[bend right=60] (d);
        \draw[arc] (e) to[bend right=60] (f);
        \draw[arc,dashed] (c) to[bend right=60] (f);
        \draw[arc,dashed] (f) to[bend left=60] (d);

        \draw[arc] (c) to[bend left=60] (d);
        \draw[arc] (c) to (a);
        \draw[arc,dashed] (a) to[bend right=60] (f);
        \draw[arc,dashed] (a) to[bend left=60] (d);
        \draw[arc,dashed] (f) to[bend left=60] (d);

        \draw[arc] (a) to[bend left=60] (b);
        \draw[arc,dashed] (f) to[bend right=60] (b);

        \draw[arc] (d) to (b);
  \end{tikzpicture}
  \caption{A graph and two elimination orders. Left, the graph; middle, induced graph associated to the
    elimination order $(7> 6>5>4>3>2>1)$. Vertices are eliminated from the largest to the smallest. The maximum size of $N_i$
    sets created during elimination is $2$ (maximum number of outgoing edges) and only one (dashed)
    fill-in edge is added when vertex $4$ is eliminated; right, induced graph associated to the elimination 
    order $(7>5>3>1>6>4>2)$. The maximum size of $N_i$ sets
    created during elimination is $3$ and 5 (dashed) fill-in edges are used.}
\label{fig: good and bad order}
\end{figure}

The induced graph $G^{ind}_\pi$ is also called the \emph{fill graph}
of $G$, and the process of computing it is sometimes referred to as ``playing the
elimination game'' on $G$, as it just simulates elimination on $G$
using the variable ordering $\pi$ (see an example on Figure \ref{fig: good and bad order}). This graph is chordal~\citep{OPT-006}.
% \textcolor{red}{MV: REALLY NOT OBVIOUS TO ME?}
It is known
that every chordal graph $G$ has at least one vertex ordering $\pi$ such
that $G^{ind}_\pi = G$ (omitting the fact that edges of $G^{ind}_\pi$ are directed), called a perfect elimination ordering~\citep{fulkerson65}.  

% \centerline{FIGURE \ref{fig: good and bad order} ABOUT HERE}

The second  notion  that enables to define the treewidth is the notion of  tree decomposition. 
Intuitively, a tree decomposition of a graph $G$ organises the
vertices of $G$ in clusters of vertices which are linked by edges such
that the graph obtained is a tree.  Specific constraints on the way vertices of $G$ are associated to clusters
 in  the decomposition tree are required. These constraints ensure that the 
resulting tree   
decomposition has properties useful for building variable elimination algorithms.
%This tree must satisfy some properties that reflect some features of $G$.

\begin{definition}[tree decomposition]

  Given a graph $G = (V, E)$, a tree decomposition of $ G $, $T$,  is a tree $(\mathcal{C}, E_T)$,
  where $\mathcal{C} = \{C_1, \ldots, C_l\}$ is a family of subsets of $V$ (called
  clusters), and $E_T$ is a set of  edges between  the subsets $C_i$,
  satisfying the following properties:
  \begin{itemize}
  \item The union of all clusters $C_k$ equals $V$ (each vertex of $G$ is
    associated with at least one vertex of $T$).
  \item For every edge $(i,j)$ in $E$, there is at least one cluster $C_k$ that
    contains both $i$ and $j$.
  \item If clusters $C_k$ and $C_l$ both contain a vertex $i$ of $G$, then all clusters
    $C_s$ of $T$ in the (unique) path between $C_k$ and $C_l$
    contain $i$ as well: clusters containing vertex $i$ form a
    connected subset of $T$. This is known as the running
    intersection property.
\end{itemize}
\end{definition}

\begin{figure}[h!]
\centering
  \begin{tikzpicture}[scale=1.5,rotate=-90]
  \tikzstyle{vertex}=[draw,circle,fill=gray!50,minimum size = 5pt, inner sep = 2pt, font=\small]
  \path (0,0) node[vertex] (a) {1}
        (0,-1) node[vertex] (b) {2}
        (1,0) node[vertex] (c) {3}
        (1,-1) node[vertex] (d) {4}
        ( $(c) + (-30:1)$ ) node[vertex] (e) {5}
        ( $(e) + (30:1)$ ) node[vertex] (f) {6}
        ( $(e) + (-30:1)$ ) node[vertex] (g) {7};
  \draw (c) -- (a) -- (b) -- (d) -- (e) -- (d) --(c) --(e) -- (f) -- (e) -- (g);
%  %\filldraw[opacity=0.1] ($(a) + (180:0.3)$) arc (180:30:0.3) -- ($(d) + (30:0.3)$) arc (30:-90:0.3) --
%        ($(b) + (-90:0.3)$) arc (-90:-180:0.3) -- cycle;
%  \filldraw[opacity=0.1] ($(a) + (210:0.4)$) arc (210:90:0.4) -- ($(c) + (90:0.4)$) arc (90:0:0.4) --
%        ($(d) + (0:0.4)$) arc (0:-120:0.4) -- cycle;
%  \filldraw[opacity=0.1] ($(e) + (-60:0.3)$) arc (-60:-240:0.3) -- ($(f) + (120:0.3)$) arc (120:-60:0.3) -- cycle;
%  \filldraw[opacity=0.1] ($(e) + (-120:0.3)$) arc (-120:-300:0.3) -- ($(g) + (60:0.3)$) arc (60:-120:0.3) -- cycle;
%  \filldraw[opacity=0.1] ($(d) + (-60:0.3)$) arc (-60:-180:0.3) -- ($(c) + (180:0.3)$) arc (180:60:0.3) --
%        ($(e) + (60:0.3)$) arc (60:-60:0.3) -- cycle;
  \end{tikzpicture}\quad\quad\quad\quad
  \raisebox{0.2\height}{\begin{tikzpicture}[scale=1.5,rotate=-90]
  \tikzstyle{vertex}=[draw,circle,fill=gray!30,minimum size = 8pt, inner sep = 2pt,font=\small]
  \path (0.2,-0.8) node[vertex] (c1) {$C_1$}
        (0.8,-0.2) node[vertex] (c2) {$C_2$}
        (1.5,-0.5) node[vertex] (c3) {$C_3$}
        (2.2,0) node[vertex] (c4) {$C_4$}
        (2.2,-1) node[vertex] (c5) {$C_5$};
  \draw (c1) -- node[midway,sloped] {\scriptsize 1~~4} (c2) -- node[midway,sloped] {\scriptsize 4~~3} (c3) -- node[midway,sloped] {\scriptsize ~~~~5} (c4) -- (c3) -- node[midway,sloped] {\scriptsize 5~~~~} (c5);
  \end{tikzpicture}}
\caption{Left: graphical representation of a graphical
  model.   Right: tree decomposition  over clusters 
  $C_1 = \{1,2,4\}, \ C_2 = \{1,3,4\},\  C_3 = \{3,4,5\}, \ C_4 = \{5,6\}$ and $C_5 = \{5,7\}$. Each edge
  between two clusters is labelled by their shared variables.}
\label{fig: tree decomp}
\end{figure}
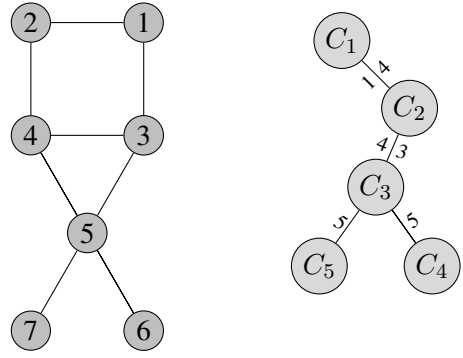

The concept of tree decomposition is illustrated in Figure \ref{fig: tree decomp}. \\

% \centerline{FIGURE \ref{fig: tree decomp} ABOUT HERE}

\begin{definition}[treewidth]
The two following definitions of the treewidth derived respectively from the notion of induced graph, and from that of tree decomposition are equivalent:
\begin{itemize}
 \item The treewidth $\TW_\pi(G)$ of a graph $G$ for the ordering $\pi$ is
 the maximum number of outgoing edges of a vertex in the induced
  graph $G^{ind}_\pi$. 
 The treewidth $\TW(G)$ of a graph $G$ is the
  minimum treewidth over all possible orderings $\pi$.
\item The width of a tree decomposition $(\mathcal{C},E_T)$ is the size of the
largest $C_i\in \mathcal{C}$ minus 1,
 and  the treewidth $\TW(G)$ of a graph is the minimum width among all its tree decompositions.
\end{itemize}

\end{definition}
It is not trivial to establish the equivalence (see \citealt[chapter 7]{Soft06}, and \citealt{TS99}). The term $\TW_\pi(G)$ is exactly the cardinality of the largest set $N_i$ created during 
variable elimination with elimination order $\pi$. 
For example, in Figure~\ref{fig: good and bad order}, the middle and
right graphs are the two induced graphs for two different orderings
 and $\TW_\pi(G)$ is equal to 2 with the first ordering and to 3 with the
second.
 It is easy to see that in this example $\TW(G) = 2$.  The treewidth of the
graph of the HMC model, and of any tree is equal to 1.

It has been established that finding a minimum treewidth ordering $\pi$ for a graph $G$, finding a minimum treewidth
 tree decomposition, or computing the treewidth of a graph are of equivalent complexity.
For an arbitrary graph, computing the treewidth is not an easy task.  Section
\ref{sec: tw}  is dedicated to this question, both from a theoretical and from a
practical point of view.

The treewidth is therefore a key indicator to answer the driving subject of this review:
will variable elimination be efficient for a given graphical model?
For instance, the principle of variable elimination was applied
to the exact computation of the normalising constant of a Markov random
field on a small $r$ by $c$ lattice in~\citet{RP04}.  For this regular
graph, it is known that the treewidth is equal to $\min(r,c)$. So exact
computation through variable elimination is  possible for 
lattices with a small  value for $\min(r,c)$ (even if $\max(r,c)$ is large).  It is however well beyond computer capacities for
real challenging problems in image analysis. In this case variable elimination can
be used to define heuristic computational solutions, such as the
algorithm of \citet{FPRW09}, which relies on the merging of exact computations on
small sub-lattices of the original lattice.

\subsection{Tree decomposition and block by block elimination}
\label{sec: tdbbe}
Given a graphical model  and a tree decomposition of its
graph, a possible alternative to solve counting or optimisation tasks
is to eliminate variables by successive blocks instead of one after
the other.
To do so, the block by block elimination
procedure~\citep{Bertele72}  relies on the tree decomposition  characterisation of the treewidth.
  The underlying idea is to apply the variable elimination
procedure on the tree decomposition, eliminating one cluster of the tree at each
step.
First  a root cluster $C_r\in\mathcal{C}$ is chosen and used to define an order of elimination 
of the clusters, by progressing from the leaves toward the root. Every eliminated cluster corresponds
 to a leaf of the  current intermediate tree.
Then each potential function $\psi_B$ is assigned  to
the cluster $C_i$ in $\mathcal{C}$ such that $B\subset C_i$ which is the
closest to the root. Such a  cluster always exists otherwise either the running intersection property would not be satisfied or the graph of the decomposition would not be a tree.
More precisely, the  procedure starts with the elimination of any leaf cluster
$C_i$ of $T$, with parent $C_j$ in $T$. Let us note
$\mathcal{B}(C_i) = \{ B \in \mathcal{B}, \psi_B \text{ assigned to } C_i\}$.
Here again, commutativity and distributivity are used to rewrite expression  (\ref{elimeq})  (with $A=V$) as follows:

$$
\BigAplus_{\bfx} \BigAtimes_{B \in \mathcal{B}} \psi_B = \BigAplus_{x_{V \setminus (C_i \setminus C_j)}} \Big[\BigAtimes_{B \in \mathcal{B} \setminus \mathcal{B}(C_i)}\psi_B \Atimes \underbrace{(\BigAplus_{x_{C_i \setminus C_j}} \BigAtimes_{B \in \mathcal{B}(C_i)} \psi_B)}_{\textrm{\scriptsize New potential function}}\Big]
$$
Note  that only variables with indices in  $C_i \setminus C_j \equiv C_i \cap (V \setminus C_j)$ are eliminated,
 even if it is common to say that the cluster has been eliminated. For instance, in the example depicted in Figure \ref{fig: tree decomp}, if the first eliminated cluster is $C_1$, the new potential
 function is $\Aplus_{x_2} \psi_{1,2}(x_1,x_2) \psi_{2,4}(x_2,x_4)$, it depends only on variables $X_1$ and $X_4$.
% All the potential functions $\psi_B$ assigned to $C_i$ are combined with
% the $\Atimes$ operator. and all variables in $C_i-C_j$ are eliminated
% together with operator $\Aplus$.
Cluster elimination continues  until no cluster is
left. The interest of this procedure is that the intermediate
potential function  created after each cluster elimination may have a scope much smaller than the treewidth,
leading to better space complexity~\cite[chapter 4]{Bertele72}. However, the time complexity is increased.

%\todoNP[inline]{est-ce que l'on arriverait a expliquer pourquoi la RIP est importante pour que la block by block elim marche?}

In summary, the lowest achievable complexity when performing variable elimination is reached for elimination orders when the cardinality of the intermediate sets $N_i$ are smaller or equal to the treewidth of $G$.
This treewidth can be determined by considering cluster sizes in tree decompositions of $G$. %\textcolor{red}{MV: READS LIKE ALL TREE DECOMPOSITION ARE NEEDED. DOESN'T SOUND VERY EFFICIENT...}
Furthermore,  any tree decomposition $T$ can be used to build an elimination order and vice versa.
Indeed, an elimination order can be defined by using  a cluster elimination order based on $T$, and by choosing 
an arbitrary order to eliminate variables with indices in the subsets $C_i \setminus C_j$.
Conversely, it is easy to build a tree
decomposition from a given vertex ordering $\pi$. Since the
induced graph $G^{ind}_\pi$ is chordal,
its maximum cliques can be identified in polynomial time. Each such  clique defines a cluster $C_i$ of the tree decomposition. Edges of $T$
can be identified as the edges of any minimum spanning tree in the graph
with vertices $C_i$ and edges $(C_i,C_j)$ weighed by $|C_i\cap C_j|$.

\paragraph{Deterministic Graphical Models.} To our knowledge, the
notion of treewidth and its properties were first identified in
combinatorial optimisation in~\citet{Bertele72}. It was then coined
``dimension'', a graph parameter which was later shown to be equivalent to the
treewidth~\citep{bodlaender98}. Variable elimination itself is related
to the Fourier-Motzkin elimination~\citep{Fourier1827}, a variable
elimination algorithm which benefits from the linearity of the handled
formulas. Variable elimination has been repeatedly rediscovered, as
non-serial dynamic programming~\citep{Bertele72}, in the David-Putnam
procedure for Boolean satisfiability problems (SAT, \citealt{Putnam60}), as Bucket elimination for the CSP
and WCSP~\citep{dechter99}, in the Viterbi and Forward-Backward
algorithms for HMM~\citep{rabiner1989} and many more.

There exists other situations where the choice of an elimination order
 has a deep impact on the complexity of the computations as in Gauss elimination scheme 
for a system of linear equations, or Choleski factorisation of very large sparse matrices, 
in which cases, the equivalence between elimination and decomposition was also used (see \citealt{bodlaender95}).

\section{Treewidth approximation for exact inference}

\label{sec: tw} 

As already mentioned, the complexity of the counting and the optimisation tasks on graphical models is strongly linked to the treewidth $\TW(G)$ 
of the underlying graph $G$.
If one could guess (one of) the optimal vertex ordering(s), $\pi^*$, leading to $\TW_{\pi^*}(G)=\TW(G)$, then, one would be able
 to achieve the ``optimal complexity'' $O(K^{\TW(G)} n) $  for solving exactly these tasks; we recall that $K$ is the maximal domain size of a variable in the graphical model.
However, the first obstacle to overcome is that the treewidth of a given graph cannot be evaluated easily: the treewidth computation 
problem is known to be NP-hard \citep{ACP1987}.
 If one has  to spend more time on finding an optimal vertex ordering than on computing probabilities
  in the underlying graphical model, 
 the utility of exact treewidth computation appears limited.
 Therefore, an alternative line of search is to look for algorithms computing a vertex
  ordering $\pi$ leading to a suboptimal width, $\TW_{\pi}(G)\geq \TW(G)$, but more efficient in terms of computational time.
In the following, we describe and empirically compare heuristics which simultaneously provide a vertex ordering and an upper bound of the treewidth. Performing inference relying on this ordering is still exact. It is not optimal in terms of time complexity, but, on  some problems, the inference can still be performed in reasonable time.

A broad class of heuristic approaches is that of greedy
algorithms~\citep{Bodlaender2010}. They use the same iterative
approach as the variable elimination algorithm (Section~\ref{sec:
  exactinference}) except that they only manipulate the graph structure. 
They do not perform any actual combination/elimination 
computation.  Starting from an empty vertex ordering and an initial
graph $G$, they repeatedly select the next vertex to add in the
ordering by locally optimising one of the following criteria:
\begin{itemize}
\item select a vertex with minimum degree in the current graph~;
\item select a vertex with minimum number of  fill-in edges in the current graph.
\end{itemize}
After each vertex selection, the current graph is modified by removing the selected vertex and making a clique
 on its neighbours. The new edges added by this clique creation are  fill-in edges. A vertex with no fill-in edges is a  simplicial vertex (see Section~\ref{var-elim-eff}).
Fast implementations of minimum degree algorithms have been developed, see  e.g., AMD~\citep{Amestoy1996} 
with time complexity in $O(nm)$~\citep{Heggernes2001} for an input  graph $G$ with $n$ vertices and $m$ edges. 
The minimum fill-in heuristics tend to be slower to compute but yield slightly better treewidth approximations in practice. 
%%I found no time complexity results on minimum fill-in algorithm
Moreover, if a  perfect elimination ordering  (i.e., adding no fill-in edges) exists, this heuristic will find it. Thus, it recognises
 chordal graphs, and it returns the optimal treewidth in this particular case. This can be easily established from results in \citealt{bodlaender2005}. 
%The proof is based on the following proposition~\cite{bodlaender2005}:\\
%{\bf Proposition:} Let $G$ be an undirected graph and $v$ be a simplicial vertex of $G$ of degree $d\geq 0$. Then,
%\begin{itemize}
% \item $\TW(G) = \max\{d,\TW(G-v)\}$,
% \item For every vertex ordering $f$ of $G-v$ of width at most $\max\{d,\TW(G-v)\}$, $(v ; f)$ is a vertex 
%ordering of $G$ with minimum treewidth.
%\end{itemize}
%From this proposition it is easy to show, by induction on the size of $G$, that a vertex ordering with minimum 
%treewidth can be obtained by successively eliminating simplicial vertices from a chordal graph (there always remains
%a simplicial vertex after, in a chordal graph, until all vertices have been removed).

Notice that there exists linear time $O(n+m)$ algorithms to detect chordal graphs as the Maximum Cardinality
 Search (MCS) greedy algorithm~\citep{Tarjan84}. MCS builds an elimination order based on the cardinality of
the already processed neighbours. However, the treewidth  approximation they return is usually worse than the previous heuristic approaches.

A simple way to improve the treewidth  bound found by these greedy algorithms is to choose between candidate vertices with same value for the selected criterion by
using a second criterion, such as minimum fill-in first and then maximum degree, or to choose at random and to 
iterate on the resulting randomised algorithms as done in \citet{Kask2011}. 
%%Other local search techniques have been tried such as~\cite{Hoesel2009} that exchange fill-in edges directly 
%on a tree decomposition representation of the solution \todoSdG{But I found no experimental results for this approach..}.

\begin{figure}[h!]
\begin{center}
\includegraphics[width=.48\textwidth]{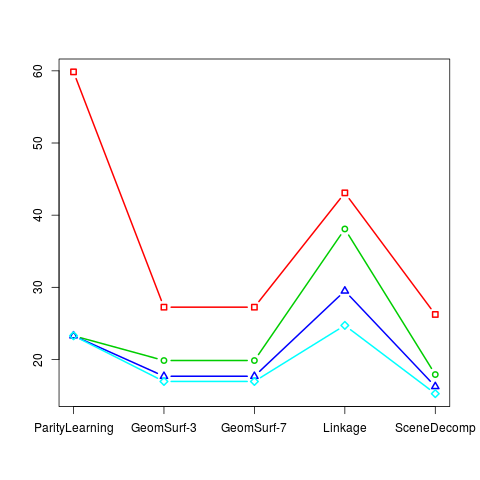}
\includegraphics[width=.48\textwidth]{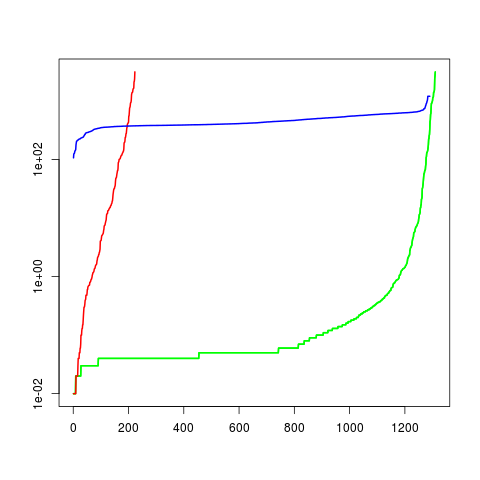}
\caption{\label{fig:joint}
Left: Comparison of treewidth upper bounds provided by \textcolor{red}{MCS} (red), \textcolor{green}{minimum degree} (green), \textcolor{blue}{minimum fill-in} (blue) and \textcolor{cyan}{randomized iterative minimum fill-in} (cyan) for  the  5 categories of problems 
%CSP/ParityLearning, MRF/Linkage, MRF/GeomSurf-3, MRF/GeomSurf-7 ,MRF/SceneDecomp 
%CSP/ParityLearning, MRF/GeomSurf-3, MRF/GeomSurf-7, MRF/Linkage, MRF/SceneDecomp.
Right: Mode evaluation by three exact methods exploiting minimum fill-in ordering or its randomized iterative version. Number of instances solved ($x$-axis) within a given CPU time in seconds (log10 scale $y$-axis) of \textcolor{red}{\sc Elim} (red), \textcolor{green}{\sc BTD} (green), and \textcolor{blue}{\sc AND/OR search} (blue).}
%CSP/ParityLearning, MRF/Linkage, MRF/GeomSurf-3, MRF/GeomSurf-7 ,MRF/SceneDecomp 
% CSP/ParityLearning, MRF/GeomSurf-3, MRF/GeomSurf-7, MRF/Linkage, MRF/SceneDecomp
% (vertical blocks from left to right). Diamond ($\diamond$, right axis): proportion (\%) of instances completely solved in less than 1 hour and 4 GB of RAM (only 20 mn for {\sc daoopt} on ParityLearning and Linkage).
\end{center}
\end{figure}

We compared the mean treewidth upper bound  found by these four   approaches (minimum degree, minimum fill-in, MCS and  randomised iterative minimum fill-in)
on a set of five WCSP and MRF benchmarks
 used as combinatorial optimisation problems in various solver competitions. 
%%\todoNP[inline]{il faudrait comparer sur autre chose que la moyenne qui n'a pas trop de sens ici. refaire
%% la table 2 en donnant plutot un rang moyen, i.e.un classement sur le nb de fois ou l'algo fournit la meilleure aprox sur l'ensemble des instances.
%%Simon et/ou Stephane, est-ce que vous pouvez le faire?}
ParityLearning is an optimisation variant of the minimal disagreement
 parity CSP problem originally contributed to the DIMACS benchmark and
 used in the Minizinc challenge~\citep{Minizinc2012}. Linkage is a
 genetic linkage analysis benchmark~\citep{UAI10}.  GeomSurf and
 SceneDecomp are respectively geometric surface labelling and scene
 decomposition problems in computer vision~\citep{OpenGM2BM}. For each  problem it is possible to vary the number of vertices and potential functions.
The number of instances per problem as well as their mean characteristics are given in Table~\ref{table-pb}.
Results
 are reported in Figure~\ref{fig:joint} (Left).The randomised iterative
 minimum fill-in algorithm used a maximum of $30,000$ iterations or
 $180$ seconds (respectively $10,000$ iterations and $60$ seconds for
 ParityLearning and Linkage), compared to a maximum of $0.37$ second
 used by the non-iterative approaches.  The minimum fill-in algorithm
 (using maximum degree for breaking ties) performed better than the
 other greedy approaches. Its randomised iterative version offers slightly 
 improved performance, at the price of some computation time.

%%TODO: replace number of factors by number of edges in the primal graph?

% \begin{table}[htb]
% \begin{center}
% %\begin{small}
% \begin{tabular}{|l|c|c|c|c|c|c|c|}
% \hline
% Problem & Nb. & n & e & MCS & Min-Degree & Min-Fill& It. Min-Fill\\
% \hline
% CSP/ParityLearning &  7 &  659 & 1246 &  59.9 &  \bf 23.3 &  \bf 23.3 &  \bf 23.3\\
% MRF/Linkage &  22 &  917 &  1560 &  43.1 &  38.1 &  29.5 &  \bf 24.7\\
% MRF/GeomSurf-3 &  300 &  505 &  2140 &  27.2 &  19.9 &  17.7 &  \bf 17.0\\
% MRF/GeomSurf-7 &  300 &  505 &  2140 &  27.2 &  19.9 &  17.7 &  \bf 17.0\\
% MRF/SceneDecomp &  715 &  183 & 672 &  26.2 &  17.9 &  16.3 &  \bf 15.3\\
% \hline
% \end{tabular}
% %\end{small}
% \caption{Comparison of treewidth (mean values on {\em Nb.} instances) produced by (iterative) greedy algorithms.
%  $n$ (resp. $e$): mean number of variables (resp. factors). Best results are in bold.\label{table-tw}}
% \end{center}
% \end{table}

% \centerline{TABLE \ref{table-pb} ABOUT HERE}
\begin{table}[h!]
\caption{Characteristics of the five optimisation problems used as benchmark.
For a given problem, several instances are available, corresponding to different numbers of variables (equal to the number of 
 vertices in the underlying graph) and different numbers of potential functions.
\label{table-pb}}
\begin{center}
%\begin{small}
\begin{tabular}{lcccc}
\hline
Problem & Nb & Mean nb & Mean nb \\
Type/Name &  of instances & of vertices &  of potential functions \\
\hline
CSP/ParityLearning &  7 &  659 & 1246 \\
MRF/Linkage &  22 &  917 &  1560 \\
MRF/GeomSurf-3 &  300 &  505 &  2140 \\
MRF/GeomSurf-7 &  300 &  505 &  2140 \\
MRF/SceneDecomp &  715 &  183 & 672\\
\hline
\end{tabular}
%\end{small}
\end{center}
\end{table}

% \begin{table}[htb]
% \begin{center}
% %\begin{small}
% \begin{tabular}{lcccc}
% \hline
% Problem & Nb & Mean nb & Mean nb \\
% Type/Name &  of instances & of vertices &  of potential functions \\
% \hline
% CSP/ParityLearning &  7 &  659 & 1246 \\
% MRF/Linkage &  22 &  917 &  1560 \\
% MRF/GeomSurf-3 &  300 &  505 &  2140 \\
% MRF/GeomSurf-7 &  300 &  505 &  2140 \\
% MRF/SceneDecomp &  715 &  183 & 672\\
% \hline
% \end{tabular}
% %\end{small}
% \caption{Characteristics of the five optimisation problems of  the benchmark.
% For a given problem, several instances are available, corresponding to differents number of variables (equal to the number of 
%  vertices in the underlying graph) and different numbers of potential functions.
% \label{table-pb}}
% \end{center}
% \end{table}

% \begin{figure}[htb]
% \begin{center}
% \includegraphics[width=.7\textwidth]{figs/Table2}
% \caption{Comparison of treewidth achieved by \textcolor{red}{MCS} (red), \textcolor{green}{Min-Degree} (green), \textcolor{blue}{Min-Fill} (blue) and \textcolor{cyan}{Min-Fill} (cyan) for problems 
% %CSP/ParityLearning, MRF/Linkage, MRF/GeomSurf-3, MRF/GeomSurf-7 ,MRF/SceneDecomp 
% CSP/ParityLearning, MRF/GeomSurf-3, MRF/GeomSurf-7, MRF/Linkage, MRF/SceneDecomp
% (vertical blocks from left to right).\label{figure-tw}}
% \end{center}
% \end{figure}

Then on the same benchmark, we  compared three exact
 methods for the task of mode evaluation that exploit either minimum fill-in ordering or its randomised iterative version: 
variable elimination (ELIM),  BTD~\citep{degivry06}, and AND/OR
Search~\citep{MarinescuDechter06}. Elim and BTD exploit the minimum fill-in ordering  while AND/OR Search used its  randomised iterative version.
 In addition, BTD and AND/OR Search exploit a  tree decomposition during a Depth First Branch and Bound method in
order to get a good trade-off between memory space and search
effort. Just like variable elimination, they have a worst-case time
complexity exponential in the treewidth. All methods were allocated
a maximum of  1 hour and 4 GB of RAM on an AMD Operon 6176 at 2.3
GHz. 
The results are reported in Figure ~\ref{fig:joint} (Right), and show that BTD was
 able to solve more problems than the two other methods for fixed CPU time.
However,  the best performing method heavily depends on the problem category. On
ParityLearning, ELIM was the fastest method, but it
ran out of memory on $83\%$ of the total set of instances, while
BTD (resp. AND/OR Search) used less than $1.7$ GB (resp.
4GB). %%0.83= (1344 - 7 - 5 - 182 - 37)/1344
The randomised  iterative minimum fill-in heuristic used by AND/OR Search in
preprocessing consumed a fixed amount of time ($\approx 180$ seconds,
included in the  CPU time measurements) larger than the
cost of  a simple minimum fill-in
heuristics run. BTD was faster than AND/OR Search to solve most of
the instances except on two problem categories (ParityLearning and
Linkage).

To perform this comparison, we ran the following implementation of each method.
The version of ELIM was the one implemented in the combinatorial
optimisation solver {\sc toolbar 2.3} (options \texttt{-i -T3},
available at
\url{mulcyber.toulouse.inra.fr/projects/toolbar}). The version of BTD was the one implemented 
in the combinatorial optimisation solver {\sc toulbar2
  0.9.7} (options \texttt{-B=1 -O=-3 -nopre}). Toulbar2 is available at
\url{www7.inra.fr/mia/T/toulbar2}. This software won  the UAI
2010~\citep{UAI10} and 2014~\citep{UAI14} Inference Competitions on
the MAP task. AND/OR Search was the version implemented in the open-source
version~1.1.2 of {\sc daoopt}~\citep{Otten12} (options \texttt{-y
  -i 35 --slsX=20 --slsT=10 --lds 1 -m 4000 -t 30000 --orderTime=180}
for benchmarks from computer vision, and \texttt{-y -i 25 --slsX=10 --slsT=6 --lds 1 -m
  4000 -t 10000 --orderTime=60} for the other benchmarks) which won the
Probabilistic Inference Challenge 2011~\citep{PIC11}, albeit with a different
closed-source version~\citep{Otten12}. 

\section{From Variable Elimination to Message Passing}
\label{sec: VE2MP}

% Message passing algorithms make use of so-called messages. These messages can be
% defined in terms of potential functions, without resorting to the definition
% of graphical models.  
On tree-structured graphical models, message
passing algorithms extend the variable elimination algorithm
by efficiently computing every marginals (or max-marginals)
simultaneously, when variable elimination only computes one.  On
general graphical models, message passing algorithms can still be
applied. They either provide approximate results efficiently, or have an exponential running cost.

We also present a less classical interpretation of the
message passing algorithms: it may be conceptually interesting to view these
algorithms as performing a re-parametrisation of the original
graphical model, i.e. a rewriting of the potentials without modifying 
the joint distribution. Instead of producing external messages, 
the re-parametrisation produces an equivalent MRF, where marginals 
can be easily accessed, and which can be better adapted that the original one for initialising further processing. 

\subsection{Message passing and belief propagation}
\label{ref:lpb}

\subsubsection{Message passing when the graph is a tree}
\label{mess-pass-graph-tree}

Message passing algorithms over trees \citep{Pearl88} can be described as an extension of
variable elimination, where the marginals or max-marginals of all variables are computed in a double
pass of the algorithm.  We depict the principle here when $G$ is a tree first and for marginal computation.
At the beginning of the first pass (the forward pass) each leaf  $X_i$ is marked as  ``processed`` and all other variables are "unprocessed". 
Then each  leaf is successively visited and the new potential $\psi_{N_i}$ is  considered as a
``message'' sent from $X_i$ to $X_{pa(i)}$ (the parents of $X_i$ in the tree),
denoted as $\mu_{i\to pa(i)}$. This message is a potential
function over $X_{pa(i)}$ only (scope of size 1). Messages are moved upward to nodes in the subgraph defined by unmarked variables. A variable is marked as processed once it has received its messages.
%Already computed messages are handled as unary potentials.

When only one variable remains unmarked (defining the root of the
tree), the combination of all the functions on this variable (messages
and possibly an original potential function involving only the root variable) will be equal to the
marginal unnormalised distribution on this variable. This
results directly from the fact that the operations performed in this 
forward pass of message passing are equivalent to
variable elimination. 
% The root of the tree defines a directed tree
% where the root is at the top, descendants are below and messages are
% flowing upwards, to the root.

To compute the marginal of another variable, one
can redirect the tree using this variable as a new root. Some subtrees will
remain unchanged (in terms of direction from the root of the subtree
to the leaves) in this new tree, and the messages in these subtrees do
not need to be recomputed. The second pass (backward pass) of the message passing algorithm exploits the fact that messages are shared between several marginal computations, to organise all these computations in a clever way, so
that in order to compute marginals of all variables,  it is enough in the second pass to send messages from the root towards the leaf. Then the marginal is computed by  combining  downward messages with upward messages arriving at a particular vertex.
%only two messages are computed for each edge, one for each
%possible direction of the edge. 
One application is the well-known Forward-Backward algorithm~\citep{rabiner1989}.

%This can be done using a simple rule:
%for an edge connecting $X_i$ and $X_j$, a message $\mu_{i\to j}$ can
%be sent to $X_j$ as soon as $X_i$ has received all the messages from
%all its neighbours different from $X_i$ (initially, there is no message
%and only leaves satisfy this condition). When each edge has
% %transmitted its 2 messages, 
% Considering an arbitrary node as the root of the tree, send messages from all leaves to 
% their single neighbour, and mark leaves as 
% it can be shown that the combination of
% all the messages and unary potentials of a variable defines a
% potential equal to the marginal unnormalised potential distribution on
% this variable.
% This follows from the fact that the message
% $\mu_{i\to j}$ obtained when eliminating $X_i$ is the marginal on
% $X_{j}$ of the problem defined by the subtree rooted in $X_{j}$ with
% just one son $X_i$ and all its descendants.

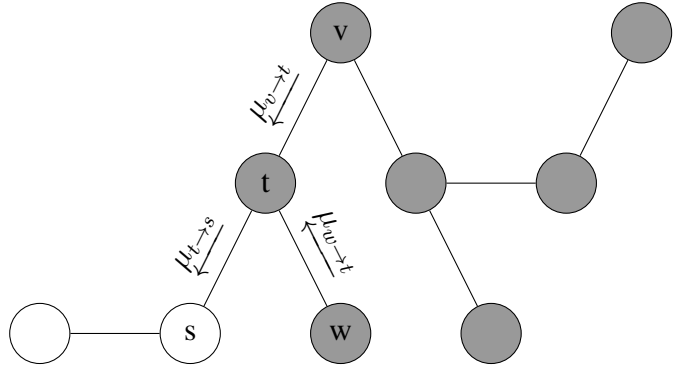
\begin{figure}[h!]
\begin{center}
\begin{tikzpicture}
 \node[draw,circle,minimum size=0.8cm] (a) at(0,0){~~};
 \node[draw,circle,minimum size=0.8cm] (s) at(2,0){s};
 \node[draw,circle,minimum size=0.8cm,fill=gray!80] (w) at(4,0){w};
 \node[draw,circle,minimum size=0.8cm,fill=gray!80] (t) at(3,2){t};
 \node[draw,circle,minimum size=0.8cm,fill=gray!80] (b) at(5,2){~~};
 \node[draw,circle,minimum size=0.8cm,fill=gray!80] (v) at(4,4){v};
 \node[draw,circle,minimum size=0.8cm,fill=gray!80] (c) at(6,0){~~};
 \node[draw,circle,minimum size=0.8cm,fill=gray!80] (d) at(7,2){~~};
 \node[draw,circle,minimum size=0.8cm,fill=gray!80] (e) at(8,4){~~};
 
 \draw (a) -- (s);
 \draw (t) -- (s) node[midway,above,sloped]{$\color{black} \mathbb{\underleftarrow{\mu_{t \to s}}}$}; % node[midway,below,sloped]{$\longrightarrow$};
 \draw (t) -- (w) node[midway,above,sloped]{$\color{black} \mathbb{\underleftarrow{\mu_{w \to t}}}$}; % node[midway,below,sloped]{$\longrightarrow$};
 \draw (t) -- (v) node[midway,above,sloped]{$\color{black} \mathbb{\underleftarrow{\mu_{v \to t}}}$}; % node[midway,below,sloped]{$\longrightarrow$};
 \draw (v) -- (b);
 \draw (b) -- (c);
 \draw (b) -- (d);
 \draw (d) -- (e);
\end{tikzpicture}
\end{center}
\caption{Example of message update on a tree. In this example, nodes $t$, $v$ and $w$ are marked, while node $s$ is still unmarked. 
$\mu_{t \to s}$ is a function of all the incoming messages to node $t$, except $\mu_{s \to t}$.\label{FigMPT}}
\end{figure}

Formally, in the message passing algorithm  for marginal evaluation over a tree $(V,E)$,
messages $\mu_{i\to j}$ are defined for each edge $(i,j)\in E$ in a 
{\em leaves-to-root-to-leaves} order; there
are $2|E|$ such messages, one for each edge direction.   Messages
$\mu_{i\to j}$ are functions of $x_j$, which are computed iteratively,
by the following algorithm:

\begin{enumerate}
\item First, messages leaving the leaves of the tree are computed:
for each $i\in V$, where $i$ is a leaf of the tree, and for $j$ the unique parent of $i$, for all $(x_i,x_j)\in\Lambda_i\times\Lambda_j$:
  $$
  \mu_{i\to j}(x_j)\leftarrow \sum_{x_i'} \psi_{ij}(x_i',x_j)\psi_i(x_i')
% \mbox{ and } \mu_{j\to i}(x_i)\leftarrow 1.
  $$
  
  Mark all leaves as {\em processed}.
\item Then,   messages are sent upward through all edges. Message updates are performed iteratively, from marked nodes $i$ to their only unmarked neighbour $j$ through edge $(i,j)\in E$.  
  Message updates take the following form for all $x_j\in\Lambda_j$:
  \begin{equation}\label{sumprod}
     \mu_{i\to j}(x_j) \leftarrow \frac{1}{K} \sum_{x_i'} \psi_{ij}(x_i',x_j)\psi_i(x_i')\prod_{k\neq j, (k,i)\in E} \mu_{k\to i}(x_i'), 
  \end{equation}
  where $K=\sum_{x_j}\sum_{x_i'} \psi_{ij}(x_i',x_j)\psi_i(x_i')\prod_{k\neq j, (k,i)\in E} \mu_{k\to i}(x_i')$.  In theory it is not necessary to normalise the messages, but this can be useful to avoid numerical problems.\\
  Mark node $j$ as {\em processed}. See Figure \ref{FigMPT} for an illustration.
\item  Send the messages downward (from root to leaves).
  This second phase of message updates takes the following form:
  \begin{itemize}
  \item Unmark root node. 
  \item While there remains a marked node, send update (\ref{sumprod}) from an unmarked node to one of its marked neighbours, unmark the corresponding neighbour.
  \end{itemize}
\item After the three above steps, messages have been transmitted through all edges in both directions. Finally, marginal distributions over variables and pairs of
  variables (linked by an edge) are computed as follows for all $(x_i,x_j) \in \Lambda_i \times \Lambda_j$:
  \begin{eqnarray}
  &  p_i(x_i) \leftarrow \frac{1}{K_i}   \psi_i(x_i)\prod_{j, (j,i)\in E} \mu_{j \to i}(x_i), \nonumber\\
   & p_{ij}(x_i,x_j) \leftarrow \frac{1}{K_{ij}}   \psi_{ij}(x_i,x_j)\prod_{k\neq j, (k,i)\in E}\mu_{k \to i}(x_i)\prod_{l\neq i, (l,j)\in E}\mu_{l \to j}(x_j)\nonumber
  \end{eqnarray}
  
  $K_i$ and $K_{ij}$ are suitable normalising constants.
\end{enumerate}

% In Step 2, the order over message updates is not specified.  In the
% case where the graphical model is a tree, it can be shown that, if an
% arbitrary root node is fixed, updating messages from leaves to the
% root node, and then back to the leaves is sufficient to compute {\em
%   exact marginals} over all variables and pairs of neighbour
% variables.

% \centerline{FIGURE \ref{FigMPT} ABOUT HERE}

Max-product and Max-sum algorithms can be
equivalently defined on a tree, for exact computation of the
$\max$-marginal of a joint distribution or its logarithm  (see chapter 8 of \citealt{B06}). In 
algebraic language, updates as defined by the formula of (\ref{sumprod}) take the general form:
$$
\forall x_j\in\Lambda_j, \mu_{i\to j}(x_j) = \Aplus_{x_i'} \psi_{ij}(x_i',x_j)\psi_i(x_i')\Atimes_{k\neq j, (k,i)\in E} \mu_{k\to i}(x_i').
$$
% As for sum-product, the resulting algorithm computes exact $\Aplus$-marginals on a
% tree-structured graphical model. It follows that the mode of the
% distribution can be computed. On general graphical models, the algorithm provides only
% approximations (see Section~\ref{fact-graph-not-tree} thereafter).  
%A {\em Max-sum} (or {\em Min-sum}) version of the
%algorithm can be defined for computing optimal solutions in cost
%function networks problems.

\subsubsection{Message passing when the factor graph is a tree}

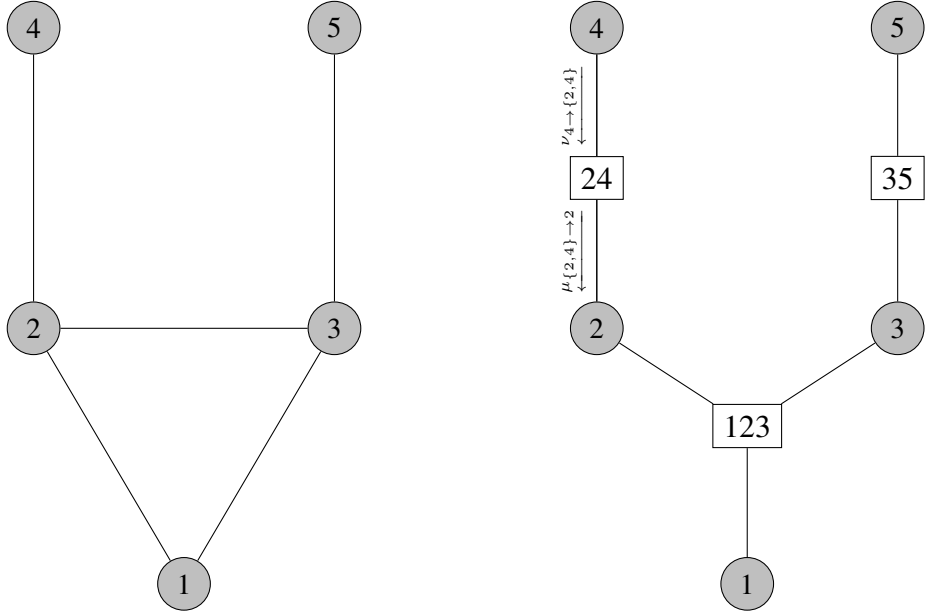
\begin{figure}[h!]
\begin{center}
 \begin{tikzpicture}[scale=1,rotate=-90]
  \tikzstyle{node}=[draw,circle,fill=gray!50,minimum size = 0.7cm, inner sep = 2pt,font=\small]
  \tikzstyle{arc}=[-]
  \path  (1.4,0) node[node] (a) {1}
         (-2, -2) node[node] (b) {2}
         (-2, 2) node[node] (c) {3}
         (-6, -2) node[node] (d) {4}
         (-6, 2) node[node] (e) {5};
  \draw[arc] (a) to (b);
  \draw[arc] (a) to (c);
  \draw[arc] (b) to (c);
  \draw[arc] (b) to (d);
  \draw[arc] (e) to (c);
  \end{tikzpicture}\quad\quad\quad\quad\quad\quad
  \begin{tikzpicture}[scale=1,rotate=-90]
  \tikzstyle{node}=[draw,circle,fill=gray!50,minimum size = 0.7cm, inner sep = 2pt,font=\small]
  \tikzstyle{factor}=[draw,minimum size = 5pt]
  \tikzstyle{arc}=[-]
  \path  (1.4,0) node[node] (a) {1}
         (-0.7,0) node[factor] (abc) {123}
         (-2,-2) node[node] (b) {2}
         (-4,-2) node[factor] (bd) {24}
         (-2, 2) node[node] (c) {3}
         (-4,2) node[factor] (ce) {35}
         (-6, -2) node[node] (d) {4}
         (-6, 2) node[node] (e) {5};
         %(-2.5, -1.5) node (f) {};
  \draw[arc] (abc) to (b);
  \draw[arc] (a) to (abc);
  \draw[arc] (abc) to (c);
  %\draw[arc] (b) to (c);
  \draw[arc] (b) to (bd);
  \draw[arc] (d) to (bd);
  \draw[arc] (c) to (ce);
  \draw[arc] (ce) to (e);
  %\draw (0,0) node{A} -- (4,2) node{B}node[midway]{M}
   \draw (bd) -- (b) node[midway,above,rotate=90]{\tiny $\color{black}\mathbb{\underleftarrow{\mu_{\{2,4\}\to 2}}}$};
    \draw (d) -- (bd) node[midway,above,rotate=90]{\tiny $\color{black}\mathbb{\underleftarrow{\nu_{4 \to\{2,4\}}}}$};
  \end{tikzpicture}
\end{center}
\caption{Left: Graphical model which structure is not a tree. Right: Corresponding factor graph, which is a tree. For applying message passing, the root is  variable 1, while variables 4 and 5 are leaves. For the left branch the first messages sent is $\nu_{4\to \{2,4\}}(x_4)$ followed by $\mu_{\{2,4\}\to 2}(x_2)$}.
\label{fig: tree factor graph}
\end{figure}

In some cases, the graph underlying the model may not be a tree, but the corresponding factor graph can be a tree, with factors potentially involving more than two variables (see Figure \ref{fig: tree factor graph}  for an example). In these cases,   message passing algorithm  can still be defined, and they lead to exact marginal value computations (or of max-marginals). However, their complexity becomes exponential in the size of the largest factor minus 1.

% \centerline{FIGURE \ref{fig: tree factor graph} ABOUT HERE}

The message passing algorithm on a tree structured factor graph exploits the same idea of shared messages than in the case a tree  structured graphical models, except that two different kinds of messages are computed: 
\begin{itemize}
\item Factor-to-variable messages: messages from a factor $B$ (we identify the factor with the subset $B$ of the potential function $\psi_B$ it represents) towards a variable $i$,  $\mu_{B\to i}(x_i)$.
\item Variable-to-factor messages:  message from a variable $i$ towards a factor $B$,  $\nu_{i\to B}(x_i)$.
\end{itemize}
These are updated in a leaf-to-root direction and then backward, as above, but two different updating rules are used instead of (\ref{sumprod}): for all  $ x_i\in\Lambda_i$
\begin{eqnarray}
  \mu_{B\to i}(x_i) & \leftarrow & \sum_{x_{B\setminus i}} 
 \left(\psi_{B}(x_B) \prod_{j \in B\setminus i} \nu_{j\to B}(x_j)\right), \nonumber\\
 %\label{FactoVar}\\
  \nu_{i\to B}(x_i) & \leftarrow & \prod_{B'\neq B, i\in B'} \mu_{B'\to i}(x_i). \nonumber 
  %\label{VartoFac}
\end{eqnarray}
Then, the marginal probabilities are obtained by local marginalisation, as in Step 4 of the algorithm of Subsection~\ref{mess-pass-graph-tree} above.
 \begin{equation}
   p_i(x_i) \leftarrow \frac{1}{K_i}   \psi_i(x_i)\prod_{B, i\in B} \mu_{B \to i}(x_i), \forall x_i\in\Lambda_i,\nonumber
\end{equation}
\noindent where $K_i$ is again a normalising constant.

\subsection{When the factor graph is not a tree}
\label{fact-graph-not-tree}

When the factor graph of the  graphical model is not a tree,
the two-pass message passing algorithm can no more be applied directly as is because of the loops.  Yet, for general graphical
models, this message passing approach can be generalised in two
different ways.

\begin{itemize}
\item A tree decomposition can be computed, as previously
  discussed in Section \ref{sec: tdbbe}. Message passing can then be applied on the resulting cluster
  tree, handling each cluster as a cross-product of variables
  following a block-by-block approach. This yields an exact algorithm,
  for which computations can be expensive (exponential in the
  treewidth) and space intensive (exponential in the separator
  size). A typical example of such algorithm is the algebraic exact
  message passing algorithm~\citep{Shafer88,Shenoy90}.
\item Alternatively, the Loopy Belief Propagation
  algorithm~\citep{FreyMacKay98} is another extension of message
  passing in which messages updates are repeated, in arbitrary order through all edges (possibly many times through each edge),
  until a termination condition is met. The algorithm returns
  approximations of the marginal probabilities (over variables and
  pairs of variables).  The quality of the approximation and the
  convergence to steady-state messages are not guaranteed, hence, the
  importance of the termination condition.  However, it has been
  observed that LBP often provides good estimates of the
  marginals, in practice. A deeper analysis of the  Loopy Belief Propagation algorithm 
  is postponed to Section~\ref{sec: variational}.
\end{itemize}

\subsection{Message Passing and re-parametrisation}

%reprendre selon le plan suivant
% -  il existe des reparametrisations (celle de koller, de soft AC, de TRW...
% elles ont un interet en preprocess ... petit schéma
% texte actuel, centré sur la reparametrisation version Koller
% petit exemple plus concret sur la reparametrisation façon Thomas en groupe de lecture

It is possible to use message passing as a re-parametrisation
technique. In this case, the computed messages are directly used to reformulate the original graphical model in a new equivalent graphical model with the same graphical structure. By ``equivalent'' we mean that the potential functions are not the same but they define  the same joint distribution as the original graphical model.

Several  methods for re-parametrisation have been proposed both in the field of probabilistic graphical models~\cite[chapters 10 and 13]{KF09}  or in the field of deterministic graphical models~\citep{cooper2010}. They all share the same advantage: the re-parameterised formulation can be computed to satisfy precise requirements. It can be designed so that the re-parameterised potential functions contains some information of interest (marginal distributions on singletons, on pairs $p_i(x_i)$, max-marginals $p^*(x_i)$, or their approximation). It can also be optimised in order to tighten a  bound on the probability of a MAP assignment~\citep{TRWS,Schiex00b,cooper2010,HK13} or on the partition function~\citep{TRWBP,liu2011,viricel2016}. Originally naive bounds can be tightened into non-naive ones by re-parametrisation.
An additional advantage of  the re-parametrised distribution is in the context of incremental updates, where we have to perform inference based on the observation of some of the variables, and new  observations (new evidence) are introduced incrementally. Since the the  re-parameterised model  already includes the result of previous inferences, it is more interesting (in term of number of message to send) to perform the updated inference when starting with this expression of the  joint distribution that with the original one ~\cite[chapter 10]{KF09}.%This idea is illustrated in Figure~\ref{Fig:reparam}.

%\centerline{FIGURE \ref{Fig:reparam} HERE}

The idea behind re-parametrisation is conceptually very simple: when a
message $\mu_{i \to j}$ is computed, instead of keeping it as a
message, it is possible to combine any potential function involving
$X_j$ with $\mu_{i \to j}$, using $\Atimes$. To preserve the joint
distribution defined by the original graphical model, we need to 
divide another potential function involving $X_j$ by the same message $\mu_{i \to j}$ using the
inverse of $\Atimes$.
% \footnote{Zeros in potential can be dealt with by
%   a proper extension of the algebraic operations, including an inverse
%   for zero. If the algebraic structure equipped with $\Atimes$ is not
%   a group but only a semi-group or a monoid, suitable pseudo inverses
%   can often be defined. See~\citet{cooper.ea04,gondran-minoux}.}

\paragraph{Example for the computation of  the max-marginals.}
We illustrate here how re-parametrisation can be exploited to extract directly all (unnormalised) max-marginals $p^*(x_i)$ from the order 1 potentials of the new model.  In this case $\psi_{ij}$ is divided by $\mu_{i\to j}$, while $\psi_{j}$ is multiplied by $\mu_{i\to j}$. The same procedure can be run  by replacing $\max$ by $+$ in the message definition to obtain all singleton marginals $P(x_i)$ instead.
% One interesting application of this latter re-parametrisation is to obtain an improved upper bound of the normalising constant $Z$. A well-know upper bound in the piece wise upper bound $\prod_i \sum_{x_i} \psi_i(x_i)$ (see \cite{KCC08}), which is a tighter ...
%pas clair encore ...

Let us consider a graphical model with 3 binary variables.  The potential functions defining the graphical model are:
\begin{eqnarray}
\psi_1(x_1)    =  (3,  1), \ \psi_2(x_2) &= &(2,6), \ \psi_3(x_3) = (3,4) \nonumber \\
\psi_{12}(x_1,x_2)   =   
\left( \begin{array}{cc}
3 & 2 \\
5 & 4
\end{array}
\right),
&  & 
\psi_{23}(x_2,x_3)   = 
\left( \begin{array}{cc}
4 & 8 \\
4 & 1
\end{array}
\right) \nonumber
\end{eqnarray}

Since the graph of the model is a single path and is thus tree-structured, we just need two passes of messages. We use vertex 2 as the root.
The first messages, from the leaves to the root, are:
\begin{eqnarray}
\mu_{1 \to 2} (x_2)& = &\max_{x_1} \psi_1(x_1) \psi_{12}(x_1,x_2) \nonumber \\
\mu_{3 \to 2} (x_2)& = &\max_{x_3} \psi_3(x_3) \psi_{23}(x_2,x_3) \nonumber
\end{eqnarray}
We obtain
\begin{eqnarray}
%\mu_{1 \to 2} (x_2)& = &  (9,6)    \nonumber \\
%\mu_{3 \to 2} (x_2)& = &  (32, 12)  \nonumber
\mu_{1 \to 2}(0) =  \max(3\times 3, 1 \times2) = 9&,  &\mu_{1 \to 2}(1) = \max( 3 \times 2, 1 \times 4) = 6   \nonumber \\
\mu_{3 \to 2}(0) = \max(3 \times 4, 4 \times 8) = 32&,  & \mu_{3 \to 2}(1) = \max(3 \times 4, 4 \times 1) = 12 \nonumber 
\end{eqnarray}

Potentials $\psi_{12}$ and $\psi_{23}$ are divided respectively by $\mu_{1 \to 2}$ and  $\mu_{3 \to 2}$, while $\psi_2$ is multiplied by these two same messages.
For instance
 \begin{eqnarray}
 \psi'_2(0) & = & \psi_2(0)  \mu_{1 \to 2}(0) \mu_{3 \to 2}(0) = 2 \times 9 \times 32 = 576\nonumber \\
 \psi'_2(1) & = & \psi_2(1)  \mu_{1 \to 2}(1) \mu_{3 \to 2}(1) =  6 \times 6 \times 12 = 532\nonumber \\
 \psi'_{12}(x_1,0) & = &  \frac{\psi_{12}(x_1,0)}{\mu_{1 \to 2}(0)}, \ \psi'_{12}(x_1,1) = \frac{\psi_{12}(x_1,1)}{\mu_{1 \to 2}(1)} \nonumber
 \end{eqnarray}
All the updated potentials are:
\begin{footnotesize}
\begin{eqnarray}
&& \psi'_1(x_1) = \psi_1(x_1)    =  (3,  1), \ \psi'_2(x_2) = (576,432), \ \psi'_3(x_3) = \psi_3(x_3) =(3,4) \nonumber \\
&& \psi'_{12}(x_1,x_2)   =   
\left( \begin{array}{cc}
3/9 & 2/6 \\
5/9 & 4/6
\end{array}
\right)
=
\left( \begin{array}{cc}
1/3 & 1/3 \\
5/9 & 2/3
\end{array}
\right) \nonumber\\
&&
\psi'_{23}(x_2,x_3)   = 
\left( \begin{array}{cc}
4/32 & 8/32 \\
4/12 & 1/12
\end{array}
\right) 
=
\left( \begin{array}{cc}
1/8 & 1/4 \\
1/4 & 1/12
\end{array}
\right) 
\nonumber
\end{eqnarray}
\end{footnotesize}

Then messages from the root towards the leaves are computed using these updated potentials:
\begin{eqnarray}
\mu_{2 \to 1} (x_1)& = &\max_{x_2} \psi'_2(x_2) \psi'_{12}(x_1,x_2) = (192,320)  \nonumber \\
\mu_{2 \to 3} (x_3)& = &\max_{x_2} \psi'_2(x_2) \psi'_{23}(x_2,x_3) = (144,144)\nonumber
\end{eqnarray}

Finally, potentials $\psi'_{12}$ and $\psi'_{23}$ are divided respectively by $\mu_{2 \to 1}$ and $\mu_{2 \to 3}$, while $\psi'_1$ and  $\psi'_3$ are multiplied by $\mu_{2 \to 1}$ and $\mu_{2 \to 3}$ respectively, leading to the re-parameterised potentials
\begin{footnotesize}
\begin{eqnarray}
&& \psi''_1(x_1)    =  (3 \times 192,  1\times 320) = (576,320), \ \psi''_2(x_2) = (576,432) \nonumber\\
&& \psi''_3(x_3) = (3\times 144,4 \times 144) = (432, 576) \nonumber \\
&& \psi''_{12}(x_1,x_2)   =   
\left( \begin{array}{cc}
\frac{1}{3\times 192} & \frac{1}{3\times 192} \\
\frac{5}{9 \times 320} & \frac{2}{3 \times 320}
\end{array}
\right)
=
\left( \begin{array}{cc}
\frac{1}{576} & \frac{1}{576} \\
\frac{1}{576} & \frac{1}{480} 
\end{array}
\right) \nonumber\\
&&
\psi''_{23}(x_2,x_3)   = 
\left( \begin{array}{cc}
\frac{1}{8 \times 144} &  \frac{1}{4 \times 144} \\
\frac{1}{3\times 144} & \frac{1}{12 \times 144}
\end{array}
\right) 
=
\left( \begin{array}{cc}
\frac{1}{1152} &  \frac{1}{576}\\
\frac{1}{432} &  \frac{1}{1728}
\end{array}
\right) 
\nonumber
\end{eqnarray}
\end{footnotesize}

Then we can directly read the (unnormalised) max marginal from the singleton potentials. For instance 
$$
\max_{x_2,x_3} \psi_1(0) \psi_2(x_2) \psi_3(x_3) \psi_{12}(0,x_2) \psi_{23}(x_2,x_3) = 576 = \psi''(0).
$$
We can check that the original graphical model and the re- parameterised one define the same joint  distribution  by comparing to the (unnormalised) probability of each possible state (see Table \ref{tab: ex reparam}).

% \begin{center}
% TABLE \ref{tab: ex reparam} ABOUT HERE
% \end{center}
\begin{table}[hb]
\caption{The unnormalised probabilities of the eight possible states in the original and re-parameterised models. One can check that the re-parameterised version describes the same joint distribution than  the original one.
\label{tab: ex reparam}}
\centering
\begin{tabular}{@{\hskip1pt}c@{\hskip1pt}c@{\hskip1pt}c@{\hskip1pt}|ccc}
\hline
      &       &       & Original & & Reparameterised \\
$x_1$ & $x_2$ & $x_3$ & $\psi_1\hfill \psi_2\hfill \psi_3 \hfill\psi_{12}\hfill \psi_{23}\!$ & & $\psi''_1 \hfill\psi''_2\hfill \psi''_3\hfill \psi''_{12}\hfill \psi''_{23}$ \\\hline
0     & 0     & 0    & $3 \times 2 \times 3 \times 3 \times 4$ & $= 216 =$ & $576 \times 576 \times 432 \times \frac{1}{576} \times \frac{1}{1152}$ \\
0     & 0     & 1    & $3 \times 2 \times 4 \times 3 \times 8$ & $= 576 =$ & $576 \times 576 \times 576 \times \frac{1}{576} \times \frac{1}{576}$ \\
0     & 1     & 0    & $3 \times 6 \times 3 \times 2 \times 4$ & $= 432 =$ & $576 \times 432 \times 432 \times \frac{1}{576} \times \frac{1}{432}$ \\
0     & 1     & 1    & $3 \times 6 \times 4 \times 2 \times 1$ & $= 144 =$ & $576 \times 432 \times 576 \times \frac{1}{576} \times \frac{1}{1728}$ \\
1     & 0     & 0    & $1 \times 2 \times 3 \times 5 \times 4$ & $= 120 =$ & $320 \times 576 \times 432 \times \frac{1}{576} \times \frac{1}{1152}$ \\
1     & 0     & 1    & $1 \times 2 \times 4 \times 5 \times 8$ & $= 320 =$ & $320 \times 576 \times 576 \times \frac{1}{576} \times \frac{1}{576}$ \\
1     & 1     & 0    & $1 \times 6 \times 3 \times 4 \times 4$ & $= 288 =$ & $320 \times 432 \times 432 \times \frac{1}{480} \times \frac{1}{432}$ \\
1     & 1     & 1    & $1 \times 6 \times 4 \times 4 \times 1$ & $= 96 =$  & $320 \times 432 \times 576 \times \frac{1}{480} \times \frac{1}{1728}$ \\
\hline
\end{tabular}
\end{table}

\paragraph{Re-parametrisation to compute pairwise or cluster joint distributions.}

One possibility is to incorporate the messages in the binary
potentials, in order to  extract directly the pairwise joint  distributions
as described in~\citet[chapter 10]{KF09}:  $\psi_{ij}$ is replaced by $\psi_{ij} \Atimes \mu_{i\to j}
\Atimes \mu_{j\to i}$ while $\psi_{i}$ is divided by $\mu_{j\to i}$ and $\psi_{j}$ by $\mu_{i\to j}$. If, for example, sum-prod messages are computed, each re-parameterised pairwise potential $\psi_{ij}$ can be shown to be equal to the (unnormalised) 
marginal distribution of $(X_i,X_j)$ (or an approximation of it if the graph is loopy).

In tree-structured problems, the resulting graphical model is said to be \emph{calibrated} to
emphasise the fact that all pairs of binary potentials sharing a common
variable agree on the marginal distribution of this common variable (here $x_i$):
$$\Aplus_{x_j} \psi_{ij} = \Aplus_{x_k} \psi_{ik}$$
In the loopy case, if an exact approach using tree decomposition is followed,
the domains of the messages have a size exponential in the size of  the intersection of pairs of
clusters, and the re-parametrisation will create new potentials of this
size. These messages are included inside the clusters. Each
resulting  cluster potential  will be the (unnormalised) marginal of the joint distribution on
the cluster variables. Again, a re-parameterised graphical model on a tree-decomposition 
is calibrated, and any two intersecting clusters agree on their marginals. This is exploited
in the Lauritzen-Spiegelhalter and Jensen sum-product-divide
algorithms~\citep{Lauritzen88,Jensen90}. Besides its interest for
incremental updates in this context, the re-parameterised graphical model using tree decomposition allows 
us to locally compute exact marginals for any set of variables in a same cluster.

If a local ``loopy'' approach is used instead, re-parameterisations do
not change scopes, but provide a re-parameterised model. Estimates
of the marginals of the original model can be read directly. For MAP, such
re-parameterisations can follow clever update rules to provide
convergent re-parameterisations maximising a well defined
criterion. Typical examples of this process are the sequential version of the tree re-weighted algorithm (TRWS \citealt{TRWS}), or the Max-Product Linear Programming algorithm (MPLP, \citealt{MPLP}) which aims optimising a bound on the non-normalised probability of the mode. 
% A seminal reference, published in Russian is~\citet{Sch76}. \textcolor{red}{MV: DO WE KEEP THIS REF IF WE DON'T EXPLICITELY USE IT (SORRY CAN'T READ RUSSIAN :-))}
These algorithms can be exact on graphical models
with loops, provided the potential functions are all submodular (often
described as the discrete version of convexity, see for instance \citealt{Topkis78, CCJK04}). 

\paragraph{Re-parametrisation in deterministic graphical models.} Re-parameterising message passing algorithms have also been used in deterministic graphical models. They are then known as ``local consistency'' enforcing or constraint propagation algorithms. On one side, a local consistency property defines the targeted calibration property. On the other side, the enforcing algorithm uses so-called Equivalence Preserving Transformations to transform
the original network into an equivalent network, i.e. defining the same
joint function, which satisfies the desired calibration/local
consistency property. Similar to LBP, Arc Consistency~\citep{Waltz72,HB06} is the most usual form of local consistency, and is related to Unit Propagation in SAT~\citep{SATHB09}. Arc consistency is exact on trees, while it is usually incrementally maintained during an exact tree search, using
re-parametrisation. Because of the idempotency of logical operators (they can be applied several time without changing the result obtained after the first application),
local consistencies always converge to a unique fix-point. 

Local consistency properties and algorithms for Weighted CSPs are
closely related to message passing for MAP. They are however 
always convergent, thanks to suitable calibration
properties~\citep{Schiex00b,cooper.ea04,cooper2010}, and  also solve
tree structured problems or  problems where all potential functions are submodular. 

These algorithms can be directly used to tackle the max-prod and sum-prod problems in a MRF. The re-parametrised MRF is then often more informative that the original one. For instance, under the simple conditions that all potential functions which scope larger than 1 are bounded by $1$, a trivial upper bound of the normalising constant $Z$ is $\prod_i \sum_{x_i} \psi_i(x_i)$. This naive upper bound can be considerably tightened by re-parameterising the MRF  using a soft-arc consistency algorithm~\citep{viricel2016}.

\section{Heuristics and approximations for inference}
\label{sec: variational}

We mainly discussed methods for exact inference in graphical models.
 They are useful if an order for variable elimination
with small treewidth is available.  In many real life applications,
interaction network are seldom tree-shaped, and their treewidth can be
large (e.g. a grid of pixel in image analysis). Consequently,  exact methods
cannot be applied anymore.  However, they can be drawn inspiration from to derive
heuristic methods for inference that can be applied to any graphical
model. What is meant by a heuristic method is an algorithm that is (a priori) not
derived from the optimisation of a particular criterion, the latter is rather termed 
an approximation method.
Nevertheless, we shall alleviate this distinction, and show that good performing message passing-based heuristics can sometimes be interpreted as approximate methods.
 For the marginalisation
task, the most widespread heuristics derived from variable elimination
and message passing principles is the Loopy Belief Propagation
% (LBP, \citealt{KFL01}) algorithm described in Section \ref{ref:lpb}, which provides approximations of the marginal distributions on singletons and pairs of variables. It  has been extended to compute an approximation of the distribution of subsets of variables with the Generalised BP algorithm~\citep{YFW05}. \textcolor{red}{MV: NUMEROUS EXTENSIONS BUT ONLY ONE CITED HERE? I found Gaussian belief propagation (GaBP), Weiss and Freeman (2001). "Correctness of Belief Propagation...". Neural Computation. 13 (10): 2173-2200, probably included in the former ref. THEN QUOTE 2-3 ALGORITHMS? MAYBE Fletcher et al. (2016) "Expectation consistent approximate...". Proceedings - ISIT 2016; 2016 IEEE International Symposium on Information Theory, 190-194 OR SURVEY PROPAGATION (SP) Braunstein et al. (2005). "Survey propagation: ...". Random Structures \& Algorithms. 27 (2): 201-226. HAPPY TO ADD THEM IF YOU WANT. OR JUST DELETE}
In the last decade, a better understanding of these heuristics
was reached, and they can now be re-interpreted as particular
instances of variational approximation methods \citep{WaJ08}.  A
variational approximation of a distribution $p$ is defined as the best
approximation of $p$ in a class $\mathcal{Q}$ of tractable
distributions (for inference), according to the Kullback-Leibler
divergence.  Depending of the application (e.g. discrete or continuous
variables), several choices for $\mathcal{Q}$ can be considered.
The connection with variable elimination principles and
treewidth is not obvious at first sight.  However, as we just emphasised, LBP can be cast in the
variational framework. The
treewidth of the chosen variational distribution depends on
the nature of the variables: $i)$ in the case of discrete variables
the treewidth need be low: in most cases, the class $\mathcal{Q}$ is formed 
by independent variables (mean field approximation), with
associated treewidth equal to 0, and some works consider a class
$\mathcal{Q}$ with associated treewidth equal to 1 (see  Section \ref{Sec:VariationApprox});
$ii)$ in the case of continuous variables, the treewidth of the
variational distribution is the same as in the original model:
$\mathcal{Q}$ is in general chosen to be the class of multivariate Gaussian
distributions, for which numerous inference tools are available.

We recall here the two key components for
a variational approximation method: the Kullback-Leibler divergence
and the choice of a class of tractable distributions. We then explain
how LBP can be interpreted as a variational approximation method. Finally we recall
the rare examples where some statistical properties of an estimator obtained using
a variational approximation
have been established. In Section \ref{sec: illust CHMM} we will illustrate how variational methods can be used to derive approximate EM algorithms for estimation in CHMM.

\subsection{Variational approximations}
\label{Sec:VariationApprox}

The Kullback-Leibler divergence $KL(q||p)= \sum_\bfx q(\bfx) \log
\frac{q(\bfx)}{p(\bfx)}$ measures the dissimilarity between two probability
distributions $p$ and $q$.  $KL$ is not symmetric, hence not a distance. 
It is positive, and it is null if and
only if $p$ and $q$ are equal.  Let us consider now that $q$ is
constrained to belong to a family $\mathcal{Q}$, which does not include
$p$.  The solution $ q^{*}$ of $ \arg\min_{q \in \mathcal{Q}} KL(q||p)
$ is then the best approximation of $p$ in $\mathcal{Q}$ according to the $KL$ 
divergence. It is called the variational distribution. 
If $\mathcal{Q}$ is a set of tractable distributions 
for inference, then marginals, mode or normalising constant of $q^{*}$ can
be used as approximations of the same quantities on $p$.
%This principle based on two  ingredients, choice of $D$ and choice of
%$\mathcal{Q}$
%has been nicely described in \cite{Minka05}: the study  presents different
%choices
%for $D$ and $\mathcal{Q}$ and discusses their properties. Here we will focus on
%the K\"ullback-Leibler divergence
%$
%KL(q||p) = \sum_x q(x) log \frac{q(x)}{p(x)}
%$
%which will enable us to revisit the LBP  heuristic as an approximation method. 

Variational approximation were originally defined in the field of statistical mechanics, as approximations
of the minimum of the free energy $F(q)$, 
$$
F(q) =  - \sum_\bfx q(\bfx)  \log \prod_{B \in \mathcal{B}} \psi_B(x_B)  + \sum_\bfx q(\bfx)  \log q(\bfx).
$$
They are also known as Kikuchi approximations or Cluster Variational Methods (CVM, 
\citealt{K51}). Minimising $F(q)$ is equivalent to minimising $KL(q||p)$, since
$$
F(q) = - \sum_\bfx q(\bfx) \log p(\bfx) - \log(Z) + \sum_\bfx q(\bfx)  \log q(\bfx) = KL(q || p) - \log(Z).
$$

The mean field approximation is the most naive approximation among the family of Kikuchi approximations. Let us consider  a binary Potts model on $n$
vertices whose joint distribution is
$$
p(\bfx) = \frac{1}{Z}  \prod_i \exp{(a_i x_i + \sum_{(i,j) \in E} b_{ij} x_i
x_j)}.
$$
We can derive its  mean field approximation, corresponding to 
the class  $\mathcal{Q}^{MF}$  of  fully factorised distributions (i.e. an
associated graph of treewidth equal to 0): 
 $\mathcal{Q}^{MF} = \{q, \text{ such that } q(\bfx) = \prod_{i \in V} q_i(x_i) \}$.

% whete $Z$ is the normalising constant. Most probability of interest can not be
% computed in this model, so it may be desirable to approximate this
%distribution
% with another distribution $\widetilde{p}(x)$ with a simple form, such as a
Since variables are binary $\mathcal{Q}^{MF}$ corresponds to joint distributions
of 
 independent Bernoulli variables with respective parameters $q_i =_{def} q_i(1)$.
Namely for all $q$ in  $\mathcal{Q}^{MF}$, we can write 
$
q(\bfx) = \prod_i q_i^{x_i} (1-q_i)^{1- x_i}.
$
The optimal approximation (in terms of Kullback-Leibler divergence) within
this class of distributions is characterised by the set of $q_i$'s which
minimise $KL(q || p)$. Denoting $E_q$ the expectation with
respect to $q$, $KL(q || p) - \log Z$ is
\begin{eqnarray*}
 & & E_q\left(\sum_i \left[ X_i \log q_i + (1-X_i) \log (1-q_i)
\right] - \sum_i a_i X_i - \sum_{(i,j) \in E} b_{ij} X_i X_j\right) \\
 & & = \sum_i \left[ q_i \log q_i + (1-q_i) \log (1-q_i) \right] -
\sum_i a_i q_i - \sum_{(i,j) \in E} b_{ij} q_i q_j.
\end{eqnarray*}
This expectation has a simple form because of the specific structure of
$q$. 
Minimising it with respect  to $q_i$ gives the fixed-point relation that each
optimal
$q_i^{MF}$'s must satisfy:
$$
\log \left[ {q_i^{MF}}/{(1 - q_i^{MF})} \right] = a_i + \sum_{j:(i,j) \in E}
b_{ij}
q_j^{MF}.
$$
leading to
$$
q_i^{MF} = \frac{e^{a_i + \sum_{j:(i,j) \in E} b_{ij}
q_j^{MF}}}{ 1+ e^{a_i + \sum_{j:(i,j) \in E} b_{ij}
q_j^{MF}}}.
$$
It is interesting to note that this expression is very close to the expression of 
  the conditional probability that $X_i=1$ given that all
other variables in the neighbourhood of $i$: 
$$
\Pr(X_i=1 \mid x_{N_i}) = \frac{e^{a_i + \sum_{j:(i,j) \in E} b_{ij}
x_j}}{ 1+ e^{a_i + \sum_{j:(i,j) \in E} b_{ij}
x_j}}.
$$

The variational distribution $q_i^{MF}$  can be interpreted as  equal to this conditional 
distribution, with neighbouring variables fixed to their expected values under 
distribution $q^{MF}$. It explains the name of mean field approximation. 
Note that in general $q_i$ is not equal to the  marginal $p_i(1)$.

The choice of the class $\mathcal Q$ is indeed a critical trade-off between opposite desirable properties: it must be large
enough to guarantee a good approximation, and small enough to contain only
 distributions for which inference in manageable. 
In the next section, a particular  choice for $\mathcal{Q}$, the Bethe class, is emphasised.
In particular, it enables us to link the LBP heuristics to variational methods.
Other choices are possible, and have been used. For instance, 
in the structured mean field setting~\citep{GJ97,WaJ08}, the distribution of a
factorial Hidden Markov Model
 is approximated in a variational approach; the multivariate  hidden state is decoupled, and
the variational distribution $q$ of the conditional distribution of hidden states is
that of  independent Markov chains (here again, the treewidth is equal to 1).
The Chow-Liu
algorithm~\citep{CL68} computes the minimum of $KL(p || q)$ for a
distribution $q$ whose associated graph is a spanning tree of the graph of $p$. 
This amounts to computing the best approximation of $p$ among graphical models
with treewidth equal to 1.
Finally, an alternative to treewidth reduction is to choose the variational approximation
 in the class of exponential distributions. This has been applied to 
Gaussian process classification~\citep{KG06} using a  multivariate Gaussian approximation 
of the posterior distribution  of the hidden field. This method relies on the use of the EP 
algorithm~\citep{Minka01}. In this algorithm,  $KL(p || q)$ is minimised instead of 
$KL(q || p)$. The choice of minimising one or the other depends on their computational tractability.

\subsection{LBP heuristics as a variational method} \label{Sec:LBP}
% The mean field approximation is the most naive approximation among the so-called Kikuchi approximations 
% from statistical mechanics, also known as  Cluster Variational Methods (CVM \citeNP{K51}).
% % They are also known as Cluster Variational
% % Methods (CVM  \cite{K51}). The CVQ method consists in
% % approximating the joint distribution $p$  by a function involving only product of
% % marginals on small subsets of variables. 
% % The idea is to express $p$ as a product  of functions of increasing order
% % (the order of a function is the dimension of its scope) and to
% % truncate this expression by replacing by 1 all functions of order larger than a
% % given one (\cite{K94, D94}).
% % Another well know Kikuchi approximation is the Bethe approximation. 
% Originally, they are not defined by a minimisation of the Kullback-Leibler divergence, 
% but as an approximation of the minimum of the free energy $H(q)$, 
% $$
% H(q) =  - \sum_x q(x)  \log \prod_{B \in \mathcal{B}} \psi_B(x_B)  + \sum_x q(x)  \log q(x).
% $$
%The two problems are equivalent since $H(q)$ is equal to $KL(q  || p) - \log Z$ and is minimum when $p = q$.
If $p$ and $q$ are pairwise MRF  whose associated graph $G=(V,E)$ is the same and is a tree, then 
$q (\bfx) = \frac{\prod_{(i,j) \in E} q(x_i,x_j)}{ \prod_{i \in V}
q(x_i)^{d_i -1}}$, where 
$\{q(x_i,x_j)\}$ and $\{q(x_i)\}$ are coherent sets of order 2 and order 1 marginals 
of $q$, respectively, and $d_i$ is the degree of vertex $i$ in the tree.
In this particular case, the  free energy  is expressed as  (see \citealt{HZW03,YFW05})
\begin{eqnarray}
F(q) &=&  - \sum_{(i,j) \in E} \sum_{x_i, x_j}   q(x_i,x_j) \log \psi(x_i,x_j) 
- \sum_{i \in V} \sum_{x_i}   q(x_i) \log \psi(x_i) \nonumber \\
&&    +  \sum_{(i,j) \in E} \sum_{x_i, x_j}   q(x_i,x_j) \log q(x_i,x_j) 
  + \sum_{i \in V} (d_i – 1) \sum_{x_i}   q(x_i) \log q(x_i) \nonumber 
\end{eqnarray}
The Bethe approximation consists in applying to an arbitrary graphical model the 
same formula of the free energy as the one used for a tree, and then in minimising it over   
the variables $ \{ q(x_i,x_j) \} $ and $ \{ q(x_i) \} $  under the constraint that they are 
probability distributions and that $ q(x_i) $ is the marginal of $ q(x_i,x_j)$. 
By extension, the 
Bethe approximation can be interpreted as a variational method
associated to the family $\mathcal{Q}^{Bethe}$ of unnormalised distributions 
that can be expressed as  $q(\bfx) = \frac{\prod_{(i,j) \in E} q(x_i,x_j)}{ 
\prod_{i \in V}
q(x_i)^{d_i -1}}$ with $\{q(x_i,x_j)\}$ and $\{q(x_i)\}$ coherent sets of order 
2 and order 1 marginals.

%The Loopy Belief Propagation (LBP) algorithm  ~\cite{FreyMacKay98} is  an
%iterative version of the Belief Propagation (BP) algorithm for marginal
%evaluation presented in Section
%\ref{sec: exactinference}.
%Since the two-steps BP is not exact for a non-tree graph, the idea of LBP is to
%send messages non only twice (upward and downward) but a large number of time
%until some convergence
%is reached. When LBP converges the algorithm outputs are approximation of all
%order 1 and order 2 marginals of $p$.

\citet{YFW05} established that the  fixed points
of LBP (when they exist, convergence is still not well understood, see
 \citealt{W00} and \citealt{MK07}) are stationary points of the problem of minimising the
Bethe free energy, or equivalently 
$KL(q||p)$ with $q$ in the class $\mathcal{Q}^{Bethe}$ of distributions. 
%$q^{2}(x) = \frac{\prod_{(i,j) \in E} q(x_i,x_j)}{ \prod_{i \in V}
%q(x_i)^{d_i -1}}$ (i.e,  order 2 truncature), where $G=(V,E)$ is still the graph
%attached to the target
%distribution $p$. 
% \todoNP[inline]{on a plutot l'habitude de dire que les points fixes de LBP sont
% des points stationnaires de l'energie libre de Bethe $F^{bethe}(q, p)$ mais
% comme 
% $F^{bethe}(q, p) = KL(q||p)  - ln(Z)$ cela revient au meme}

Furthermore, \citet{YFW05}  showed that for any  class of distributions
$\mathcal{Q}$ corresponding to a particular CVM method, 
 it is possible to define a generalised BP algorithm whose fixed  points are stationary
points of the problem of minimising $KL(q||p)$  in $\mathcal{Q}$. 

The drawback of the LBP algorithm and its extensions~\citep{YFW05} is that they
are not
associated with any theoretical bound on the error made on the marginals
approximations.
Nevertheless, LBP  is increasingly used for inference in graphical
models for its good behaviour in practice ~\citep{MKW99}. It is implemented in 
software packages for inference in graphical models like 
libDAI~\citep{libDAI} or OpenGM2~\citep{opengm2}.

\subsection{Statistical properties of variational estimates} 
Maximum-likelihood parameter estimation in graphical model is often intractable because it could require to compute marginals or normalising constants. A computationally efficient alternative to Monte-Carlo estimates are variational estimates, obtained using a variational approximation of the model.
From a statistical point-of-view, because variational estimation  is only an approximation of maximum-likelihood estimation, the resulting parameter estimates do not benefit of the typical properties of maximum-likelihood estimates (MLE), such as consistency or asymptotic normality. 
Unfortunately, no general theory exists for variational estimates, and results are available only for some specific models (see e.g. \citealt{HOW11} for the consistency in the Poisson log-normal model and \citealt{BKM17} for some other examples). 
From a more general point of view, in a Bayesian context, \citet{WaT05} and \citet{WaT06} studied the properties of variational estimates. 
They proved that the approximate conditional distribution are centred on the true posterior mean, but with a too small variance. 
\citet{CDP12} proved the consistency of the (frequentist) variational estimates of the Stochastic Block Model (SBM), while \citet{GDR12} empirically established the accuracy of their Bayesian counterpart. 
Variational Bayes estimates are also proposed by \citet{JaJ00} for logistic regression, and the approximate posterior also turns out to be very accurate. 
A heuristic explanation for these two positive examples (SBM and logistic regression) is that, in both cases, the class $\mathcal{Q}$ used for the approximate conditional (or posterior) distribution $q$ is sought so as to asymptotically contain the true conditional distribution.

\section{Illustration on  CHMM}
\label{sec: illust CHMM}
In this last section,  we illustrate how the different discussed algorithms, in the CHMM framework, perform in practice for marginal inference when the model parameters are known, and how concretely they can be exploited in the EM algorithm to perform parameter estimation.

\subsection{Comparison of exact variable elimination, variational inference and Gibbs sampling in practice}
\label{sec: expesCHMM}

We compared the following inference algorithms on the problem of computing the marginals of all the hidden variables of the CHMM model of pest propagation described in Section \ref{sec : def CHMM}, conditionally to the observed variables.  We simulated 10 datasets with the following parameters values: $\rho =0.2,  \nu =  0.5, \epsilon = 0.15, f_n = 0.3$  and $f_p  = 0.1$. For each data set, we ran the following algorithms, using libDAI software \citep{libDAI}: junction tree (JT, exact method using the principles of tree decomposition and block by block elimination); loopy belief propagation (LBP); mean field approximation (MF); and Gibbs sampling (GS, \citealt{Geman84}), with 10,000 runs, each with a burn-in of 100 iterations and then 10,000 iterations. We compared the algorithms on three criteria: running time (\textit{time} variable), mean absolute difference between the true marginal probability of state 0 and the estimated one, over all hidden variables (\textit{diff-marg} variable), and percentage of hidden variables which are not restored to their true value with the mode of the estimated marginal (\textit{error-resto} variable). The results (see Table \ref{tab: res CHMM}) are presented for increasing values of $n$, the number of rows (and also of columns) of the square grid of fields (i.e. $I = n^2$). Beyond $n=3$, JT cannot be run, so for computing \textit{diff-marg} we used the GS marginals instead of the true marginals. These results illustrate well the fact that approximate inference methods based on the principle of variable elimination are very time efficient compared to Monte-Carlo methods (less than 4 minutes for a problem with $I = 10,000$ hidden variables),  while being still very accurate. Furthermore, even a naive variational method like the mean field one can be interesting if accurate marginal estimates are not required but we are only interested in preserving their mode.\\

% \centerline{ TABLE \ref{tab: res CHMM} ABOUT HERE}
\begin{table}
 \caption{Comparison of Junction Tree (JT), Loopy Belief Propagation (LBP), Mean Field (MF) and Gibbs sampling (GS) inference algorithms on the CHMM model of pest propagation: (a) running time, in second; (b) mean difference between the true and the estimated marginal of state 0 (when JT cannot be ran we use GS marginals as true ones); (c) percentage of hidden variables not restored to their true value when using the mode of the marginals.
\label{tab: res CHMM}}
\begin{center}
 \begin{tabular}{cc} 
\begin{tabular}{lrrrr}
\hline
 \textit{time} &  JT &  LBP & MF & GIBBS\\
\hline
$n=3$ &  $0.04$  &  $0.04$ &   $0.03$ &   $1.05$ \\
$n=5$  & $-$ & $0.19$ & $0.14$ & $3.30$\\
$n=10$ & $-$ & $1.07$ & $0.65$ & $13.99$  \\
$n=100$ & $-$ & $219.31$ & $134.31$ & $3,499.6$ \\
$n=200$ & $-$ & $1,026.2$ & $746.68$ & $29,341.0$ \\
\hline
\end{tabular}
 &
  \begin{tabular}{lrrr}
\hline
\textit{diff-marg} & LBP & MF & GIBBS\\
\hline
$n=3$ & $0.001$ & $0.032$ & $0.032$ \\
$n=5$ & $0.003$ & $0.037$ & $-$ \\
$n=10$ & $0.003$ & $0.032$ & $-$   \\
$n=100$ & $0.003$ & $0.032$ & $-$\\
$n=200$ & $0.003$ & $0.032$ & $-$ \\
\hline
\end{tabular}
\\
(a) & (b)
\end{tabular}

\vspace*{1cm}
 
 \begin{tabular}{lrrrr}
\hline
\textit{error-resto} &   JT & LBP & MF & GIBBS\\
\hline
$n=3$  &  $20.00$ & $19.80$ & $19.26$ & $20.19$   \\
$n=5$  &   $-$ & $18.60$ & $19.27$ & $18.93$  \\
$n=10$ &  $-$ & $17.87$ & $17.70$ & $17.83$   \\
$n=100$ &  $-$ & $18.19$ & $18.39$ & $18.20$   \\
$n=200$ &  $-$ & $18.18$ & $18.40$ & $18.18$  \\
\hline
 & & (c) & & 
\end{tabular}
\end{center}
\end{table}

\subsection{Variational approximation for estimation in CHMM}
\label{sec: EM CHMM}
%\paragraph{Coupled hidden Markov model.} 
We now illustrate how variational approximations have been used for parameter estimation  using an EM algorithm in the case of  CHMM.

% \begin{figure}[ht]
%   \begin{center}
%   \input{figs/Fig:CoupledHMM}
%   \caption{Graphical representation of $p(h,o)$ for a coupled HMM\label{Fig:CoupledHMM}}
%  \end{center}
% \end{figure}

%\paragraph{EM and variational EM inference.}
\paragraph{Exact EM algorithm}
CHMM are examples of incomplete data models, as they
involve variables $(\bfO, \bfH)$, and only variables $\bfO$ are observed. Maximum
likelihood inference for such a model aims at finding the value of the parameters
$\theta$ which maximise the (log-)likelihood of the observed data $\bfo$, i.e.
to solve $\max_\theta \log \Pr^\theta(\bfo)$. The most popular algorithm to achieve
this task is the EM algorithm \citep{DLR77}. One of its formulation reads as an iterative 
maximisation procedure of the following functional:
% Observe that
% $$
%   \log p^\theta(o)  = E(\log p^\theta(o, H) | o) - E(\log p^\theta(H|o) | o) 
% = \max_q F(\theta, q), 
% $$
% \noindent where 
$$
 F(\theta, q)  = E_q(\log {\Pr}^\theta(\bfo, \bfH)) - E_q(\log q(\bfH)) = \log {\Pr}^\theta(\bfo) - KL(q(\bfH) || {\Pr}^\theta(\bfH|\bfo)),
$$
\noindent where $q$ stands for any distribution on the hidden variables $\bfH$, and $E_q$ stands for
the expectation under the arbitrary
distribution $q$. The EM algorithm  consists 
in alternatively maximising $F(\theta, q)$ with respect to  $q$ (E-step) and to $\theta$ (M-step). 
The solution of the E-step is $q(\bfh) = {\Pr}^\theta(\bfh|\bfo)$, since the Kullback-Leibler divergence is then minimal, and even null in this case. When replacing $q(\bfh)$ by $q(\bfh) = {\Pr}^\theta(\bfh|\bfo)$ in $F$, we obtain that the M-step amounts to maximising $ E \left[ \log {\Pr}^\theta(\bfo, \bfH) | \bfo \right] $.
%  However, this conditional distribution is intractable in
% many models and approximate inference strategies are required. \\

% \begin{figure}[ht]
%   \begin{center}
%   \input{Fig:CoupledHMM-DBN}
%   \caption{Complete graphical model of the DBN version\label{Fig:CoupledHMM-DBN}}
%  \end{center}
% \end{figure}

%\paragraph{Back to coupled HMM.}
%These two approaches have been applied to the case of  coupled HMM. 

%\paragraph{Exact calculation.} 

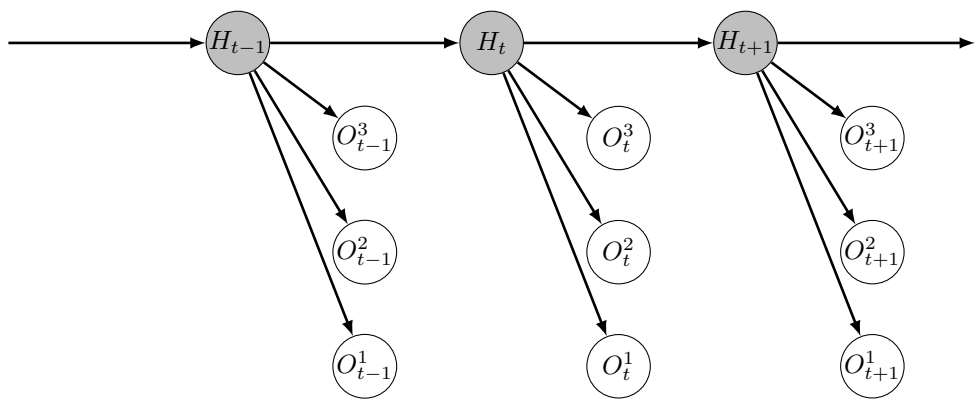
\begin{figure}[h!]
  \begin{center}
  \begin{tikzpicture}
  \node[empty] (Etm2) at (-1*\heu, 0) {\footnotesize $\quad$};
  \node[hidden] (Htm1) at (0, 0) {\footnotesize $H_{t-1}$};
  \node[hidden] (Ht) at (\heu, 0) {\footnotesize $H_{t}$};
  \node[hidden] (Htp1) at (2*\heu, 0) {\footnotesize $H_{t+1}$};
  \node[empty] (Etp2) at (3*\heu, 0) {\footnotesize $\quad$};
  
  \draw[arrow] (Etm2) to (Htm1);  \draw[arrow] (Htp1) to (Etp2);   
  \draw[arrow] (Htm1) to (Ht);   \draw[arrow] (Ht) to (Htp1);  
  
  \node[observed] (O1tm1) at (.5*\heu, -1.7*\veu) {\footnotesize $O^{1}_{t-1}$};
  \node[observed] (O1t) at (1.5*\heu, -1.7*\veu) {\footnotesize $O^{1}_{t}$};
  \node[observed] (O1tp1) at (2.5*\heu, -1.7*\veu) {\footnotesize $O^{1}_{t+1}$};
  \node[observed] (Oitm1) at (.5*\heu, -1.1*\veu) {\footnotesize $O^{2}_{t-1}$};
  \node[observed] (Oit) at (1.5*\heu, -1.1*\veu) {\footnotesize $O^{2}_{t}$};
  \node[observed] (Oitp1) at (2.5*\heu, -1.1*\veu) {\footnotesize $O^{2}_{t+1}$};
  \node[observed] (OItm1) at (.5*\heu, -.5*\veu) {\footnotesize $O^{3}_{t-1}$};
  \node[observed] (OIt) at (1.5*\heu, -.5*\veu) {\footnotesize $O^{3}_{t}$};
  \node[observed] (OItp1) at (2.5*\heu, -.5*\veu) {\footnotesize $O^{3}_{t+1}$};
  
  \draw[arrow] (Htm1) to (O1tm1);   
  \draw[arrow] (Ht) to (O1t);
  \draw[arrow] (Htp1) to (O1tp1);   
  \draw[arrow] (Htm1) to (Oitm1);   
  \draw[arrow] (Ht) to (Oit); 
  \draw[arrow] (Htp1) to (Oitp1);  
  \draw[arrow] (Htm1) to (OItm1);  
  \draw[arrow] (Ht) to (OIt);
  \draw[arrow] (Htp1) to (OItp1);  

  \end{tikzpicture}
  \caption{Graphical representation of $\Pr(\bfh,\bfo)$ for a coupled HMM when merging hidden variables at each time step \label{Fig:CoupledHMM-merge}}
 \end{center}
\end{figure}

Exact  computation of ${\Pr}^\theta(\bfh|\bfo)$ can be performed by observing that
(\ref{Eq:cHMM-joint}) can be rewritten as
$$
{\Pr}^\theta(\bfh,\bfo) \propto  \psi^{init'}(h_1) \left(\prod_{t=2}^T\psi^{M'}(h_{t-1}, h_t)\right) 
\times \left(\prod_{i=1}^I \prod_{t=1}^T \psi^E(h^i_t, o^i_t)\right),
$$
where $\psi^{init'}$ is the global initial distribution, equal to $\prod_{i=1}^I \psi^{init}(h_1^i)$, and  $\Psi^{M'}$ is the global transition probability, equal to $\prod_{i=1}^I \psi^M(h_{t-1}^i, h_{t-1}^{L-i}, h_t^i)$. This writing
 is equivalent to merging all hidden variables of a given time step. It corresponds to the graphical model given in \figurename~\ref{Fig:CoupledHMM-merge}. 
Denoting $K$ the number of possible values for each hidden variables, we end up with a regular hidden Markov model 
with $K^I$ possible hidden states. Both ${\Pr}^\theta(\bfh|\bfo)$ and its mode can then be computed in an 
exact manner with either the forward-backward recursion or the Viterbi algorithm for the mode evaluation. Both procedures have the same complexity: $O(TK^{2I})$. The exact calculation 
can therefore be achieved provided that $K^I$ remains small enough, but becomes intractable when the number of signals $I$ exceeds a few tens.

% \centerline{FIGURE \ref{Fig:CoupledHMM-merge} ABOUT HERE}

% \begin{figure}[ht]
%   \begin{center}
%   \input{figs/Fig:CoupledHMM-merge}
%   \caption{Graphical representation of $p(h,o)$ for a coupled HMM when merging hidden variables at each time step \label{Fig:CoupledHMM-merge}}
%  \end{center}
% \end{figure}

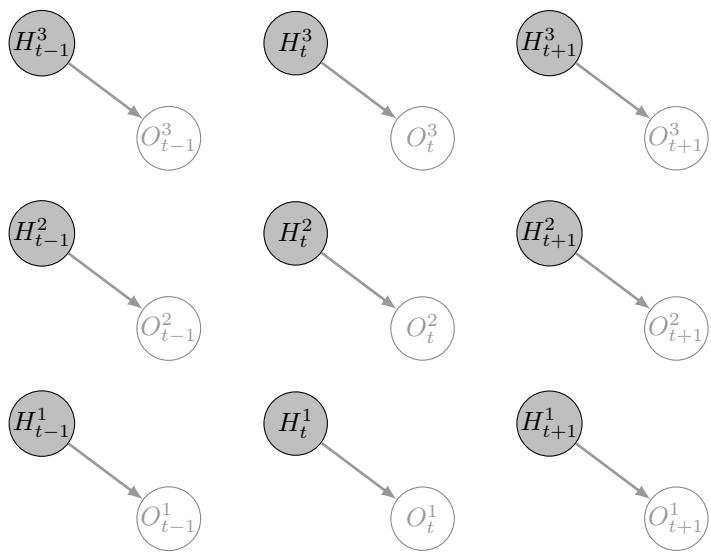
\begin{figure}[h!]
  \begin{center}
  \begin{tikzpicture}
  \node[empty] (E1tm2) at (-1*\heu, 0) {\footnotesize $\quad$};
  \node[hidden] (H1tm1) at (0, 0) {\footnotesize $H^{1}_{t-1}$};
  \node[hidden] (H1t) at (\heu, 0) {\footnotesize $H^{1}_{t}$};
  \node[hidden] (H1tp1) at (2*\heu, 0) {\footnotesize $H^{1}_{t+1}$};
  \node[empty] (E1tp2) at (3*\heu, 0) {\footnotesize $\quad$};

  \node[empty] (Eitm2) at (-1*\heu, \veu) {\footnotesize $\quad$};
  \node[hidden] (Hitm1) at (0, \veu) {\footnotesize $H^{2}_{t-1}$};
  \node[hidden] (Hit) at (\heu, \veu) {\footnotesize $H^{2}_{t}$};
  \node[hidden] (Hitp1) at (2*\heu, \veu) {\footnotesize $H^{2}_{t+1}$};
  \node[empty] (Eitp2) at (3*\heu, \veu) {\footnotesize $\quad$};

  \node[empty] (EItm2) at (-1*\heu, 2*\veu) {\footnotesize $\quad$};
  \node[hidden] (HItm1) at (0, 2*\veu) {\footnotesize $H^{3}_{t-1}$};
  \node[hidden] (HIt) at (\heu, 2*\veu) {\footnotesize $H^{3}_{t}$};
  \node[hidden] (HItp1) at (2*\heu, 2*\veu) {\footnotesize $H^{3}_{t+1}$};
  \node[empty] (EItp2) at (3*\heu, 2*\veu) {\footnotesize $\quad$};
   
  \node[observedcond] (O1tm1) at (.5*\heu, -.5*\veu) {\footnotesize $O^{1}_{t-1}$};
  \node[observedcond] (O1t) at (1.5*\heu, -.5*\veu) {\footnotesize $O^{1}_{t}$};
  \node[observedcond] (O1tp1) at (2.5*\heu, -.5*\veu) {\footnotesize $O^{1}_{t+1}$};
  \node[observedcond] (Oitm1) at (.5*\heu, .5*\veu) {\footnotesize $O^{2}_{t-1}$};
  \node[observedcond] (Oit) at (1.5*\heu, .5*\veu) {\footnotesize $O^{2}_{t}$};
  \node[observedcond] (Oitp1) at (2.5*\heu, .5*\veu) {\footnotesize $O^{2}_{t+1}$};
  \node[observedcond] (OItm1) at (.5*\heu, 1.5*\veu) {\footnotesize $O^{3}_{t-1}$};
  \node[observedcond] (OIt) at (1.5*\heu, 1.5*\veu) {\footnotesize $O^{3}_{t}$};
  \node[observedcond] (OItp1) at (2.5*\heu, 1.5*\veu) {\footnotesize $O^{3}_{t+1}$};
  
  \draw[lightarrow] (H1tm1) to (O1tm1);   
  \draw[lightarrow] (H1t) to (O1t);
  \draw[lightarrow] (H1tp1) to (O1tp1);   
  \draw[lightarrow] (Hitm1) to (Oitm1);   
  \draw[lightarrow] (Hit) to (Oit); 
  \draw[lightarrow] (Hitp1) to (Oitp1);  
  \draw[lightarrow] (HItm1) to (OItm1);  
  \draw[lightarrow] (HIt) to (OIt);
  \draw[lightarrow] (HItp1) to (OItp1);  

  \end{tikzpicture}
  \caption{Graphical representation   for the  mean-field approximation of $\Pr(\bfh,\bfo)$ in a coupled HMM.
Observed variables are indicated in light grey since they are not part of the variational distribution which is a distribution only on the hidden variables. \label{Fig:CoupledHMM-indep}}
 \end{center}
\end{figure}

\paragraph{Several variational approximations for the EM algorithm}
For more complex graphical structure, explicitly determining ${\Pr}^\theta(\bfh|\bfo)$ 
can be too expensive to perform exactly. A first approach to derive an approximate E-step
is to seek for a variational approximation  of ${\Pr}^\theta(\bfh|\bfo)$  assuming that  $q(\bfh)$ is 
restricted to a family $\mathcal{Q}$ of tractable distributions, as described in 
Section \ref{Sec:VariationApprox}. 
% This approach results in the maximisation of a lower bound of the original log-likelihood. 
The choice of $\mathcal{Q}$ is  critical, and requires achieving an acceptable balance 
between approximation accuracy and computation efficiency. 
%Some possible $\mathcal{Q}$ for coupled HMM are given below.  \\
Choosing $\mathcal{Q}$ typically amounts to breaking down some dependencies in the 
original distribution to end up with some tractable distribution. 
In the case of CHMM, the simplest distribution is the class of fully factorised 
distributions (i.e. mean field approximation), that is 
$$
\mathcal{Q}_0 = \{q: q(\bfh) = \prod_{i=1}^I \prod_{t=1}^T q_{it}(h^i_t) \}.
$$
Such an approximation of ${\Pr}^\theta(\bfh|\bfo)$ corresponds to the graphical model of \figurename~\ref{Fig:CoupledHMM-indep}. Intuitively, this
 approximation replaces the stochastic influence between the hidden variables by its mean value. 
%This approximation is often referred to as the mean-field approximation.

% \centerline{FIGURE \ref{Fig:CoupledHMM-indep} ABOUT HERE}

% \begin{figure}[ht]
%   \begin{center}
%   \input{figs/Fig:CoupledHMM-indep}
%   \caption{Graphical representation   for the independent mean-field approximation of $p(h,o)$ in a coupled HMM.
% Observed variables are indicated in light gray since they are not part of the variational distribution which is a distribution only on the hidden variables. \label{Fig:CoupledHMM-indep}}
%  \end{center}
% \end{figure}

\begin{figure}[h!]
  \begin{center}
  \begin{tikzpicture}

  \node[empty] (E1tm2) at (-1*\heu, 0) {\footnotesize $\quad$};
  \node[hidden] (H1tm1) at (0, 0) {\footnotesize $H^{1}_{t-1}$};
  \node[hidden] (H1t) at (\heu, 0) {\footnotesize $H^{1}_{t}$};
  \node[hidden] (H1tp1) at (2*\heu, 0) {\footnotesize $H^{1}_{t+1}$};
  \node[empty] (E1tp2) at (3*\heu, 0) {\footnotesize $\quad$};

  \node[empty] (Eitm2) at (-1*\heu, \veu) {\footnotesize $\quad$};
  \node[hidden] (Hitm1) at (0, \veu) {\footnotesize $H^{2}_{t-1}$};
  \node[hidden] (Hit) at (\heu, \veu) {\footnotesize $H^{2}_{t}$};
  \node[hidden] (Hitp1) at (2*\heu, \veu) {\footnotesize $H^{2}_{t+1}$};
  \node[empty] (Eitp2) at (3*\heu, \veu) {\footnotesize $\quad$};

  \node[empty] (EItm2) at (-1*\heu, 2*\veu) {\footnotesize $\quad$};
  \node[hidden] (HItm1) at (0, 2*\veu) {\footnotesize $H^{3}_{t-1}$};
  \node[hidden] (HIt) at (\heu, 2*\veu) {\footnotesize $H^{3}_{t}$};
  \node[hidden] (HItp1) at (2*\heu, 2*\veu) {\footnotesize $H^{3}_{t+1}$};
  \node[empty] (EItp2) at (3*\heu, 2*\veu) {\footnotesize $\quad$};
  
  \draw[arrow] (E1tm2) to (H1tm1);  \draw[arrow] (H1tp1) to (E1tp2);   
  \draw[arrow] (H1tm1) to (H1t);   \draw[arrow] (H1t) to (H1tp1);  
  \draw[arrow] (Eitm2) to (Hitm1);  \draw[arrow] (Hitp1) to (Eitp2);   
  \draw[arrow] (Hitm1) to (Hit);   \draw[arrow] (Hit) to (Hitp1);  
  \draw[arrow] (EItm2) to (HItm1);  \draw[arrow] (HItp1) to (EItp2);   
  \draw[arrow] (HItm1) to (HIt);   \draw[arrow] (HIt) to (HItp1);  
  
  \node[observedcond] (O1tm1) at (.5*\heu, -.5*\veu) {\footnotesize $O^{1}_{t-1}$};
  \node[observedcond] (O1t) at (1.5*\heu, -.5*\veu) {\footnotesize $O^{1}_{t}$};
  \node[observedcond] (O1tp1) at (2.5*\heu, -.5*\veu) {\footnotesize $O^{1}_{t+1}$};
  \node[observedcond] (Oitm1) at (.5*\heu, .5*\veu) {\footnotesize $O^{2}_{t-1}$};
  \node[observedcond] (Oit) at (1.5*\heu, .5*\veu) {\footnotesize $O^{2}_{t}$};
  \node[observedcond] (Oitp1) at (2.5*\heu, .5*\veu) {\footnotesize $O^{2}_{t+1}$};
  \node[observedcond] (OItm1) at (.5*\heu, 1.5*\veu) {\footnotesize $O^{3}_{t-1}$};
  \node[observedcond] (OIt) at (1.5*\heu, 1.5*\veu) {\footnotesize $O^{3}_{t}$};
  \node[observedcond] (OItp1) at (2.5*\heu, 1.5*\veu) {\footnotesize $O^{3}_{t+1}$};
  
  \draw[lightarrow] (H1tm1) to (O1tm1);   
  \draw[lightarrow] (H1t) to (O1t);
  \draw[lightarrow] (H1tp1) to (O1tp1);   
  \draw[lightarrow] (Hitm1) to (Oitm1);   
  \draw[lightarrow] (Hit) to (Oit); 
  \draw[lightarrow] (Hitp1) to (Oitp1);  
  \draw[lightarrow] (HItm1) to (OItm1);  
  \draw[lightarrow] (HIt) to (OIt);
  \draw[lightarrow] (HItp1) to (OItp1);  

  \end{tikzpicture}
  \caption{Graphical representation for the  approximation of  $\Pr(\bfh,\bfo)$ in a coupled HMM by independent heterogeneous Markov chain. 
Observed variables are indicated in light grey since they are not part of the variational distribution which is a distribution only on the hidden variables.
\label{Fig_CoupledHMM-Markov}}
 \end{center}
\end{figure}
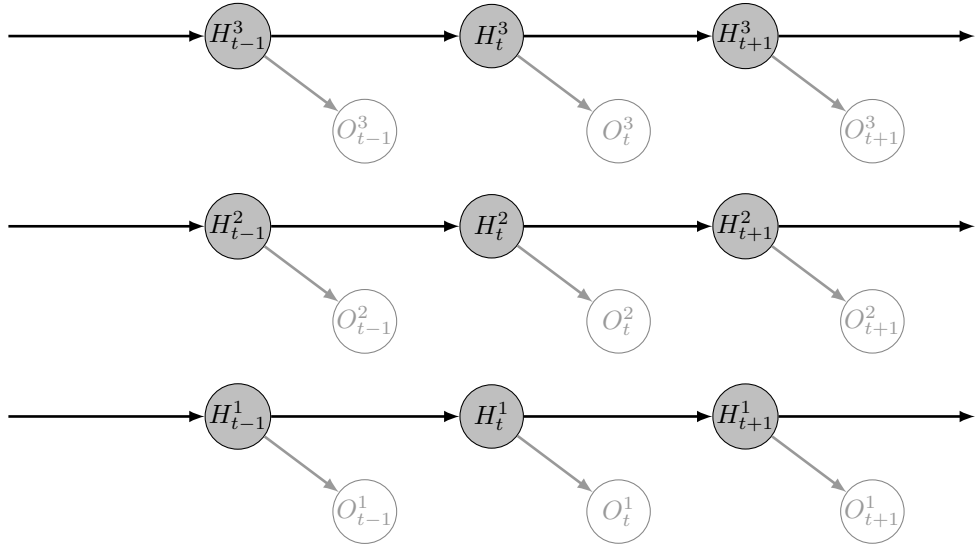

As suggested in \citet{WaJ08}, a less drastic approximation of ${\Pr}^\theta(\bfh|\bfo)$ can be obtained using the distribution family of independent heterogeneous Markov chains:
$$
\mathcal{Q}_M = \{q: q(\bfh) = \prod_i \prod_t q_{it}(h^i_t|h^i_{t-1}) \}
$$
which is consistent with the graphical representation of an independent HMM, as depicted in \figurename~\ref{Fig_CoupledHMM-Markov}.

% \centerline{FIGURE \ref{Fig_CoupledHMM-Markov} ABOUT HERE}

% 
% \begin{figure}[ht]
%   \begin{center}
%   \input{figs/Fig:CoupledHMM-Markov}
%   \caption{Graphical representation for the independent Markov approximation of  $p(h,o)$ in a coupled HMM. 
% Observed variables are indicated in light gray since they are not part of the variational distribution which is a distribution only on the hidden variables.
% \label{Fig:CoupledHMM-Markov}}
%  \end{center}
% \end{figure}

An alternative is to  use the Bethe approximation of  $F(\theta, q) $.
% consists in seeking for the maximum of an  approximation $\widetilde{F}(\theta, q)$ of $F(\theta, q)$ %which involves 
%only marginals of $p(h|o)$ on subsets of variables in  $H$ of limited size.
Then the LBP algorithm can be used to provide an approximation of the conditional marginal distributions on singletons and pairs of variables (no other marginals are involved in the E step of EM). This approach has been proposed in
%so that a LBP algorithm can apply. In this approach the optimum of approximate function $\widetilde{F}(\theta, q)$ is not guaranteed to be a bound of the original log-likelihood.
%As described above, an alternative strategy consists in maximising  an approximate version $\widetilde{F}(\theta, q)$ of ${F}(\theta, q)$. 
\citet{HZW03}.
% where the authors  approximated the negative entropy term $ E_q(\log q(H))$  in $F(\theta, q)$ by its so-called Bethe approximation as follows 
% (the first term in $F$, $E_q(\log p^\theta(o, H))$, by definition depends only on  marginals of  variables involved in the potential functions $\psi^M, \psi^C, \psi^E$).
% \begin{align}
%  & \sum_i \sum_t q^M(h^i_{t-1}, h^i_t) \log q^M(h^i_{t-1}, h^i_t)
% + \sum_t  q^C(h_t) \log q^C(h_t)  \nonumber\\
%  & + \sum_i \sum_t q^E(h^i_t, o^i_t) \log q^E(h^i_t, o^i_t)
% - \sum_i \sum_t I q(h^i_t) \log q(h^i_t) \nonumber
% \end{align}
% because each hidden variable $H^i_t$ has degree $I+1$ in the  original graphical model given in Figure \ref{Fig:CoupledHMM}.
%They derive a message passing algorithm to get the  distributions $q^M$, $q^C$ and $q^E$ minimising . 
The advantage of this approach compared to the variational approximations based on families $\mathcal{Q}_0$ or $\mathcal{Q}_M$,
is that it provides an approximation of the joint conditional distribution of  pairs of hidden variables within a same time step, instead of assuming that they are independent.

% \SR{}{
% \newcommand{\qb}{\overline{q}}
% The Bethe approximation described in Section \ref{Sec:LBP} can also be considered. In this case, the approximate distribution is characterized by the marginal distributions of each pair associated with the edge of the graphical model displayed in \figurename~\ref{Fig:CoupledHMM-LBP}, that is
% $$
% \mathcal{Q}_B = \left\{q: q(H) =  \left(\prod_{i, t} \qb(H^i_{t-1}, H^i_t) \right) \left( \prod_t \prod_{i < j} \qb(H^i_t, H^j) \right) 
% \left/ 
% \left(\prod_{i, t} \qb(H^i_t)^I \right) 
% \right. \right\}
% $$
% because each hidden variable $H^i_t$ has degree $I+1$ is this graph. Note that no inclusion hold between $\mathcal{Q}_B$ and $\mathcal{Q}_M$, so their respectively accuracy can not be compared in general.
% }
% 

\section{Conclusion and discussion}
\label{sec: conclu}
%%conclusion

This tutorial on variable elimination for exact and approximate
inference is an introduction to the basic concepts of variable
elimination, message passing and their links with variational methods.
It introduces  these fields  to statisticians
confronted with inference in graphical models. The main message is that
exact inference should not be systematically ruled out.  Before looking for an efficient approximate method, a wise advice would be to
try to evaluate the treewidth of the graphical model. In practice, this
question is not easy to answer. Nevertheless several  algorithms exist that provide  an upper bound of the treewidth together with the associated  variable elimination order (minimum degree, minimum fill-in, maximum cardinality search, ...). Even if it is not optimal, this ordering can be used to perform exact inference if the bound is small enough.

%%EXAMPLES OF APPLICATIONS. TO GM FOR THE FIRST 2 PARAGRAPHS, THEN MORE APPLICATIONS WITH REAL DATA.
% One of the many uses of the treewidth can be found in the algorithm  by~\citet*{Lauritzen88}  to  solve  an  inference  problem (e.g. graph reconstruction) on  so-called Bayesian networks. Typically, this family of algorithms first performs a tree decomposition. Next, a dynamic programming algorithm is performed from this decomposition. The dynamic programming algorithm has exponential complexity in the treewidth of the obtained decomposition. It is hence only feasible for reasonable values of the width, a feature often encountered in many real-world applications.
Examples where the low treewidth of the graphical model has been successfully exploited to perform exact inference in problems apparently too complex are numerous.
\Citet{korhonen2013} simplified the NP-hard problem of learning the structure of a Bayesian network from data %(hence from variable distributions via their independence relationships)
when the underlying network has ``low'' treewidth. They proposed an exact score-based algorithm to learn graph structure using dynamic programming. \Citet{berg2014} compared their approach with an encoding of the algorithm in the framework of Maximum Satisfiability and improved performances on classical Machine Learning datasets with networks up to 29 nodes.
%Divyanshu Vats and José M. F. Moura , “Graphical Models as Block-Tree Graphs”, https://arxiv.org/pdf/1007.0563.pdf -> block-tree instead of junction tree. STrength as that clusters of the block-tree are a little bit larger but disjoint. They hence can be constructed via a proposed algorithm.
\Citet{akutsu2009} tackled the problem of Boolean acyclic network completion. More specifically, the aim is to achieve the smallest number of modifications in the network, so that the distribution is consistent with the binary observations at the nodes. The authors established the general NP-completeness of the problem, even for tree-structured networks. They however reported that these problems can be solved in polynomial time for network with bounded treewidth and in-degree, and with enough samples (in the order of at least $ \log $ of the number of nodes). Their findings were applied~\citep{tamura2014} to obtain the sparsest possible set of modifications in the activation and inhibition functions of a signalling network (comprising 57 nodes and 154 edges) after a hypothesised cell-state alteration in colorectal cancer patients.
%% NP : j'ai enlevé le suivant car il me semble qu'il n'y a pas de modèle graphique dans cet article, que des graphes.
% Incidentally, \citet{akiba2012} noted that methods for shortest path length queries could be applied to real-world complex network when assuming a bounded treewidth on the fringe of a dense core. In particular, they implement an improved tree-decomposition-based technique for complex networks.
\Citet{xing2004} introduced two Bayesian probabilistic graphical modelling of genomic data analysis devoted to (i) the identification of motifs and cis-regulatory modules from transcriptional regulatory sequences, and (ii) the haplotype inference from genotypes of SNPs (Single Nucleotide Polymorphisms). The inference for these two high-dimensional models on hybrid distributions is very complex to compute. The author noted that the exact computation (e.g. of MAP or marginal distributions) might be feasible for models of bounded tree-width, if a good variable ordering was available. However, the question on how to find this latter one is not addressed, and an approximate generalised mean field inference algorithm is developed.
Finally, the reader can find in \Citealt{berger2008} more illustration of how the notion of treewidth can help simplifying the parametrisation of many algorithms in bioinformatics.
%\citet{liu2006} introduces a graph decomposition of mass peak data (to identify and analyse proteins)  to obtain an optimal de novo sequencing and spectral alignment in linear and quadratic time respectively. -> NO DISTRIBUTION BUT CLEVER USE OF A LIMITED TREEWIDTH!
%Liu, C., Song, Y., Yan, B., and Cai, L.L. (2006) Fast de novo peptide sequencing and spectral alignment via tree decomposition. In Pacific Symposium on Biocomputing. 
%\citet{peng2015} obtained 3d coordinates of proteins. Complexes unlike monomer have large treewidth. Algo to treat complex as monomer by excluding interactions to be neglected. Novel decomposition here. Overcomes the large treewidth problem by a clever decomposition. Solved by sub-gradient algorithm. -> NO DISTRIBUTION BUT WORTH HAVING A LOOK! NOT PUBLISHED.
%Peng., J., Hosur, R., Berger, B., and Xu, J. (2015) iTreePack: Protein Complex Side-Chain Packing by Dual Decomposition. 
% They allowed
% % \cite{F81} to build phylogenetic trees from DNA
% %sequences,
%  \cite{PSD00} to classify individuals in a structured
% population with genetic marker data or \cite{F04} to reconstruct regulatory
% networks from gene expression data with Bayesian Networks  to cite only a
% few of some fruitful marriages in bioinformatics applications. In epidemiology,
% the pioneering work of \cite{BYM91} is at the origin of numerous modeling works
% for disease mapping based on Markov random Field. In ecology,
% Hidden Markov Random Fields have been successfully used for the detection
% of spatial structures in populations \cite{FAG06}.

For the reader interested in testing the inference algorithms presented in this article, the list  provided by  Kevin Murphy (https://www.cs.ubc.ca/~murphyk/Software/bnsoft.html), even though slightly out-dated, gives a good idea of the variety of existing software packages, most of them being dedicated to a particular family of graphical model (directed, or undirected). 
One of the reason why variable elimination based technique for inference in graphical model is not well widespread outside the communities of researchers in  Computer Science and Machine Learning is probably that there exist no software being both generic and with an easy interface from R, Python or Matlab.

Obviously this tutorial is not exhaustive, since we chose to
focus on fundamental concepts. While many important results on
treewidth and graphical models have several decades in age, the area
is still lively, and we now broaden our discussion  to a few recent works which 
tackle some challenges related to the computation of the treewidth.

Because they offer efficient algorithms, graphical models with a bounded
treewidth offer an attractive target when the aim is to learn a model
that best represents some given sample. In~\cite{KB12}, the problem
of learning the structure of an undirected graphical model with
bounded treewidth is approximated by a convex optimisation problem. The
resulting algorithm has a polynomial time complexity. As discussed
in~\cite{KB12}, this algorithm is useful to derive tractable
candidate distributions in a variational approach, enabling to go
beyond the usual variational distributions with treewidth zero or 1.

For optimisation (MAP), other exact techniques are offered by tree
search algorithms such as Branch and Bound~\citep{LW66}, that
recursively consider possible conditioning of variables.  These
techniques often exploit limited variable elimination processing to
prevent exhaustive search, either using message-passing like
algorithms~\citep{cooper2010} to compute bounds that can be used for
pruning, or by performing ``on-the-fly'' elimination of variables with
small degree~\citep{Larrosa00}.

Beyond pairwise potential functions, the time needed for  simple update rules of
message passing becomes exponential  in the size of the scope of
the potential functions. 
However, for specific potential  functions involving many (or all) variables, 
exact messages  can be computed
in reasonable time, even in the context of convergent message passing
for optimisation. This can be done using polytime graph optimisation algorithms such as
shortest path or mincost flow algorithms. Such functions are known as
global potential functions~\citep{vicente2008,werner2008} in probabilistic
graphical models, and as global cost functions~\citep{lee2009,BES12,LL2012asa} in deterministic Cost Function Networks.

Different problems appear with continuous variables, where counting
requires integration of functions. Here again, for specific families
of distributions, exact (analytic) computations can be obtained for
distributions with conjugate distributions. For message passing,
several solutions have been proposed. For instance, a recent message
passing scheme proposed by \cite{NW13} relies on the combination of
orthogonal series approximation of the messages, and the use of
stochastic updates. We refer the reader to references in~\cite{NW13}
for a state-of-the-art of alternative methods dealing with continuous
variables message passing. Variational methods are also largely exploited for continuous variables, in particular in Signal Processing \citep{SQ06}.

Finally, we have excluded Monte-Carlo methods from the scope of our
review. However the combination of the inference methods presented in this article and stochastic methods for inference is a new area that researchers start exploring. Recent sampling algorithms have been proposed that use
exact optimisation algorithms to sample points with high probability
in the context of estimating the partition function. Additional control in the sampling method is needed
 to avoid biased estimations: this may be hashing functions
enforcing a fair sampling~\citep{ermon2014} or randomly perturbed
potential functions using a suitable noise
distribution~\citep{hazan2013}. More recently, Monte-Carlo and variational approaches have been combined to propose Discrete Particle Variational Inference~\citep{Saeedi17}, an algorithm that benefits from the accuracy of the former and the rapidity of the latter.

We hope this review will enable more cross-fertilisations of this
sort, combining statistics and computer science, stochastic and
deterministic algorithms for inference in graphical models.

\bibliography{twreview}

\bibliographystyle{chicago}

\newpage

\include{figures}
\include{tables}

\end{document}